\documentclass{article}

\usepackage{PRIMEarxiv}

\usepackage[utf8]{inputenc} 
\usepackage[T1]{fontenc} 
\usepackage[colorlinks,
      linkcolor=black,    
      anchorcolor=blue, 
      citecolor=black,    
      ]{hyperref}
\usepackage{url}   
\usepackage{upgreek}
\usepackage{booktabs}  
\usepackage{amsfonts}  
\usepackage{nicefrac}  
\usepackage{microtype}  
\usepackage{lipsum}
\usepackage{amsmath}   
\usepackage{fancyhdr}  
\usepackage{graphicx}  
\usepackage{subfigure}  
\usepackage{float}    
\usepackage{ragged2e} 
\usepackage{booktabs,makecell, multirow, tabularx}
\usepackage{authblk}
\usepackage{CJKutf8}
\graphicspath{{media/}} 

\newcounter{para}[subsubsection]
\newcommand\numberedParagraph{\par\refstepcounter{para}\thepara)\space}

\title{On the Opportunities of Green Computing: A Survey
}

\author[17,*]{You Zhou}
\author[17,*]{Xiujing Lin}
\author[5,*]{Xiang Zhang}
\author[14,17,*]{Maolin Wang}
\author[3,14,*]{Gangwei Jiang}
\author[6,*]{Huakang Lu}
\author[6,*]{Yupeng Wu}
\author[3,*]{Kai Zhang}
\author[3,*]{Zhe Yang}
\author[3,*]{Kehang Wang}
\author[3,*]{Yongduo Sui}
\author[1,*]{Fengwei Jia}
\author[4,*]{Zuoli Tang}
\author[17,*]{Yao Zhao}
\author[16,*]{Hongxuan Zhang}
\author[8,*]{Tiannuo Yang}
\author[12,*]{Weibo Chen}
\author[12,*]{Yunong Mao}
\author[9,*]{Yi Li}
\author[9,*]{De Bao}
\author[10,*]{Yu Li}
\author[2,*]{Hongrui Liao}
\author[7,*]{Ting Liu}
\author[7,*]{Jingwen Liu}
\author[7,*]{Jinchi Guo}
\author[14]{Xiangyu Zhao}
\author[13]{Ying WEI}
\author[6]{Hong Qian}
\author[3]{Qi Liu}
\author[3]{Xiang Wang}
\author[1]{Wai Kin (Victor) Chan}
\author[15]{Chenliang Li}
\author[8]{Yusen Li}
\author[12]{Shiyu Yang}
\author[9]{Jining Yan}
\author[10,11]{Chao Mou}
\author[2]{Shuai Han}
\author[7]{Wuxia Jin}
\author[17]{Guannan Zhang}
\author[17,+]{Xiaodong Zeng}
\affil[1]{Tsinghua-Berkeley Shenzhen Institute (TBSI), Tsinghua Shenzhen International Graduate School (SIGS), Tsinghua University}
\affil[2]{Shanghai Jiaotong University}
\affil[3]{University of Science and Technology of China}
\affil[4]{Key Laboratory of Aerospace Information Security and Trusted Computing, Ministry of Education, School of Cyber Science and Engineering, Wuhan University}
\affil[5]{Huazhong University of Science and Technology}
\affil[6]{Shanghai Institute of AI for Education and School of Computer Science and Technology, East China Normal University}
\affil[7]{Xi'an Jiaotong University}
\affil[8]{College of Computer Science, Nankai University}
\affil[9]{School of Computer Science, China University of Geosciences}
\affil[10]{School of Information Science and Technology, Beijing Forestry University}
\affil[11]{Engineering Research Center for Forestry-Oriented Intelligent Information Processing of National Forestry and Grassland Administration}
\affil[12]{Guangzhou University}
\affil[13]{Nanyang Technological University}
\affil[14]{City University of Hong Kong}
\affil[15]{Wuhan University}
\affil[16]{State Key Laboratory for Novel Software Technology, Nanjing University}
\affil[17]{Ant Group}
\affil[*]{Authors contribute equally. The ranked order is based on the survey's order of sections.}
\affil[+]{Corresponding author: xiaodong.zxd@antgroup.com}

\begin{document}
\begin{CJK}{UTF8}{gbsn}
\maketitle
\begin{abstract}
Artificial Intelligence (AI) has achieved significant advancements in technology and research with the development over several decades, and is widely used in many areas including computing vision, natural language processing, time-series analysis, speech synthesis, etc. During the age of deep learning, especially with the arise of Large Language Models, a large majority of researchers' attention is paid on pursuing new state-of-the-art (SOTA) results, resulting in ever increasing of model size and computational complexity. The needs for high computing power brings higher carbon emission and undermines research fairness by preventing small or medium-sized research institutions and companies with limited funding in participating in research.
To tackle the challenges of computing resources and environmental impact of AI, Green Computing has become a hot research topic. In this survey, we give a systematic overview of the technologies used in Green Computing. We propose the framework of Green Computing and devide it into four key components: (1) Measures of Greenness, (2) Energy-Efficient AI, (3) Energy-Efficient Computing Systems and (4) AI Use Cases for Sustainability. For each components, we discuss the research progress made and the commonly used techniques to optimize the AI efficiency. We conclude that this new research direction has the potential to address the conflicts between resource constraints and AI development. We encourage more researchers to put attention on this direction and make AI more environmental friendly.
\end{abstract}


\newpage

\tableofcontents

\newpage

\section{Introduction}

Artificial Intelligence (AI) aims to mimic human cognitive abilities and perform tasks with varying degrees of autonomy. It involves processes such as problem-solving, learning, reasoning, perception, and language understanding\cite{stuart2016artificial}. With the development over several decades, AI has achieved significant advancements in technology and research. And in recent years, to tackle the challenges of computing resources and environmental impact of AI, Green Computing has become a hot research topic\cite{Schwartz_Dodge_Smith_Etzioni_2020, Xu_Zhou_Fu_Zhou_Li_2021, Chen_Wu_Chan_Li_Ong_2023, Wu_Raghavendra_Gupta_Acun_Ardalani_Maeng_Chang_Behram_Huang_Bai_et_al_2021}. In this chapter, we analyse the Research and Development (R\&D) trend of AI, why we need green computing, and give the outline of this survey.

\subsection{The AI's Research and Development (R\&D) Trend}
\label{ai_r_and_d_trend}

The early stage of AI is mainly based on symbols, logic-theory and expert rules to mimic the decision-making of human, which requires significant human efforts. And with the rise of machine learning, especially neural networks, AI has entered the stage of deep learning, where computers learn from data instead of being explicitly programmed\cite{goodfellow2016deep}. Deep learning allows the AI systems to automatically learn and extract abstract fetures or representations from large amount of data, thus requires more computing resources and training data but less human involvements. Since deep learning is widely adopted both in academic and industry now, we focus on the deep learning stage of AI and summarize its R\&D trends as below:
\textit{\numberedParagraph{Researchers pay more attention on accuracy versus efficiency}}

\begin{figure}[!htb]
  \centering
  \subfigure[Number of Public Benchmarks for Different Tasks]{
   \label{fig:1_number_of_benchmarks} 
   \includegraphics[width=.45\textwidth]{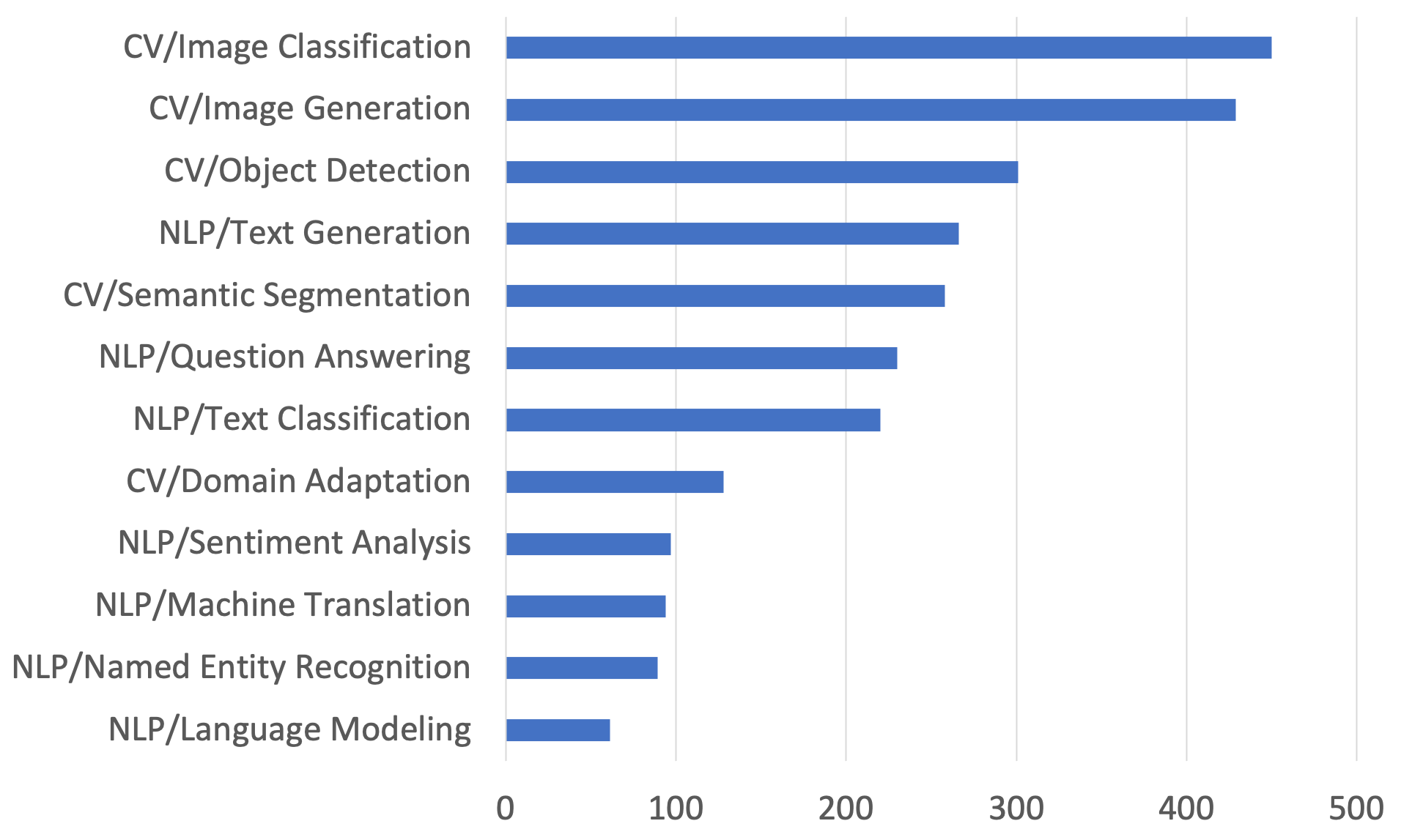}}
  \hspace{0.5in} 
  \subfigure[Number of Papers per Year (2014-2023) for Image Segmentation benchmark ADE20K]{
   \label{fig:1_number_of_papers_for_ade20k} 
   \includegraphics[width=.45\textwidth]{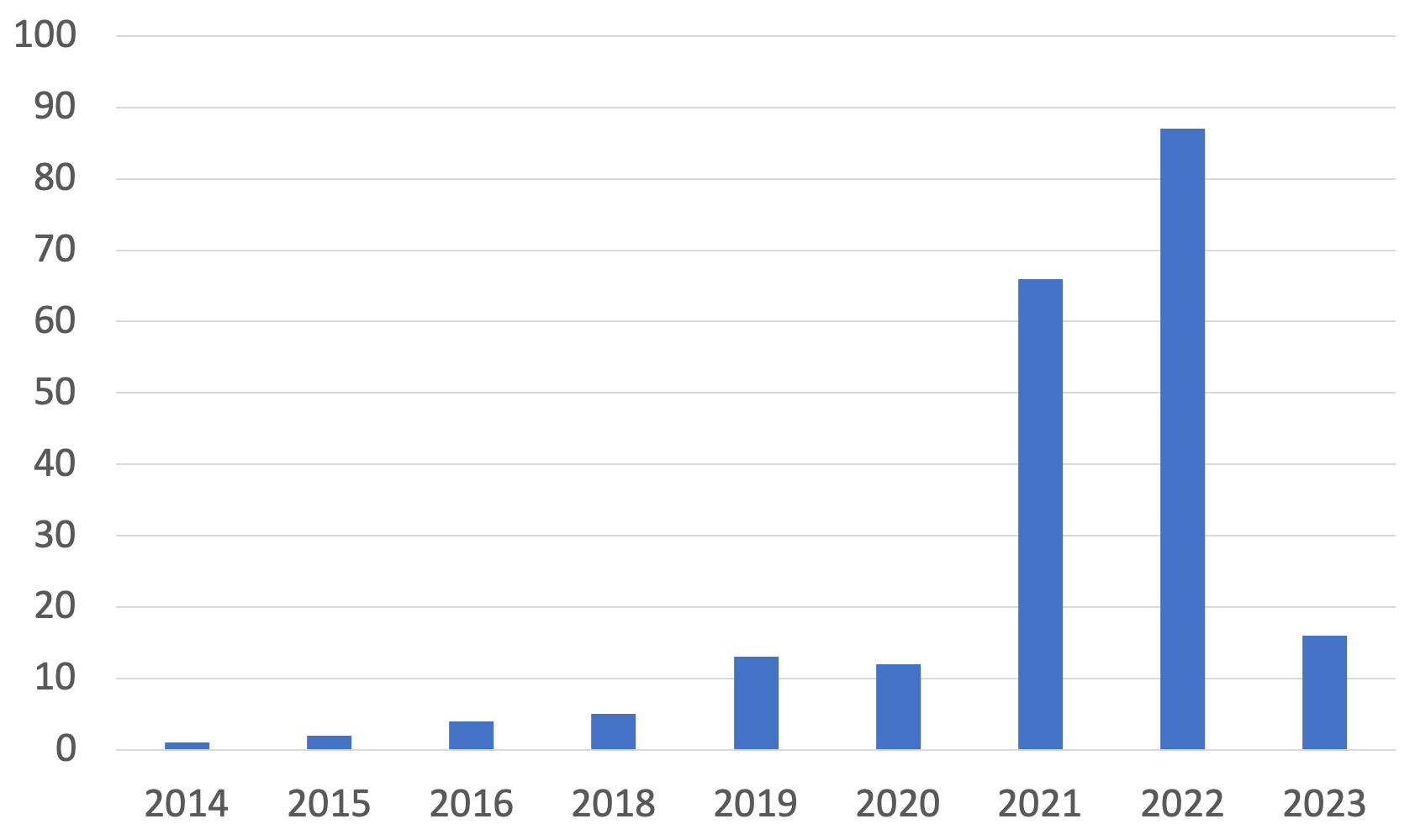}}
  \caption{Statistics of papers on public benchmarks. The data is collected from PapersWithCode\footref{fn:paperswithcode}}
  \label{fig:1_number_of_benchmarks_analysis} 
\end{figure}

In recent AI community, reporting results on publicly available datasets/benchmarks has became a widely accepted way to demonstrate the contribution of a work. To analyze the research trend on public benchmarks, we collected data from PepersWithCode\footnote{\url{https://paperswithcode.com/sota}\label{fn:paperswithcode}}. Up to Oct. 2023, there are over 11k benchmarks in total, and as shown from Figure \ref{fig:1_number_of_benchmarks}, over half of the tasks has more than 100 benchmarks available. As an example, we counted the number of papers per year (2014-2023) reporting results on ADE20K, an image segmentation benchmark. As shown from Figure \ref{fig:1_number_of_papers_for_ade20k}, starting from year 2021, there is a significant increase of papers targeting this benchmark. To find out the exact metrics that most researches are targeting, we sampled 4 tasks of CV and NLP and for each task we chose 5 common used datasets/benchmarks. For every benchmark, we counted the number of models that report metrics related to accuracy (e.g. IoU, Accuracy, F1, EM, BLEU score, etc.) or efficiency (e.g. GFLOPs, Time used, number of parameters, etc.). As the result shown from Table \ref{tab:number_of_papers_accuracy_efficiency}, Over 80\% of the papers are reporting metrics related to accuracy instead of efficiency. And there are 10/20 benchmarks have zero efficiency metrics reported. This result indicates that common interests of research community are targeting measures of performance like accuracy, and efficiency based performance like execution time, model size, etc. are ignored. 

\begin{table}[!htb]
  \centering
  \caption{Number of Papers Targeting Accuracy virsus Efficiency for Different Benchmarks}
   \begin{tabularx}{\textwidth}{llXXX}
   \toprule
   \textbf{Category} & \textbf{Task} & \textbf{Benchmark} & \textbf{Number of Papers Reporting Accuracy \newline(e.g. mIoU, Accuracy, F1, BLEU score, etc.)} & \textbf{Number of Papers Reporting Efficiency \newline(e.g. GFLOPs, Time, Number of Parameters)} \\
   \midrule
   \multirow{10}[20]{*}{\centering CV} & \multirow{5}[10]{*}{\centering Semantic Segmentation} & ADE20K & 206  & 6 \\
 \cmidrule{3-5}     &    & Cityscapes test & 101  & 3 \\
 \cmidrule{3-5}     &    & NYU Depth v2 & 99  & 0 \\
 \cmidrule{3-5}     &    & PASCAL Context & 62  & 0 \\
 \cmidrule{3-5}     &    & S3DIS & 49  & 4 \\
 \cmidrule{2-5}     & \multirow{5}[10]{*}{\centering Image Classification} & ImageNet & 931  & 480 \\
 \cmidrule{3-5}     &    & CIFAR-10 & 236  & 74 \\
 \cmidrule{3-5}     &    & CIFAR-100 & 193  & 20 \\
 \cmidrule{3-5}     &    & STL-10 & 119  & 9 \\
 \cmidrule{3-5}     &    & MNIST & 91  & 5 \\
   \midrule
   \multirow{10}[20]{*}{\centering NLP} & \multirow{5}[10]{*}{\centering Question Answering} & SQuAD1.1 & 211  & 0 \\
 \cmidrule{3-5}     &    & BoolQ & 42  & 0 \\
 \cmidrule{3-5}     &    & TriviaQA & 38  & 0 \\
 \cmidrule{3-5}     &    & Natural Questions & 36  & 0 \\
 \cmidrule{3-5}     &    & PIQA & 30  & 0 \\
 \cmidrule{2-5}     & \multirow{5}[10]{*}{\centering Machine Translation} & WMT2014 English-German & 90  & 13 \\
 \cmidrule{3-5}     &    & WMT2014 English-French & 55  & 0 \\
 \cmidrule{3-5}     &    & IWSLT2014 German-English & 32  & 2 \\
 \cmidrule{3-5}     &    & ACES & 21  & 0 \\
 \cmidrule{3-5}     &    & WMT2016 English-Romanian & 20  & 0 \\
   \bottomrule
   \end{tabularx}%
  \label{tab:number_of_papers_accuracy_efficiency}%
\end{table}%

\textit{\numberedParagraph{The model size and complexity are keeping increasing}} 

\begin{figure}[ht]
	\centering
	\includegraphics[width=0.75\textwidth]{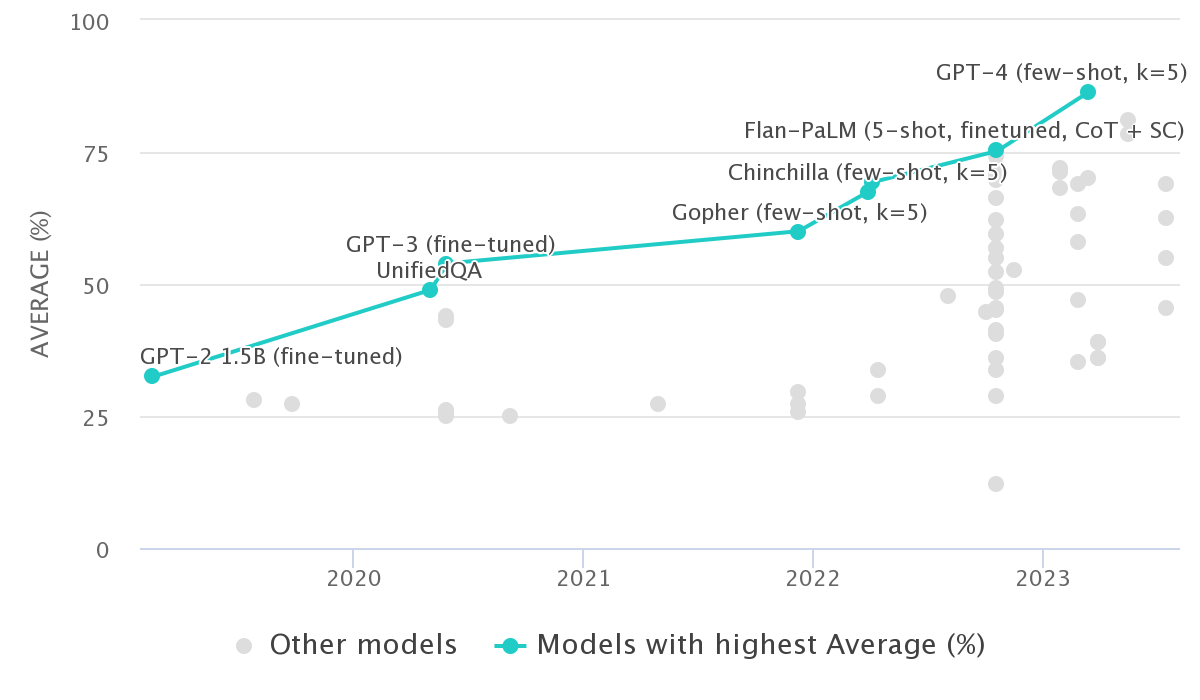}
	\caption{Leaderboard of Performance on MMLU(Multi-task Language Understanding), the chart is from PapersWithCode\footref{fn:paperswithcode_mmlu}}
	\label{fig:1_mmlu_charts}
\end{figure}

In recent years, there has been a rapid increase in the size and complexity of models, driven by advancements in hardware and computational power. As mentioned in the previous trend, most research efforts focus on pushing the boundaries of performance (such as accuracy) on public benchmarks, especially achieving new state-of-the-art (SOTA) results. And since increasing model size and complexity is a simple yet promising way to improve the model performance, researchers tend to use this approach to achieve better score on public benchmarks. We took MMLU\footnote{\url{https://paperswithcode.com/sota/multi-task-language-understanding-on-mmlu}\label{fn:paperswithcode_mmlu}}, a multi-task language understanding benchmark of NLP, as an example. As shown from Figure \ref{fig:1_mmlu_charts}, GPT-3(175B) achieved 1.5x scores to GPT-2(1.5B), and Flan-PaLM(540B) achieved 2.3x scores. However, if we consider the number of parameters, Flan-PaLM(540B) has 360x parameters to GPT-2(1.5B).

\begin{figure}[htbp]
	\centering
	\includegraphics[width=0.75\textwidth]{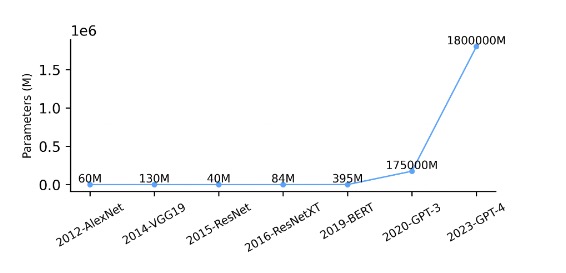}
	\caption{The Rapid Growth of Model Size}
	\label{fig:1_growth_model_size}
\end{figure}

While these larger models have shown promising results in terms of accuracy, their energy and carbon footprint have also grown at an exponential rate. This is an important consideration for AI practitioners. For example, the model AlexNet, developed by OpenAI, had 60 million parameters, but now, a large-scale text generation model called GPT-4 has 1.8 trillion parameters, a 30,000-fold increase in just 6 years. Figure \ref{fig:1_growth_model_size} illustrates the rapid growth in model parameters from 2012 to 2023. The latest version of Generative Pre-trained Transformers GPT-4 with 1.8 trillion parameters, can emit between 12,456 and 14,994 metric tons CO2e if it was trained on normal grid electricity in California, according to Kasper Groes Albin Ludvigsen(2023), while GPT-3 with 175B parameters can emit almost 500M carbons.

Based on \cite{Schwartz_Dodge_Smith_Etzioni_2020}, we categorize the aforementioned trends as \textcolor{red}{\textbf{Red Computing}}, where researchers prioritize enhancing accuracy (or similar metrics) on benchmarks by utilizing extensive computational resources, often at the expense of disregarding cost considerations, essentially "purchasing" higher performance, or even new state-of-the-art(SOTA) on leaderboards. However, this approach has several drawbacks, one of which is the ever increasing of energy comsumption and carbon emissions. As shown from \cite{Schwartz_Dodge_Smith_Etzioni_2020}, a diminishing return of model performance is obtained if we keep increasing the model complexity like number of parameters. And as we entered the stage of Large Language Models(LLM), the model size starts to increase ever faster. According to the statistics of \cite{Luccioni_Viguier_Ligozat_2022}, the training of BLOOM, a 176 billion parameter language model used 1,082,990 hours of total GPU training time, used 433,196 kWh energy and emitted approximately 50.5 tonnes of $CO_2eq$. And approximate 19 kgs of $CO_2eq$ is emitted per day of BLOOM's API deployment on inference time. Due to the fact that most large models like Llama2\cite{touvron2023llama}, PaLM\cite{chung2022scaling} and BLOOM\cite{scao2022bloom} require NVIDIA A100 or equivalent GPU with high computational power for training, small and medium-sized research institutions or companies often face difficulties in participating in research due to limited funding or computing power, which hinders the development of the entire community.

\subsection{Green Computing}

The approach of \textcolor{red}{\textbf{Red Computing}} has achieved significant advancements by pushing the boundaries of AI, but it also poses a threat to environment and natural resources. According to Gartner's Research Report\footnote{\url{https://www.gartner.com/en/articles/keep-ai-from-doing-more-climate-harm-than-good}}, AI is already consuming about 2\% of the electricity usage for the whole country. Since making AI more environmental friendly is a key goal, here we refer to \cite{Schwartz_Dodge_Smith_Etzioni_2020} and use the term \textcolor{green}{\textbf{Green Computing}}, which refers to researches that try to balance the performance of AI solutions and the cost of computational resources and environmental impacts. 

\begin{figure}[htb]
	\centering
	\includegraphics[width=0.75\textwidth]{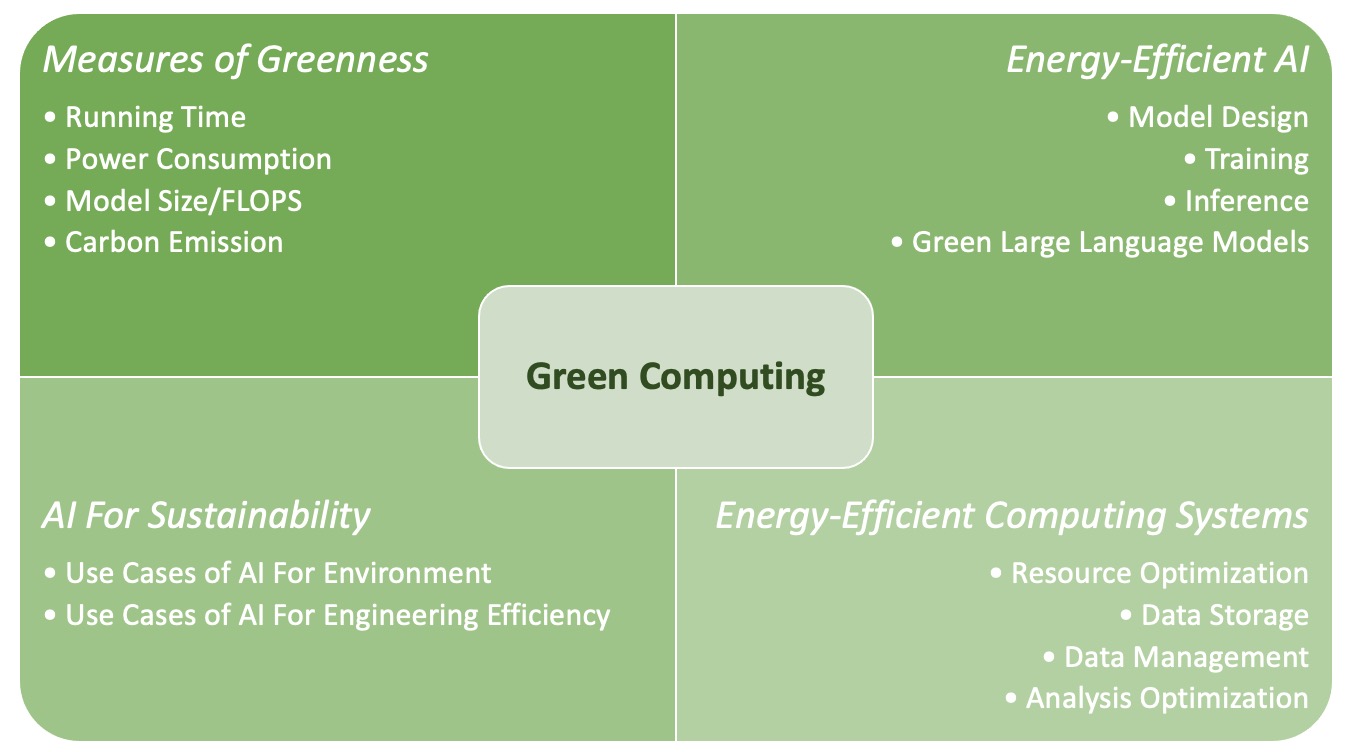}
	\caption{Green Computing Framework}
	\label{fig:1_green_computing}
\end{figure}

As shown from Figure \ref{fig:1_green_computing}, the framework of Green Computing contains below key components:

\begin{itemize}
  \item [(1)]
  \textbf{Measures of Greenness}: Key factors and methods to measure the computational resources needed for a intelligent system, or the "greenness" in computing. Common measurements include direct metrics like running time, power consumption (like electricity usage) and model size, also include indirect metrics like carbon emission.
  \item [(2)]
  \textbf{Energy-Efficient AI}: Energy-efficient methods to optimize the whole lifecycle of AI models, including model design, training, inference. It also includes optimization techniques for large language models to reduce power consumption for training and inference.
  \item [(3)]
  \textbf{Energy-Efficient Computing Systems}: Techniques to optimize the resources consumption in computing systems, including cluster resource scheduling, partitioning and data management optimization.
  \item [(4)]
  \textbf{AI for Sustainability}: Use cases to adopt AI for improving sustainability, including applications for environmental benefits(Green Computing For Environment) and improving engineering efficiency(Green Computing For Engineering). The Green Computing For Enviornment includes use cases like leveraging satellite imaging CV to monitor air pollution emission and carbon sequestration estimation, and the Green Computing For Engineering includes us cases like optimized cryptography for database security.
\end{itemize}

As opposed to Red Computing, the adoption of Green Computing has several opportunities or benefits. Firstly, the reduced consumption of computing resources introduces lower cost for real AI applications like autonomous driving, and the lower requirements for computing power can also introduce more opportunities for edge computing scenarios (e.g. mobile computing and IoT use cases). Secondly, green computing also reduces environmental impact and carbon emission. Finally, Green Computing also improves research equality by reducing the needs for computing power and can promote the overall development of AI community.

\subsection{Outline of the Survey}

In this survey, we give a systematic review of Green Computing based on the framework illustrated on Figure \ref{fig:1_green_computing}. The outline of this survey is organized as follows:

\begin{itemize}
  \item \textbf{Introduction}: This chapter gives a overview of the current Research and Development Trends of AI, also noted as Red Computing. it also discusses the motivation for Green Computing, the framework and opportunities for adopting Green Computing technologies.
  \item \textbf{Measures of Green Computing}: We describe the key factors that affect the computing resources consumption and common used ways to measure the "greenness" of computing.
  \item \textbf{Energy-Efficient Model Design}: In this chapter we list the energy-efficient modules to design an efficient AI model, and we also describe the strategy and NAS(Neural Architecture Search) methods to continue optimizing the model.
  \item \textbf{Energy-Efficient Training}: This chapter lists methods for optimizing the consumption of computational resources and data usage during model training, all while maintaining the performance of the trained model.
  \item \textbf{Energy-Efficient Inference}: This chapter describes the common used techniques to optimize a trained model, including model pruning, low-rank factorization, quantization, distillation and early-exit strategies. 
  \item \textbf{Green Computing Systems}: In this chapter we describe the techniques used to optimize the resource consumption in deployment environment, including cluster resource scheduling in cloud environment, server resource partitioning and data management.
  \item \textbf{Green Large Language Models}: In the age of Large Language Models(LLM), the requirements for computational resources have been increasingly high. This chapter lists novel approaches to optimize both training and inference of LLMs. 
  \item \textbf{Applications of Green Computing}: This chapter refers to AI for Sustainability. In this chapter we list several uses cases to adopt AI for environmental and engineering benefits across industries.
  \item \textbf{Conclusion}: This chapter concludes the survey and discusses possible future directions for Green Computing.
\end{itemize}

\section{Measures of Green Computing}
\subsection{Key Factors}
From many AI algorithm training and inference cases, we consider model size, parameter tuning and training data are the three major factors that affects computational resources.

\paragraph{Model Size} Model size refers to the number of parameters contained in a model. Larger model sizes usually require more computational resources and energy for training and inference, resulting in higher energy consumption and carbon emissions. Conversely, smaller model sizes can reduce the demand for computational resources and energy, thereby reducing the environmental impact of the model.

\paragraph{Parameter Tuning}
A series of experiments and evaluation steps conducted to validate and optimize model performance. During this process, researchers design and implement various algorithms, model architectures, feature engineering techniques, and training methods in order to achieve more accurate and efficient models. This model exploration process is computationally-intensive. In Facebook's ML research cluster, approximately half (p50) of the ML training experiments require a maximum of 1.5 GPU days, while almost all (p99) of the experiments are completed within 24 GPU days \cite{wu2022sustainable}. During the training process, researchers require adopt additional hyper-parameter tuning to optimize the algorithm. Based on experience, at least 50\% of experiments are conducted to select optimal experimental parameters.

\paragraph*{Training Data}
Last but not the least, the training data is also an important factor influencing the environmental impact of a model. The scale and quality of the training data directly affect the model's performance and training effectiveness. Typically, larger training datasets require more storage space and computational resources for processing, resulting in increased energy consumption and carbon emissions. Additionally, low-quality training data may lead to over-fitting during the training process, requiring more training iterations to adjust the model parameters, further increasing energy consumption. Therefore, selecting appropriate training datasets and optimizing the data processing process can reduce the environmental impact of the model.

As the same with model size, the training data increased rapidly. In the NLP field, BERT\cite{devlin2019bert} pre-trained a Transformer encoder on 3 billion word pieces. GPT-3, a powerful generated model, pre-trained in 2022 on 45TB data. Compared to the previous models, these massive training examples largely increase the training computation costs.

\subsection{Measurements}
In this section, we describe the common ways to measure "greenness" in computing.
\subsubsection{Running Time}
Running Time refers to the total time of model training and inference. It is easy to collect by just add timer to the program. When all models adopt the same infrastructure including hardware and software, compare running time is an effective approach. But in reality, adopting the same infrastructure is difficult, and representing "greenness" solely based on runtime lacks accuracy and hard to compare horizontally
\subsubsection{Model Size (e.g. Number of Parameters)}
As mentioned earlier, model size has a significant impact on "greenness". So model size can partly represent the algorithm's "greenness" while it can not reflect the impact of factors such as data volume, number of training iterations, and other aspects of the model training and inference process on "greenness".
\subsubsection{FPO/FLOPS (Floating Point Operations)}
Floating-Point Operations (FLOPs) measure the number of operations required to execute a specific instance when running a model. FLOPs, being hardware and software agnostic, are a straightforward measure that enables fair comparisons between different models and serves as a metric for assessing computational efficiency. Nevertheless, FLOPs serve as theoretical metrics and do not provide an accurate representation of actual runtime due to varying degrees of parallelism (DOP) across different algorithms.
\subsubsection{Hardware Power Consumption}
The most common negative impacts of AI on the environment include increases hardware power Consumption. Many hardware manufacturers provide interfaces to obtain machine-level energy consumption. However, machine-level energy consumption is usually much larger than the model actually use.
\subsubsection{Energy Consumption}
Energy consumption refers to the overall energy usage of a facility, which should not be confused with the energy delivered to the hardware\cite{parcollet2021energy}. The difference can be defined as below:
\begin{equation}
  e_{\mathrm{total}}=\text{PUE}*e_{hardware}
\end{equation}
where $PUE$ means Power Usage Effectiveness, which represent the compensatory energy consumption, such as that used for cooling purposes. According to the 2020 Data Center Industry Survey Results, the average Power Usage Effectiveness (PUE) worldwide was reported as 1.59\cite{uptimeInstitue2019}. It is worth noting that the actual PUE value is heavily influenced by the specific computing infrastructure in use. For example, Google boasts a remarkable trailing twelve-month PUE ratio of 1.11\cite{google2020efficiency}, as stated in their efficiency report for the year 2020. In comparison, Amazon (AWS) and Aliyun reported PUE ratios of 1.2\cite{aws2020aws} and 1.3\cite{aliyun2019aliyun} respectively in their respective reports.

\subsubsection{Carbon Emission}
\begin{figure}[htb]
	\centering
	\includegraphics[width=0.75\textwidth]{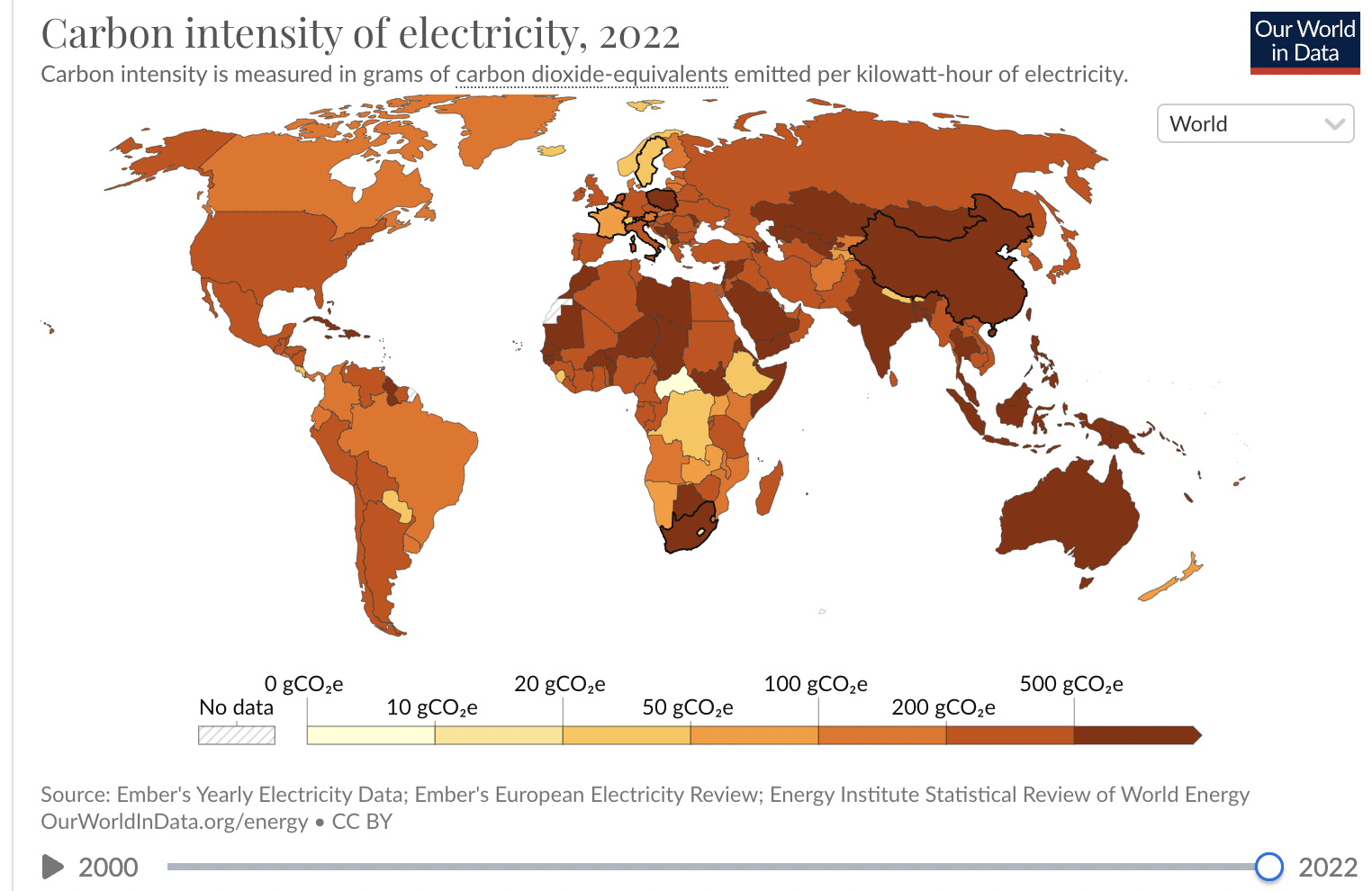}
	\caption{Different area has different carbon-intensity}
	\label{fig:2_carbonintensity}
\end{figure}

Assessing carbon emissions is the most direct method for evaluating the environmental impact. As it is hard to directly collect the statistics for carbon emission, We usually estimate it by energy consumption and carbon intensity.
\begin{equation}
  \text{CE} =\text{CI}*e_{\mathrm{total}}
\end{equation}
Quantifying this precisely is a challenging task due to variations in carbon intensity across different energy grids, which refers to the quantity of carbon emitted for every kilowatt-hour of energy consumed. The carbon intensity of electricity generation differs across regions and relies on the energy sources utilized to generate power for the local electrical grid. We get the carbon intensity of electricity for year 2022\footnote{\url{https://ourworldindata.org/grapher/carbon-intensity-electricity}} as shown from Figure \ref{fig:2_carbonintensity}.

To make the carbon costs of model training transparent, we encourage more investigators to measure energy usage and $CO_2$.
\subsection{Tool-kits}
In the past few years, tools for tracking "greenness" measurements have been poured significant attention into as a major topic in this field. 
\begin{itemize}
\item \textbf{tfprof}:
Tfprof is a profiling tool in TensorFlow that helps you analyze the performance of your TensorFlow models. It uses the static tf graph to calculate the operates.However, tfprof does not count FLOPs for operators with unknown shapes, so it is necessary to calculate the shape of tensors before statistics. In the absence of additional information, a common practice is to set the unknown dimension of the multidimensional matrix (tensor) to the batch size. This operation has a small error in the CV scene, but it will lead to significant errors in the recommendation scene. 
\item \textbf{Green Algorithms}:
Lannelongue et al. introduced Green Algorithms\cite{lannelongue2021green}, a methodological framework outlined in their study, which provides a standardized and reliable approach to estimating the carbon footprint of computational tasks. This framework takes into account factors such as processing time, computing core types, available memory, and the location and efficiency of the computing facility. The Green Algorithms tool employs algorithmic calculations to estimate carbon footprints, making it compatible with computational processes without the need for extensive information or code modifications. It accommodates a wide range of computing systems, including CPU, GPU, desktop computers, local servers and cloud computing. However, it's important to note that the estimated solutions may not fully capture the complexities of real-world operating environments, such as fluctuating machine workloads, which can affect the accuracy of the metrics.

\item \textbf{CodeCarbon}: Alexandre Lacoste et al.\cite{mlco2} are currently developing a Python package named "CodeCarbon" that enables the tracking of carbon emissions generated by different types of computer programs. This package covers a wide range of applications, from simple algorithms to complex deep neural networks. This package is user-friendly and easily visualize the track result. However it gather the machine's level energy emission which is much bigger than our special program.

\item \textbf{Carbontracker}: "Carbontracker"\cite{anthony2020carbontracker} is a similar python packages like CodeCarbon, which tracks and predicts energy consumption and carbon emissions for training deep learning models. This provides a more accurate estimation as it can track Single-GPU card emissions rather than machine level energy.

\item \textbf{Automatic AI Model Greenness Track Toolkit}: "Automatic AI Model Greenness Track Toolkit"\footnote{\url{https://github.com/alipay/Automatic_AI_Model_Greenness_Track_Toolkit}}  is a Python package developed by Ant Group. It enables the measurement of the greenness of AI processes through tracking FLOPs, electricity usage, and carbon emissions. This user-friendly toolkit requires minimal code and simple configuration to quantify AI greenness without disrupting the existing code structure. It provides a more accurate estimation of process-level energy consumption and employs a runtime statistical approach for precise FLOPs calculation, instead of relying on static analysis like tf.profiler. However, it currently only supports TensorFlow for FLOPs statistics.

\end{itemize}

\section{Energy-Efficient Model Design}

\subsection{Green Compact Module}

\subsubsection{Compact Convolution}

\textbf{\numberedParagraph{Depth-wise Separable Convolution}}

For image classification, target detection and other AI tasks, some classical neural network models of deep learning, such as LeNet, VGG, GooogleNet, etc., these have been a fairly good response using these models, but limitations of these models is very obvious, have lots of arguments, and require redundant computation. It is hard to apply some practical situations in which light-weight devices are used, such as robots, autonomous vehicles, recognition tasks, etc., which need to be executed in a timely manner on a platform with limited computing. This is where lightweight networks emerge that can target these mobile scenarios.

\paragraph{MobileNet} \cite{howard2017mobilenets} proposes a kind of efficient models called MobileNets, is an efficient CNN(convolutional neural network)model. MobileNets is based on a streamlined architecture, and the lightweight DNN can use the depth-wise separable convolution to construct. MobileNets is built primarily from the deep separable convolution originally introduced by \cite{sifre2014rigid}. There are three versions of MobileNets, namely MobileNetV1, MobileNetV2 and MobileNetV3, all of which employ some lightweight techniques such as deepseparable convolution, inverted residual structure, linear bottleneck layer, etc., to decrease the complexity of arguments and the redundancy of computation in the model while maintaining a high accuracy rate.

\begin{itemize}
 \item MobileNetV1: The first version of MobileNets was introduced in April 2017. Its core idea is finding a substitution of the standard convolution, thereby reducing the complex arguments and the redundant computation involved. The convolution is a separate convolution of each access of the input, and the point-wise convolution is a combination of 1*1 convolution checks against the output of a deep convolution. This allows efficient extraction of local and global features while reducing computational complexity. The standard convolutional filter is replaced by two layers and a depth-separable convolutional filter is constructed with point-by-point convolution. The standard convolution step both filtrates and composes the inputs into a new set of results. Depth-separable convolution disassembles it into a part for filtrating, and a part for composing.

 \item MobileNetV2: The second version builds on MobileNetV1 by introducing an inverted residual structure and a linear bottleneck layer to better promote the efficiency and performance of the structure. Inverted residual structure refers to switching the bottleneck layer and the extended layer in the standard residual structure. This avoids non-linear transformations in low-dimensional Spaces and preserves more information.

 \item MobileNetV3: This is the third version of MobileNets, which uses neural architecture search and network structure optimization methods again to better the efficiency and energy of the model. Neural architecture search refers to the use of automated methods to search for optimal network structures, rather than manual design. Network structure optimization refers to some fine-tuning and improvement of the network structure based on neural architecture search to adapt to different tasks and scenarios.

\end{itemize}

\paragraph{Xception} Both Xception and MobileNets are deep learning models based on depthwise separable convolution. Both of them can be used for learning tasks with high efficiency and performance.However, from the perspective of Inception module, Xception completely decomparts the convolution operation in Inception module into two convolutions, thereby reducing the complexity of arguments and the redundancy of computation, while improving the effect of feature extraction. Inception module is a typical network structure in DL, which is characterized by combining convolution cores of different sizes layers and extract information of different scales at once, finally splicing all the outputs together to form a deeper and wider feature map. The Inception module is designed to reduce the field of acceptance and multi-scale features of the layer while maintaining the spatial information, meanwhile the complexity of arguments and redundancy of computation.

Xception is the deep learning model based on Depth-wise separable convolution. Xception is Extreme Inception, meaning that it is an extreme Inception model. Standard convolution is a convolution kernel that handles all channels, called a single-segment case. Deep separable convolution is one convolution kernel dealing with one channel. This is called one segment per channel. Inception is in an intermediate state, i.e. not one convolution kernel handles all channels, nor an "extreme" one convolution kernel handles one channel. The Inception module is a structure that divides the input feature graph into several channels with different receptive field sizes, then performs a separate convolution operation on each channel, and finally concatenates the outputs of all channels. This can increase the receptive field and multi-scale features of the network while maintaining the spatial information, and the complexity of arguments and redundancy of computation. Xception also combines a residual connection with a linear bottleneck layer to enhance the stability and expressiveness of the network.

Xception can increase network efficiency, as well as being superior to Inception-V3 \cite{chollet2017xception}. It can run large datasets with complex arguments. It sparked some ideas: in the case of given hardware resources, as far as possible to increase the network efficiency and performance, can also be understood as the full use of hardware resources. As you can see, the Xception unit can greatly reduce the number of parameters. Xception also uses residual connections and linear bottleneck layers to enhance the stability and expressiveness of the network. A residual connection is the addition of a jump connection between input and output to better convey gradients and information during training. The linear bottleneck layer refers to the removal of the ReLU activation function behind the last point-by-point convolution layer in favor of a linear activation function or no activation function. In this way, the information loss caused by ReLU can be reduced and the feature expression ability can be improved.

\textbf{\numberedParagraph{Fire Convolution}}

Fire Convolution was proposed by SqueezeNet, a lightweight neural network model designed for mobile and embedded vision applications. A convolutional structure designed for lightweight neural networks is full name Flexible, Inexpensive, Reusable, and Efficient Convolution, meaning that it is a flexible, low-cost, reusable, and efficient convolutional structure.

Fire Convolution and SqueezeNet are closely related concepts. Both of them are convolutional structures designed for lightweight neural networks. They can be used for image classification, target detection, image generation and other tasks with high efficiency and performance at the same time. However, there are some differences between Fire Convolution and SqueezeNet. Fire Convolution is a convolutional structure that composed of squeeze part and expand part. The squeeze part compresses the input with 1x1 convolution checks; the expand part can expand output of the squeeze part with 1*1 and 3*3 convolution checks to increase channel amount.

Fire Convolution greatly reduces the complexity of arguments and footprint of access meanwhile keeping network performance. SqueezeNet is a neural network model that uses Fire Convolution instead of normal convolution in order to reduce parameters and memory footprint. SqueezeNet also uses several other strategies to reduce the model size, such as reducing the amount of channels to 64, delay downsampling, and replacing the full connection layer with global mean pooling. Fire Convolution is a core component of SqueezeNet, but it's not the only one. SqueezeNet also contains several other ordinary convolution layers and pooling layers, as well as a fully connected layer. Fire Convolution can be used by other neural network models, not just SqueezeNet. SqueezeNet aims to compress the model size to less than 0.5M while maintaining AlexNet level accuracy. To achieve this, SqueezeNet uses Fire Convolution to replace normal convolution to decrease the number of arguments and address footprint.

\textbf{\numberedParagraph{Flattened Convolution}}

Flattened Convolution is a convolutional structure designed for lightweight neural networks. The full name of Flattened Convolution is Flattened Convolutional Layer, meaning that it is a structure that fuses the convolutional and fully connected layers together. The characteristic of Flattened Convolution is that it connects each channel of the feature distribution of input information to the fully connected part separately, and then splices the outputs of all channels together to form a deeper and wider feature map. The conventional 3D convolution filter is achieved in the training phase by dividing it into three successive 1D filters. The purpose of Flattened Convolution is to augment the field of acceptance and multi-scale characteristic and meanwhile keeping the information of the network, while reducing the complexity of arguments and the redundant computation.

\cite{jin2014flattened} A Flattened Convolutional neural network aimed for fast feedforward acceleration is proposed. The excessiveness of arguments, particularly the data complexity of the convolutional filters in CNNs, has been wide designed, and diverse methods which to build the low-rank part of the trained filters are need to put forward. From this perspective, the paper trains flattened networks, which includes a series of one-dimensional part, to gain performance advantage to that of traditional convolutional networks. They tested flat models and found flat layers can effectually replace 3D filters with high accuracy. Because the learning parameters are significantly reduced, the flat convolutional pipe provides approximately twice the acceleration during the feedforward pass.

\textbf{\numberedParagraph{Shrinked Convolution}}

Shrinked Convolution is a convolutional structure designed for lightweight neural networks. The full name of Shrinked Convolution is Shrinked Convolutional Layer, meaning that it is a structure that blends the convolutional and pooled layers together. The feature of Shrinked Convolution is that it can connect each channel of the input feature graph to a convolutional kernel separately, and then average pool the outputs of all channels to form a smaller and more compact feature graph. Shrinked Convolution's purpose is to reduce the receptive field and multi-scale features of the network while maintaining its spatial information, while reducing the complex arguments and the redundant computation involved.

\subsubsection{Compact Graph Computation}

Graph representation is everywhere; Objects in the real world are often defined by how they are connected with others. The cluster of objects, and the their relationship, can reasonably represented as a data structure graph. For more than a decade, researchers have been developing neural networks (called graph neural networks or GNNS) based on graph data. Graphs are a very powerful and versatile representation of data, and there are actually some kinds of element can be construct into graphs,such as images or text.

\textbf{\numberedParagraph{ImprovedGCN}}

Motivated by numerous research finds, \cite{dhawan2022improvedgcn} proposes the ImprovedGCN, which contains the main necessary component in GCN. This is CF, CF is an existing method for building robust recommendation systems. A common sample is to regard the item and the user as an embed, then to understand the diverse limitations of the embed between these. After that, the information by embeddings can connected with the sum to obtain the final embeddings for prediction. The above framework is simple and unsophisticated, and not only is it easy to train, but it also has good validity.

The ImprovedGCN model is a recommendation system based on a lightweight graph convolutional network, which basically contain the following three portions: (1) The construction of a user-item dichotomy graph, based on the user's historical behavior and social relationship, to construct a user-item dichotomy graph, which is used to show the relations between users and items; (2) Lightweight graph convolutional network learning, which uses lightweight graph convolutional layer to learn the representation vector of users and items on the bipartite graph, while introducing attention mechanism to enhance the similarity calculation between users and items; (3) Recommendation result generation, according to the representation vector of the user and the item, calculate the user's preference score for the item, and rank according to the score to generate the recommendation result. The ImprovedGCN model can effectively utilize complex relationships between users and items and lightweight graph convolution layers to improve recommendation accuracy and variety.The lightweight graph convolution layer is a method used to decrease the complex arguments and the redundant computation in GNN, which mainly consists of the following two steps: using sparse matrix multiplication to implement the graph convolution operation, avoiding dense processing of the adjacency matrix, thus reducing memory consumption and computational complexity; The adjacency matrix compression is used to reduce the amount of neighbor nodes of the graph convolution, and the input dimension and output dimension of the graph convolution are reduced by hash coding and bucket splitting of the adjacency matrix. The lightweight graph convolution layer can better the speed and efficiency of the GNN, and is suitable for large-scale recommendation scenarios.

\textbf{\numberedParagraph{SeHGNN}}

Heterogeneous characteristic means more complete and richer semantic information, requiring specially customized models to handle them. Heterogeneous Graph Neural Network (HGNN) is a wonderful method to embed those graph information into vetcor. Most of the existing HGNN methods inherit the graph neural network mechanism designed for isomorphic graphs, such as the attention mechanism and multi-layer structure. These mechanisms lead to unnecessary complexity. They have conducted extensive research on this and proposed a straightforward and effective network :SeHGNN.

To get structural information more easily, SeHGNN uses a lightweight average aggregator to aggregate precomputed neighbor representations, thereby reducing complexity, eliminating excessive attention on neighbors, and avoiding reduplicative neighbor vertex aggregation in each training cycle. For utilizing semantic information, SeHGNN uses a single-layer idea with long-distance meta-paths which can expand the forward domain, and a transformer-based semantic fusion part to mix features of different meta-paths. Therefore, SeHGNN can achieve high precision and fast learning speed with simple structure characteristics.

SeHGNN uses two innovative designs: One is to use a lightweight mean aggregator to anticipate neighbor aggregations, avoiding the overuse of neighbor attention and the overhead of repeating neighbor aggregations every training period; The second is the use of a single-layer module and long-distance meta-paths which can expand the forward domain, besides the transformer-based semantic fusion part to mix features of different meta-paths is also a good design.

\subsubsection{Time-Series Model}

Time series classification has important applications in many fields, such as health, industrial automation, network services and network security. Most of the most advanced classification methods rely on ensemble learning, which uses multiple basic models to make a classification. The advantage of ensemble learning is the ability to synthesize the wisdom of different models to improve classification accuracy. However, ensemble learning also requires a lot of computational resources, making them hard to utilize that to put into use resource-constrained devices.

\textbf{\numberedParagraph{LightTS}}

As the process digitization develop to maturation, a new-style time-series structured data will be generated. The accuracy of modern classification is usually based on communities centered around some basic models. High accuracy learning requires a large amount of computing power, which is impossible in environments with limited resources such as edge equipment. In order to raise adaptability of learning,\cite{campos2023lightts} proposes a simple framework that integrates large-scale inclusion into light-weight moduels meanwhile providing accuracy for higher competitive. Firstly, it propose lighting models that encourage the fusion of different basic models with different weight factor; Second, this method allows users with limited resource budget to use the cheapest model based on the model for highly accuracy and greater size.

The goal of the LightTS framework is to break up large integration models for lightweight resources and maintain highly accuracy. It consists of two main parts: Adaptive integrated distillation: This part is responsible for converting the classification results of multiple base models into training targets for lightweight models. It uses an adaptive weight allocation method,which portions different indexs based on the ability to classify on different classes, allowing them to participate meaningfully in the training of lightweight models. This has the advantage of balancing the contributions of different base models, avoiding overfitting or underfitting, and improving the generalization ability of lightweight models.

Pareto optimality selection: This section is responsible for selecting the best model from multiple lightweight models after the training is complete. It uses a method to identify the Pareto optimal setting between model accuracy and model size, allowing the user to choose the most accurate lightweight model with a space budget. The benefit of this is the flexibility to adjust the size and performance of the lightweight model according to different resource constraints.

LightTS is a new flexible architecture aimed at expanding the advanced time-series sort to resource constrained machines. Firstly, it can convert large baseline models into small models. In this case, two optimization levels are used to evaluate the effectiveness of each basic model processing process. Secondly, the method of finding the pareto optimal parts solution by using simple machine with different spatial request.

\textbf{\numberedParagraph{LightCTS}}

Correlation time series (CTS) haves an important role in various applications such as traffic and server control. Several DL models have been proposed to improve the accuracy of CTS predictions. However, as models become more and more inclusive, they hard to boost precision. \cite{lai2023lightcts} The goal of the research is to achieve better results, lighter designs that maintain precision and be able to use non-functional materials. For that, a popular CTS prediction model has emerged and two findings have been shown to predict. With this, the LightCTS architecture is proposed that uses a simple superposition of time and space wiyhout using more intensive devices. In addition, LightCTS has lightweight spatio-temporal operator parts, called L-TCN and GL-Former, which improve the performance of the calculation without compromising its abstract function. It as well includes a final compression model to reduce downtime and speed up subsequent calculations. Tests on data mining show that LightCTS can achieve high-fidelity graphics with less code and flash overhead.

Begin with a thorough review of the above models, put those in a unified architecture, and carefully study the calculation and storage costs of the models. The test result yielded key discoveries and indicated two ideas to get lightness: 1) deformalizing the calculations related to information abstruction and 2) enhaving the general CTS framework as well comcompression of the excessivebess time dimension for the expensive operator.

The LightCTS framework is a CTS prediction model based on neural network, which is composed of four main parts: 1) Embedded module: This module is responsible for converting the multi-dimensional CTS of the input into a high-dimensional vector representation for subsequent feature extraction. It uses two embedding methods: positional embedding and channel embedding. Location embedding is used to capture timing information in a time series, and channel embedding is used to capture correlations between different time series. 2) Spatiotemporal operator: This module is responsible for extracting spatiotemporal features from the vector representation obtained by the embedded module, i.e. features that take into account both time and spatial dimensions. It uses two lightweight space-time parts: L-TCN and GL-Former. L-TCN is a module based on causal convolution. GL-Former is a module on account of the self-attention mechanism, that is flexible in capturing spatial correlations between different time series. Both modules use deep separable convolution and multi-head attention mechanisms to reduce computational complexity and number of parameters. 3) End compression strategy: This strategy is used to reduce the redundant features of the spatiotemporal operator's output and speed up the subsequent calculation process.It uses an adaptive pooling method to optimise the size of the pooling window with the information of different time series, thus retaining more useful information while reducing useless information. 4) Aggregation and output module: This module is responsible for aggregating the compressed spatio-temporal features and outputting the predicted results.A key feature of the LightCTS framework is its approach to combining spatio- temporal operators in parallel stacks, rather than alternating stacks. This has the advantage of reducing the computational overhead while maintaining high accuracy.

\subsubsection{Transformer-Based Model}
Transformer is a powerful sequence model, but requires a quadratic increase in time and memory with sequence length. In theory, Self Attention's time complexity and memory usage are reach $O(n^2)$ level ($n$ is the sequence length), Self-Attention is $O(n^2)$ because it computes correlation for any two vectors to get the $n^2$ correlation matrix. However, the memory and computational requirements of such networks grow by the power of two with sequence length, which precludes their use on long sequences.

\textbf{\numberedParagraph{Efficient Attention}}

\paragraph{Sparse Attention} To save memory and speed up computation, a basic idea is to reduce associative computation, that is, to think that each element is only related to a part of the sequence, which is the basic principle of Sparse Attention. Sparse Attention is a way to use the sparsity of the attention matrix to reduce the computation and memory footprint. The basic idea of Sparse Attention is to compute and retain only a portion of the non-zero elements in the attention matrix, while ignoring or approximating the others. This reduces the complexity of the attention matrix to $O(n)$ or $O(nlogn)$, in which $n$ means sequence length. Sparse Attention can be implemented in a variety of ways, such as fixed sparse patterns (such as local attention, block attention, sparse multi-head attention, etc.), or dynamic sparse patterns (such as sparse attention based on hashing or gradients, etc.). \cite{child2019generating} separates a complete attention calculation into several faster attention operations that, when combined, can approximate intensive attention operations. Use it to apply self-attention to sequences of unprecedented length.

\paragraph{Attention Approximation} Attention Approximation is a method of approximating the value of attention by using the low-rank properties of the attention matrix. The basic idea of Attention Approximation is to use a low-dimensional feature mapping function (such as a kernel or random feature mapping) to project the query, key, and value matrix into a low-dimensional space, then perform a dot product attention calculation in the low- dimensional space, and finally map back to the original space. This reduces the rank of the attention matrix, thereby reducing the amount of computation and memory footprint.

Reformer and Performer are two kernel-based approximation methods that can both transform the dot product attention mechanism into a linear operation, thereby avoiding quadratic dependence on the length of the input sequence. Reformer is a method that uses Gaussian kernel functions to implement Attention Approximation that can transform the dot product attention mechanism into a linear operation, thereby avoiding quadratic dependence on the length of the input sequence. First, the input sequence is transformed linearly to get the query representation MathbfQ, the key representation MathbfK, and the value representation MathbfV. Then, Gaussian kernel functions are used to project the query matrix and key matrix into an infinite dimensional feature space.

Performer is a method of implementing Attention Approximation using orthogonal random feature mapping that transforms the dot product attention mechanism into a linear operation and maintains the same theoretical properties as the traditional Transformer model. Performe first performs a linear transformation on the input sequence to get MathbfQ, MathbfK, and MathbfV. Then, orthogonal random feature mapping function $p(cdot)$ is used to project the query matrix and key matrix into a low-dimensional space. The orthogonal random feature mapping function guarantees that the dot between projected query and key matrix is equal to the dot between the original representation matrix.

\textbf{\numberedParagraph{EdgeBERT}}

EdgeBERT is a model based on ALBERT that enables inference for multi-tasking natural language processing on chips with low power consumption. The goal of EdgeBERT is to adapt to edge computing scenarios with as little memory footprint and computational effort as possible while guaranteeing a certain accuracy rate. ALBERT is a lightweight variant of BERT that was proposed by Google in 2020 to reduce the complex arguments and expensive computation with the improvement for BERT. There are three main improvements in ALBERT: Embedding layer decomposition: Separating the dimensions in the word embedding and the hidden layer, decreasing the amount of arguments in embedding layer; Parameter sharing: all Transformer layer parameters are shared between layers to decrease the amount of arguments; for example,sentence order prediction task: replace the initial BERT's task for forecast subsequent statements with judging the order between two sentences to improve the learning effect of text consistency.

EdgeBERT, an ALBERT model optimized for edge computing, was proposed in 2021 by Harvard University and others to minimize energy consumption while meeting latency requirements. EdgeBERT's main optimization has two points: the entropy-based early exit mechanism: The number of Transformer layers is dynamically adjusted according to the complexity of each sentence. When the output entropy of a certain layer is lower than the threshold, the calculation of subsequent layers will be stopped to save operation time; Dynamic attention range: According to the different attention range of each head, a mask is set for each token, so that it can only calculate attention to the surrounding tokens, reducing the overhead of matrix operation.

\cite{tambe2021edgebert} EdgeBERT uses a number of algorithms and hardware techniques to compress model size and increase reasoning speed. The implementation of EdgeBERT mainly includes the following aspects: 1) Entropy-based early exit mechanism: The number of Transformer layers is dynamically adjusted according to the complexity of each sentence. When the output entropy of a certain layer is lower than the threshold, the calculation of subsequent layers will be stopped to save operation time. 2) Dynamic attention range: According to the different attention range of each head, a mask is set for each token, so that it can only calculate attention to the surrounding tokens, reducing the cost of matrix operation. 3) First-order network pruning: according to the change amplitude of each parameter in the fine-tuning process, some parameters are selectively zeroed, thus reducing the number of model parameters and memory occupation. 4) floating-point quantization: the 32-bit floating point number is converted to an 8-bit integer, thereby reducing the model storage space and computational complexity. 5) Hardware accelerator system: A dedicated hardware accelerator system is designed to achieve fast switching of voltage frequency and high-density embedding of non-volatile memory, thus reducing energy consumption and latency.

\textbf{\numberedParagraph{R2D2}}

The R2D2 model is a network based on Transformer Networks that can efficiently process long text sequences while maintaining efficient computational complexity. Its main features are: it uses a new Attention mechanism called Sparse Recomputable Attention, which can dynamically select a portion of key value pairs that are most relevant to each query, thus reducing the size and computation of the attention matrix; A new pruning and growing algorithm is used, which can greatly reduce the computation and memory overhead without affecting the tree structure and model performance, and realize the linear time coding; And a new pre-trained goal was used, which implemented a recursive Transformer based on differentiable Cky-style binary trees, attempting to predict each word in terms of its left and right abstract nodes.

The sparse repeatable attention mechanism based on the R2D2 model is a novel attention mechanism that can greatly reduce the computational and memory overhead without losing accuracy. Its main idea is to select some key key-value pairs from the original key-value pairs, called Anchors, and recalculate other key-value pairs according to the anchors and queries. Specifically, the attention mechanism consists of the following steps: First, the input sequence is divided into into multiple subsequences, and each subsequence only performs attention computation with itself and neighboring subsequences, thus achieving sparsity;Then, for each subsequence, a gradient-based pruning method is used to select some key key-value pairs from the original key-value pairs as anchor points and store them in the replay buffer, thus achieving repeatability; Finally, when attention needs to be calculated, other key value pairs are recalculated based on anchor points and queries, and a sparse matrix multiplication-based method is used to transform the attention calculation into an efficient matrix operation, thus achieving high efficiency.

Specifically, the R2D2 model includes the following parts: 1) Input encoder: It converts the input sequence into a matrix composed of word vectors and position vectors, which serves as the input to Transformer Networks. 2) Sparse repeatable attention: it is one of the core components of the R2D2 model, which makes use of Transformer Networks' self-attention machinery and multi-dimension attention machinery for boosting representation mean and generalization ability of model. First, the input sequence is divided into multiple subsequences, and each subsequenceperforms attention calculation only with itself and neighboring subsequences, thus achieving sparsity ; Then, for each subsequence, a gradient-based pruning method is used to select some key key-value pairs from the original key-value pairs as Anchors and store them in the replay buffer, thus achieving repeatability. Finally, when attention needs to be calculated, other key value pairs are recalculated according to anchor points and queries, and a method based on sparse matrix multiplication is used to transform the attention calculation into an efficient matrix operation, thus achieving high efficiency. 3) Pruning and growth algorithm: It is one of the core components of the R2D2 model, and it is used to optimize the computation and memory overhead of sparse repeatable attention mechanisms. 4) Output decoder: It converts the output vector obtained by the sparse repeatable attention mechanism into a matrix composed of word vectors and gets the final prediction result through a linear layer. 5) Pre-training goal: It is one of the core components of the R2D2 model, and it is used to implement a recursive Transformer that tries to predict each word based on its left and right abstract nodes. First, using a binary tree based on differentiable CKy-style, the input sequence is divided into multiple subtrees and an abstract node is assigned to each subtree; Then, using a bidirectional language model, predict each word based on its left and right abstract nodes and calculate the prediction loss; Finally, use an optimizer that updates the model parameters based on the predicted losses and adjusts the structure of the binary tree.

\subsubsection{Lightweight Softmax}

The softmax uses the vector $z$ of $N$ real numbers as parameter and normalizes it to the probability distribution of $N$, it’s easy to see probabilities are proportional to the exponent of the parameter number. In most situation, many vector elements may be negative, or greater than 1 before using softmax. But after applying softmax, each part will be in range $(0,1)$ and these parts add up to 1 so that they can be viewed as probabilities. Accordingly, input parameter values are in proportion to probabilities.

The usual softmax function $\sigma:R^N\rightarrow(0,1)^N,N\gg1$, the formula $\sigma$ is formulated as follows:

$$\sigma(Z)_i=\frac{e^{Z_i}}{\sum^N_{j=1}e^{Z_j}}\ for\ i=1,\dots,N\ and\ Z=(Z_1,\dots,Z_N)\in R^N$$

Namely, for each $z_i$ of the input element set $z$, it uses the standard exponential function, and normalize those results by dividing those values by the sum exponentials' value. This operation is to guarantee that the total components of the output vector is equal to 1.

In general, you can replace $e$ with a different base of $b\ (b > 0)$. When $0 < b < 1$, a smaller input component leads to a obvious result probability, as well concentrating the probability distribution around the minimum input value in which $b$ is reducing. On the contrary ($b > 1$), a larger input component leads to a larger result probability, the similarity is an increase in the $b$ value will make the probability distribution more concentrated, while the position is maximum input value.

The softmax is often used in ANN for multi-class differentiation tasks. In these networks, the softmax transformation results also sum to 1, and the following loss function is using by the optimized model, utilizing the maximum likelihood principle. However, in the case of high-dimensional classification, softmax leaves a lot of room for the loss function to perform optimization operations, which results in the performance reduction to some extent.when it comes to classification problems with high-dimensional outputs(more than 100 categories empirically), standard softmax and backpropagation do not take advantage of the sparsity of the categories, and as a result, softmax converces slowly on high- dimensional classification tasks. The softmax properties are often questioned to find better alternatives to the above problems. In particular, the first idea is an approximation of sampling methods, where a small portion of the result measurements is calculated. The second is to modify the softmax output layer by introducing a heuristic tree for the high-dimensional probiems.

For settling a matter that traditional softmax in high-dimensional classification, a simple and concise variant of softmax, sparse-softmax, is empirically studied. \cite{sun2021sparse} proposes this simple and scalable alternative to softmax, Sparse-softmax, which is specifically used for high-dimensional classification problems. One problem in the traditional softmax function is the probability distribution of the outcome, for each $Z_i$, $softmax (Z_i) \ne 0$. This is a drawback in dealing with multi-dimensional differentiation that require sparse probability distributions.For that we can manually set a hyperparameter $k$, and then select only the largest $k$ input values as vectors for the exponential normalization function $\Omega_k\in R^K$, while the others are masked to zero. Accordingly, softmax's cross entropy loss function will be modified accordingly.

\subsubsection{Compact Embedding}

Embedding is a technology used to reduce the dimensionality of data, which can map high-dimensional data (such as text, images, audio, etc.) into a low-dimensional data meanwhile retaining the local configuration and data semantic information.A advantages of Embedding are that it can reduce the storage space and computational complexity of the data, improve the visualization and interpretability of the data as well the generalization ability, and gain data robustness. A principle is to utilize the local linear relationship of data, that is, it is assumed that each data point can be approximated by the linear combination of other data points in its neighborhood. Therefore, Embedding is designed for a low-dimensional data space to get the coordinates about data element,it an also be approximated by the linear combination of other data points in its neighborhood, while keeping the neighborhood relationship in the original space unchanged. This can be done by minimizing an objective function that measures the reconstruction error between the data points in the original space and the lower-dimensional space.

It is a popular dimensionality reduction technology, which is widely used in many domain, for intance NLP, CV, bioinformatics, social network analysis and so on. The compact Embedding can be used to draw the data characteristic information, thereby improving the performance of subsequent tasks such as classification, clustering, regression, etc. The compact Embedding can also be used to explore and analyze the inherent structure and patterns of the data, thereby increasing the understanding and insight of the data.

\textbf{\numberedParagraph{AdaEmbed}}

Deep Learning recommendation models (DLRMs) are using increasingly large embedded tables to represent classified sparse features, such as video types. Each sparse feature is typically associated with an embedded table, where each instance of that feature is represented by a trainable embedded row (weight vector).

Unlike existing work that focuses primarily on optimizing DLRM for a given embed, \cite{lai2023adaembed} proposes a complementary system, AdaEmbed, that reduces the embedding size required for the same DLRM accuracy by performing embedding pruning in training. The key insight is that the access patterns and weights of different embeddings are heterogeneous across the embedded rows and change dynamically during training, which means that the importance of embeddings is different relative to model accuracy. However, for modern DLRms with billions of embeddings (terabytes), identifying the important embeddings and performing pruning is a challenge. Given the total embed size, AdaEmbed considers the embeddings with high runtime access frequency and large training gradients to be more important, and dynamically trims the less important embeddings by scale to automatically determine the embeddings for each feature.

An automatic training pruning system is introduced to actively optimize each feature embed (AdaEmbed) for better model accuracy. For a given embedding size, AdaEmbed extensibly identifies and preserves the embeddings that are more important to model accuracy at a particular time during the training process. AdaEmbed is introduced as a pruning system in automatic training to adaptively optimize the embedments of each feature on a large scale to obtain better model accuracy. Unlike existing model pruning work, which focuses on traditional models and or pruning model sizes after training is completed, AdaEmbed automatically identifies and retains important embedments of a given embedding size as training progresses to improve performance. The evaluation in an industrial setting shows that AdaEmbed provides superior model accuracy for its post-training pruning pairs, in addition to saving resources throughout the training process. Contains several parts: Embedding Monitor: Identifies important embeddings ; Intra-Feature Embedding Importance : For embedding of the same features, a data and model-perceived importance metric $EI(i)$ is introduced to capture the importance of each line i to the accuracy of the model. $EI(i)$ does not rely on an embedded weight that has been pruned to become obsolete, but rather a runtime combination of access frequency and gradient ; AdaEmbed Coordinator: Trim in time. To find the sweet spot between pruning overhead and quality, the AdaEmbed coordinator determines the correct pruning time to reduce the number of pruning rounds required, and instructs the memory manager to minimize the overhead per round of pruning when pruning embedded weights.

\textbf{\numberedParagraph{Graph Embedding}}

Graphic analysis allows for better quantification and control of complex networks, but traditional methods have high computational costs and excessive memory requirements associated with the high inconsistency characteristics of industrial networks. The graph embedding technique allows for the effective conversion of multi-dimensional sparse graphs into low density, dense, and continuous vector spaces while maintaining the structural characteristics of the graph. Another new class of embedding way uses Gaussian distribution, which includes crucial assessments of uncertainty. The main purpose of using this is to wrap the attributes of each node in smaller vectors; Therefore, the similarity of nodes in the original irregular space can be easily quantified using standard metrics that integrated into vector space. so, it is easy to quantify the similarity of nodes in the original complex irregular space with the measuring method. In addition to node embeddings, other ways of embedding such as edge, subgraph and fullgraph embeddings also can use.

The graph embedding methods used fall roughly into three broad categories: those based on matrix decomposition, those based on random walks, and those based on neural networks. Matrix factory-based methods: high-order methods construct matrices based on transformation probabilities and decompose them to get node inputs, however those are not easily scalable to deal with larger networks. The key point is the wider range of random walking methods and based-NN methods.Except the method of static graph, a class of dynamic graph embedding based on DL was also discussed.

Graph embedding methods based on vector points have three main cases: based on matrix factorization, random walk, and deep learning. Graph embedding depended on vector points is map high-dimensional graph vertexs into low-dimensional vectors of the underlying constructure while maintaining the structural information of the original graph.

Graph embeddings based on Gaussian distributions. An emerging method of graph embedding, called "graph embedding based on Gaussian Distribution" or "probability graph embedding", having great potential to deal with random graph embedding. Inspired by the word2Gauss method, the graph Gaussian embedding technique embeds words as Gaussian distribution potential functions into infinite-dimensional function Spaces. Thus, evey element is projected to a "soft region" in the potential data space, which can provide a better quantification of the word properties.

\subsection{Efficient Strategy}

In addition to the specialized design of various neural network components discussed in the previous chapters, there are some general strategies that can be used for efficient neural network structure design. In this chapter, we primarily introduce strategies such as low-rank module strategies, static parameter sharing, dynamic networks, and super networks. These strategies can be seamlessly integrated into any parameterized structure, as they are not specific to any particular architecture.

\subsubsection{Low-rank Module Strategies}
\label{Sec:lowrank1}
This section will discuss how to organically integrate low-rank strategies with network design, emphasizing the differences and connections with Section~\ref{Sec:lowrank2}. If trained parameters are decomposed using matrix or tensor decomposition and then replaced in the original modules, this also leads to improvements in model inference. However, for many scenarios, such as multi-source information fusion or multi-task learning, the computational cost of the original model becomes unsustainable (due to the curse of dimensionality). To avoid redundancy and inefficiency, this section primarily focuses on the low-rank format modeling design without using decomposition algorithms.

In recent times, the utilization of low-rank modules for constructing various common neural networks has reached a high level of maturity~\cite{DBLP:journals/corr/abs-2302-09019}. For instance, in order to expedite the training process of classical Convolutional Neural Networks (CNNs), low-rank formatted CP-convolution~\cite{DBLP:conf/nips/HayashiYSM19,phan2020stable,nekooei2022compression} has been developed by replacing classical convolutional weights matrix into the CP decomposed format. Similarly, additional CNNs can be implemented by applying other low-rank formats. Tucker decomposition, which is a widely employed tensor format, is frequently utilised in the context of Convolutional Neural Networks (CNNs).~\cite{DBLP:journals/ijon/PanWX22,Liu2022DeepNN}. Unlike basic Tucker formats, BTT-CNNs~\cite{li2017bt} integrate multiple Tucker decompositions through summation. everal other variations of BTT-CNNs~\cite{li2017bt} have also been suggested, and they have demonstrated superior performance compared to Tucker CNNs~\cite{li2017bt} owing to their enhanced capabilities.
Furthermore, the integration of highly compact Tensor Train (TT) formats into Convolutional Neural Networks (CNNs) has been proposed as an approach of developing TT-CNNs.~\cite{DBLP:journals/corr/GaripovPNV16,liu2022tt,qi2023exploiting}. When compared to TTs, Tensor Ring (TR) formats are typically more space-efficient~\cite{DBLP:conf/cvpr/WangSEWA18}, making TR-CNNs~\cite{DBLP:conf/cvpr/WangSEWA18} more potent than TT-CNNs.
Interestingly, to address issues related to degeneracy in tensorial layers, a stable decomposition method known as CPD-EPC~\cite{phan2020stable} has been proposed. It incorporates a minimal sensitivity design for both CP convolutional layers and hybrid Tucker2-CP convolutional layers.

Similar to CNNs, we can also employ low-rank modules to design other networks like, RNNs and Transformers.
For instance, using the CP and Tucker formats, respectively, the CP-RNN and Tucker-RNN~\cite{DBLP:journals/ijon/PanWX22} can be constructed directly. The CP-RNN consistently obtains the smallest parameter size among various tensor formats due to its highly compact low-rank structure.
To achieve a high parameter compression ratio, the TT-RNN~\cite{DBLP:conf/icml/YangKT17} adopts the TT format in an RNN. However, the TT-RNN's linear structure restricts the flexibility and capacity of TT-based models for data representation. To overcome this limitation, Tensor Rings (TRs) were proposed as a solution, connecting the endpoints and forming a ring structure~\cite{zhao2016tensor} to unleash the full potential of linear designs.

The fundamental basis of BTT-RNN~\cite{Ye_2018_CVPR,li2017bt} is in the aggregation of Tucker decompositions. The framework has the capability to automatically acquire knowledge about inter-parameter correlations, hence enabling the implicit removal of redundant dense connections and enhancing overall performance.
The concept of the MPO structure was introduced to decompose each matrix within the Transformer model~\cite{DBLP:conf/acl/LiuGZXLW20}. This approach aims to establish the MPO structure by decomposing the matrices in order to improve the efficiency and performance of the Transformer model. This decomposition yields both small auxiliary tensors and core tensors, which capture the essential information. The auxiliary tensors are then trained again using a tuning approach, which increases performance. In order to preserve the fundamental elements of the initial matrix, the weight of the central tensor is temporarily held constant.
Specially designed low-rank structures can be utilized to enhance performance in scenarios involving multiple tasks. Yang et al.~\cite{yang2017deep} introduced the Tensor Train Multitask (TTMT) and Tucker Multitask (TMT) models, which leverage the TT and Tucker formats respectively. The TMT models address the challenge of negative transfer in a sharing architecture, while also minimizing parameter volume in a flexible framework, mitigating any potential performance degradation.
The capacity to create innovative neural network structures using diverse low-rank formats is at our disposal, even for those without known numerical decomposition procedures.

\subsubsection{Static Weight Sharing}

Static weight sharing is a method for reusing weights in neural networks; unlike intermediate vectors, weights are shared among all instances and remain constant throughout inference. Multiple model optimisation algorithms are utilised to recycle parameters over many layers or for diverse activities, ensuring efficient memory consumption. Cross-task parameter sharing and cross-layer parameter sharing are two well-liked approaches for exchanging static weights.

\textbf{Cross-task sharing} is widely adopted in the context of multi-task, multi-domain, or multi-lingual scenarios, as evidenced by several studies~\cite{DBLP:journals/corr/RamsundarKRWKP15,DBLP:conf/acl/DuongCBC15,DBLP:conf/acl/SogaardG16,DBLP:conf/emnlp/HashimotoXTS17,DBLP:conf/iclr/YangSC17,DBLP:journals/jmlr/RaffelSRLNMZLL20,yang2017deep}. The fundamental idea behind cross-task sharing is to make parameter sharing possible across all tasks, languages, and domains. A common strategy is multi-task learning, which may be accomplished by hard or soft parameter sharing. Hard parameter sharing minimises the number of parameters, while soft parameter sharing does not entail exchanging network components across activities. This research focuses especially on the usage of hard parameter sharing.

\textbf{Cross-layer sharing} is another technique widely used to improve parameter efficiency. Extensive research has been conducted to explore parameter sharing across layers, as evidenced by various studies~\cite{DBLP:conf/iclr/DehghaniGVUK19,DBLP:conf/nips/BaiKK19,DBLP:conf/iclr/LanCGGSS20}.

Savarese and Mottini~\cite{DBLP:conf/iclr/SavareseM19} proposed a significant parameter sharing approach that involves the use of a global library of templates. This scheme allows for the derivation of parameters for each layer of a convolutional neural network (CNN) by linear combinations.
Similarly, the Universal Transformer model proposed by Dehghani et al.\cite{DBLP:conf/iclr/DehghaniGVUK19} shares the same parameters across all layers. Building on these foundations, Lan et al.\cite{DBLP:conf/iclr/LanCGGSS20} extended the application of cross-layer sharing mechanisms to both pre-training and fine-tuning scenarios, while diverse sharing strategies were proposed by Li et al.~\cite{DBLP:journals/corr/abs-2104-06022}.

\subsubsection{Dynamic Networks}

Furthermore, in addition to the direct implementation of static sharing, there have been introductions of dynamic solutions aimed at determining the specific layers or components that should be shared. Dynamic networks refer to neural networks that possess dynamic model structure, whereby the computational structure and parameters are dynamically defined based on the specific requirements of the task at hand. 
Consequently, this type of network can mitigate computational expenses and enhance the adaptability of networks. For the implementation of dynamic networks, generally, the following common paradigms are available:

\textbf{Cascading-style dynamic networks} 
These cascade designs~\cite{viola2001rapid,lienhart2002extended,viola2004robust} were first introduced with the purpose of addressing imbalanced binary classification issues. The researchers used a cascading approach by integrating many rudimentary models and selectively passing the input to the subsequent model only when the current model exhibited uncertainty in its prediction. 

\textbf{Early-exiting-style dynamic networks} have been proposed in previous works such as Teerapittayanon et al.\cite{DBLP:conf/icpr/Teerapittayanon16}, Bolukbasi et al.\cite{DBLP:conf/icml/BolukbasiWDS17}, and Huang et al.\cite{DBLP:journals/corr/abs-2103-01148}. 
These networks are designed with the incorporation of multiple internal classifiers on intermediate layers, allowing for intermediate predictions and the ability to determine whether to continue the forward process or halt. One specific instance of this early exiting technique is dynamic halting, where parameters, including the final classifier, are shared across layers. This approach facilitates iterative inference through a shared layer instead of processing samples with multiple individual stacked layers.
The concept of adaptive computation time (ACT) was introduced by Graves~\cite{DBLP:journals/corr/Graves16} for recurrent models. This mechanism automatically determines the necessary number of iterations to compute each input symbol or token. 
Expanding on the advancements made in this work, such mechanism has been successfully applied to various architectural designs, such as ResNets and Transformers. Notably, SACT~\cite{DBLP:conf/cvpr/FigurnovCZZHVS17} implemented dynamic halting in a multidimensional manner, incorporating both coarse structures across multiple layers within the same block, as well as fine-grained structures that encompasses all spatial positions.
This implementation enhances the efficiency of the network. Another example is the Universal Transformer~\cite{DBLP:conf/iclr/DehghaniGVUK19}, where all layers within the Transformer model are shared. This shared architecture improves computational efficiency while maintaining performance.
In summary, the incorporation of multiple internal classifiers and the utilization of dynamic halting techniques, such as ACT, have paved the way for more efficient neural networks. These techniques have been successfully applied to various architectures, allowing for adaptive and shared computation, ultimately improving the overall computational efficiency of the models.

\textbf{Skipping-style dynamic networks} have been successfully applied in various models, including SkipNet~\cite{DBLP:conf/eccv/WangYDDG18}, ConvNet-AIG~\cite{DBLP:journals/ijcv/VeitB20}, and BlockDrop~\cite{DBLP:conf/cvpr/WuNKRDGF18}. These networks improve efficiency during the forward process by introducing additional policy modules that decide which layers to skip or not.

\textbf{Mixture-of-Experts style} is another notable example among dynamic models, as highlighted in works such as Lepikhin et al.~\cite{DBLP:conf/iclr/LepikhinLXCFHKS21} and Switch Transformer~\cite{DBLP:journals/corr/abs-2101-03961}. These models employ a layered structure where each layer consists of multiple experts, but only a subset of experts is activated for each input instance.
For instance, Switch Transformer introduces a switch feed-forward layer, replacing the conventional feed-forward layer in the Transformer model. This layer comprises a routing module and multiple structurally identical experts. During execution, only a single expert is activated for each token in each switch layer. This approach allows for efficient and selective computation.
MMoE~\cite{ma2018modeling} is another example that leverages multiple expert submodels and a shared gating network to implicitly model relationships among multiple tasks with different label spaces. By utilizing this approach, Mixture-of-Experts-style networks offer a cost-effective and pragmatic solution to adapt and train large models with sparse activation, compared to typical dense computation architectures.
In summary, the incorporation of Mixture-of-Experts style in dynamic models provides a practical and efficient way to adapt and train large models with sparse activation. These models, such as Switch Transformer and MMoE, demonstrate the effectiveness of selectively activating experts for specific input instances, resulting in improved computational efficiency and performance.

\subsubsection{Super Network}

Recently, there has been a growing interest in the concept of a super network. These studies involve training a super-network alongside various sub-networks using task-specific loss functions. The appropriate sub-network is obtained based on the given resource constraints.
One notable example is the introduction of slimmable neural networks by Yu et al.~\cite{DBLP:conf/iclr/YuYXYH19}. These networks offer instant and adaptive trade-offs between accuracy and efficiency by allowing selection from several predefined widths. Expanding on this idea, Yu and Huang~\cite{DBLP:conf/iccv/YuH19} introduced US-Nets, which enable arbitrary width selection.
Another approach proposed by Fan et al.~\cite{DBLP:conf/iclr/FanGJ20} is an elastic network that can obtain sub-networks of varying depths from a huge neural networks without the need for additional fine-tuning. This further enhances adaptability. Additionally, adaptability extends to temporal or input length selection.
For instance, Kim and Cho~\cite{DBLP:conf/acl/KimC20} introduced the length-adaptive Transformer, which enables flexible and progressive length reduction. This length-adaptive Transformer seamlessly integrates into downstream tasks and meets efficiency requirements by adjusting the length configuration accordingly.
Overall, the concept of a super network, along with the advancements in adaptive selection of sub-networks and lengths, has opened up new possibilities in achieving a balance between accuracy and efficiency in neural networks.

\subsection{Green NAS}

In addition to model design, research efforts have aimed at creating search-efficient networks for resource-constrained devices, such as mobile devices. These studies draw inspiration from neural architecture search for green machine learning. However, traditional NAS consumes significant resources, leading to the emergence of various paradigms for green NAS.


\subsubsection{One-shot NAS}
One-shot NAS~\cite{hu2018manifold, fu2020scene, zhao2019object} represents a novel paradigm that decouples architecture search from supernet training. In one-shot NAS, evolutionary algorithms are employed to sample numerous architectures during the training phase of the supernet.
For instance, B. Gabriel et al.~\cite{bender2018understanding} trains the over-parameterized network while gradually dropping out operators, allowing their weights to co-adapt. SPOS~\cite{guo2019single} builds upon this idea and introduces uniform sampling for supernet training, where only one path is activated during each optimization step, optimized using standard gradient-based methods. FairNAS~\cite{chu2019fairnas} further enhances the one-shot approach by enforcing strict fairness in both supernet sampling and training. AutoSlim~\cite{Yu2019AutoSlimTO} improves correlation by optimizing the maximum, minimum, and intermediate paths through in-place distillation. 
\subsubsection{Zero-shot~(Training Free) NAS}
Unlike one-shot NAS methods, where one model still requires training, zero-shot NAS approaches aim to predict performance before actual training. These methods evaluate networks within the search space based on a score, without relying on post-training performance or weights. In recent years, researchers have increasingly recognized the utility of training-free metrics in assessing the capacity of neural networks. 

Mohamed et al.\cite{abdelfattah2021zero} assesses the effectiveness of various pruning-at-initialization criteria in the context of NAS. NASWOT~\cite{mellor2021neural} utilizes the count of linear regions to rank different networks, while TE-NAS~\cite{chen2021neural} extends this by incorporating linear regions with the neural tangent kernel to rank networks based on their expressivity and trainability. Their approach demonstrated remarkable efficiency, delivering results within seconds and relying solely on a single device. Jisoo et al.\cite{mok2022demystifying} have highlighted the instability of NTK-based metrics across different search spaces and initializations.
KNAS~\cite{xu2021knas} introduces a hypothesis that gradients can be an evaluation measure for randomly-initialized networks. KNAS undertook a theoretical analysis to substantiate this hypothesis and identify a suitable feature, MGM (mention the full name if needed).

\subsubsection{Resource Constrained NAS}

In order to balance performance and computational cost, several follow-up research have now moved their attention to resource-aware NAS (Neural Architecture Search) techniques. By adopting a policy-based reinforcement learning search algorithm, MONA~\cite{hsu2018monas}, for instance, considers the quantity of multiply and accumulation operations as a restriction that can be instantly included into the reward function.

In differentiable NAS frameworks, incorporating computational cost metrics into the loss function as penalty terms is a straightforward approach. Metrics such as FLOPs, parameter size, and latency play a crucial role in differentiable NAS frameworks. These metrics are commonly used to evaluate the computational cost of architectures. For example, ProxylessNAS~\cite{cai2018proxylessnas} takes a unique approach by modeling the latency of the architecture as a continuous valued loss. Thus, ProxylessNAS effectively considers both accuracy and efficiency during the optimization process. Another method, FBNet~\cite{wu2019fbnet}, employs a lookup table to estimate the latency associated with candidate operations. 
 This allows FBNet to make informed decisions about the computational cost of different operations during the NAS search.
In the case of SNAS~\cite{xie2018snas}, the focus is on ensuring the differentiability of resource constraints.SNAS accomplishes this by representing the FLOPs and memory access cost of the child network through linear functions modeled by binary random variables. This enables efficient gradient-based optimization while accounting for computational constraints.
Addressing the limitations of previous differentiable NAS methods, RecNAS~\cite{peng2022recnas} is specifically designed to search for superior performance architectures that satisfy specific given constraints. RecNAS introduces improvements in the search space, search strategy, and adaptability to resource constraints. By refining the search process and optimizing the utilization of computational resources, RecNAS aims to discover architectures that meet specified constraints without compromising performance.

\section{Energy-Efficient Training}
\subsection{Efficient Training Paradigm}
Conventional machine learning models rely on capturing a static data distribution, which makes them prone to overfitting~\cite{ying2019overview} and limits their ability to generalize well~\cite{Jiang2020Fantastic}, especially when training data is limited. Additionally, this approach is difficult to scale to multitask scenarios within memory and computation constraints, (e.g. for on-device applications), as it requires storing a full set of model parameters for each task and training from scratch~\cite{guo2020parameter}. Recently, researchers have started exploring more advanced training paradigms to fully utilize available data and computational resources. These paradigms optimize the training process, improve generalization, and enable learning in resource-limited or dynamic environments.

\subsubsection{Pre-training and Fine-tuning}
With the rapid development of deep neural network architectures, the demand for data by models has gradually increased, prompting researchers~\cite{abdelhamed2018high,shao2019objects365} to invest a great deal of effort in manually constructing high-quality datasets to ensure effective learning of neural models for specific tasks.
However, manual annotation of large-scale data is time-consuming and economically costly~\cite{liu2021self}. To address this issue, researchers~\cite{weiss2016survey, zhuang2020comprehensive} have proposed the paradigm of transfer learning, which is also an important milestone in machine learning. Instead of training models from scratch with a large amount of data, transfer learning imitates the ability of humans to learn new tasks from limited samples by drawing on past knowledge. Specifically, it establishes a two-phase learning framework: the pre-training phase, where models are trained on large-scale datasets to capture general knowledge and learn basic representations, followed by the fine-tuning stage, where this knowledge is transferred to smaller datasets or target tasks.

The pre-training and fine-tuning learning framework is initially popularized in the field of computer vision (CV). Several convolutional neural networks (CNNs), such as AlexNet~\cite{krizhevsky2012alexnet} and VGG~\cite{simonyan2014very}, are pre-trained on the ImageNet, a supervised visual recognition dataset~\cite{deng2009imagenet}, to learn generic features. 
By leveraging the abundant data from ImageNet, these models faster convergence on the target tasks compared to training from scratch. ResNet~\cite{he2016deep} further proposes shortcut connections with residual layers and its combination with ImageNet pre-training achieved impressive performance on downstream tasks. This triggers the wave of exploring pre-trained models (PTMs) in various CV tasks, including image classification~\cite{he2016deep, lee2015deeply}, object detection~\cite{girshick2015fastrcnn}, and image segmentation~\cite{long2015fullyseg}. 
The success of transfer learning in computer vision has also led to its adoption in other domains. In NLP, early PTMs focused on the construction of word embeddings. Word2Vec~\cite{mikolov2013word2vec} trains shallow neural network model using the Continuous Bag of Words (CBOW) architecture to produce word embeddings that measured word similarity. Glove~\cite{pennington2014glove}, on the other hand, constructed a word-word co-occurrence matrix and performed matrix factorization to generate word embeddings that captured linear relationships based on co-occurrence probabilities.
With the development of deep models, GPT~\cite{radford2018gpt} and BERT~\cite{kenton2019bert} are proposed inspired by the success of Transformers~\cite{vaswani2017transformer}. GPT combined the Transformer architecture with an autoregressive language modeling objective, the likelihood of predicting the next word in a sequence given the previous words. BERT utilized a bidirectional Transformer as the backbone and applied masked language modeling, where tokens were randomly masked with the objective to recover them. 
These pre-training objective allows the model to learn rich contextual representations of words and capture long-range dependencies in language. Through fine-tuning these large-scale PTMs with a relatively small amount of task-specific data, they can exhibit remarkable performance on downstream NLP tasks by benefiting from the learned contextual knowledge from pre-training. The introduction of self-supervised learning and Transformers have revolutionized Artificial Intelligence field and propelling large-scale PTMs to the forefront.

The pre-training and fine-tuning learning paradigm offers significant benefits for efficient training.
First, the pre-training phase provides a wealth of knowledge that allows the fine-tuning phase to effectively handle target tasks even with limited samples.
Recent studies have shown that pre-training on a large-scale dataset such as ImageNet allows models to achieve state-of-the-art performance on specific image classification tasks with only a few labeled samples available~\cite{donahue2014decaf, matthew2014visualizing}.
Second, pre-training initializes the model with robust and transferable representations. These initial weights serve as a good starting point for the target task, enabling the model to converge faster during fine-tuning.
For example, ~\cite{he2019rethinking} shows that pre-training on ImageNet accelerates convergence, particularly in the early stages of training.
Finally, although pre-training large models can be computationally expensive, it can be done once and shared across multiple target tasks. 
For instance, the pre-trained models T5 is used as a base model for various downstream tasks, such as question answering, summarization, or machine translation, saving computational resources and training time~\cite{raffel2020t5}.
Once a model is pre-trained, fine-tuning can be performed on various smaller datasets, reducing the computational requirements.

Up to now, the emergence of increasingly larger pre-trained models has brought numerous benefits, including improved effectiveness on both known and unknown tasks. But one problem is the need to store modified copies of all the LM parameters for each task, which becomes prohibitively expensive due to the large size of LMs. Furthermore, adapting hundreds of millions of parameters incurs high computational costs~\cite{abs-2203-06904survey}. To tackle these issues, researchers have been working on parameter-efficient tuning methods that offer a promising solution ~\cite{LesterAC21Prompt, HoulsbyGJMLGAG19Adapter}. These methods aim to stimulate models with only a small portion of tunable parameters, thereby significantly reducing the computational and storage requirements for model adaptation. These approaches enable more efficient utilization of pre-trained LMs, making it feasible to adapt them to specific tasks without incurring overwhelming costs. Some notable examples of these parameter-efficient tuning methods include Prompt tuning~\cite{LesterAC21Prompt}, Adapter~\cite{HoulsbyGJMLGAG19Adapter}, and Lora~\cite{HuSWALWWC22Lora}. In the following section (Sec.~\ref{peft}) we will delve into these methods in detail.

\subsubsection{Progress Learning}
In mainstream deep learning training schemes, every training iteration is typically treated equally, where all network parameters participate in training on randomly shuffled datasets. However, recent studies have revealed that training in a structured manner, gradually increasing the complexity or difficulty of the training process over time, can lead to accelerated model training and more efficient data utilization. This new training paradigm is called progressive learning. It has been applied in various ways and has shown good acceleration performance in a number of works, e.g., AutoProg~\cite{li2022automated} achieves training acceleration of up to 85.1\% for Vision Transformer~\cite{dosovitskiy2020image} against the conventional training scheme. They can be categorized into two main approaches: data-centric and model-centric.

\textbf{Data-centric progressive learning}. It is based on the intuition that information is better learned by starting with simple tasks and gradually increasing task complexity. Several techniques have been developed in this area. Progressive Resizing~\cite{karras2018progresize} trains GAN model by progressively growing both the generator and discriminator, initially using low-resolution images and gradually introducing higher-resolution details as training progresses. EfficientTrain~\cite{wang2022efficienttrain} introduces a clipping operation in the Fourier spectrum of the inputs, enabling the model to focus on learning from low-frequency components. Kocyigit et al.~\cite{koccyiugit2023accelerating} leverage progressive resolution training, learning rate scheduling, and hard augmentation selection to accelerate and stabilize self-supervised learning of ViT. In the field of NLP, ShortFormer~\cite{press2021shortformer} treats token length as the resolution in CV and show that training transformers with shorter subsequences initially and gradually transitioning to longer subsequences can achieve significant speedups. In similar, SLW~\cite{li2022stability} employs Sequence Length Warmup, which progressively increases the sequence length during the early stages of training. This stabilizes the training process for models like GPTs, resulting in improved training efficiency and wall-clock speedup. Zhang et al.~\cite{zhang2021reducing} explore curriculum learning for pre-training by selecting the next training samples based on their loss reduction. In these experiments, the training of GPT-2 and BERT models can be accelerated by a factor of 2 to 4, surpassing the performance of the original models while utilizing fewer data.

\textbf{Model-centric progressive learning}. Recently, the training costs associated with large-scale pre-trained model have become increasingly expensive. To address this issue, the concept of model-centric progressive learning has gained traction.
Model-centric progressive learning focuses on exploring the limited capacity of large-scale models during the early periods of training. As the training progresses, the model's capacity gradually increases, and all parameters are restored to complete the entire training process. Several approaches have been developed in this field to improve training efficiency and reduce costs.
Net2Net~\cite{chen2015net2net} accelerates training by transferring the knowledge from the previously trained network to a deeper or wider network using function-preserving transformations. EfficientNetV2~\cite{tan2021efficientnetv2}, on the other hand, combines training-aware neural architecture search and scaling techniques to optimize both training speed and parameter efficiency. AutoProg~\cite{li2022automated} introduces an automated sub-network architecture selection mechanism via estimating the performance of sub-networks through an elastic supernet, enabling efficient model training. Budget training~\cite{xia2023budgeted} dynamically controls the activation rate of the model by assessing redundancies within different modules, such as attention heads, hidden dimensions in MLP, and visual tokens.
For NLP, there is a growing body of literature focused on accelerating BERT pre-training through progressive techniques. These include progressively stacking~\cite{gong2019efficient}, dropping layers~\cite{zhang2020accelerating} or expanding in multiple dimensions of the models~\cite{gu2021Progressive}.

\subsubsection{Continual learning}
The current prevailing paradigm for machine learning is to train on a given dataset to produce a trained model specifically for a particular purpose. However, a practical intelligent system should be able to incrementally learn from evolving data and knowledge. For example, a classifier trained to recognize cats and dogs should be able to continuously learn and adapt when presented with new types of animals. 
Traditional learning approaches require retraining the entire model from scratch whenever new data or information is introduced. This can be time-consuming, resource-intensive~\cite{guo2020parameter}. Alternatively, solely fine-tuning a trained model with new data might lead to the loss of previously acquired knowledge, a phenomenon known as Catastrophic Forgetting~\cite{mccloskey1989catastrophic}. To this end, Continual Learning (CL) has garnered increasing attention as a solution to address these challenges. CL focuses on training models with streaming tasks that entail non-stationary distributions. By incorporating CL, models can incrementally update their knowledge and adapt to new information without discarding what they have previously learned. 

Assume there is a sequence of tasks $\mathcal{T} =\{T_1,T_2,...T_k\}$, each $T_i$ contains a set of $N_i$ labeled pairs $\{\mathbf{X}_{i},\mathbf{Y}_i\}=\{(X_{ij},Y_{ij})\}_{j}^{N_i}$, where these pairs are independently and identically drawn from a specific data distribution $p_i(x,y)$ defined on the input space $\mathcal{X}$ and the label space $\mathcal{Y}$. The objective of continual learning is to handle the arrival of each new task $T_i$ while ensuring that previously learned knowledge from tasks $\{T_1, ... T_{i-1}\}$ is retained, when having no access to old data. Continual learning algorithms can be broadly categorized into three main groups:

\textbf{Replay-based Methods.} In recent years, exemplar replay has emerged as one of the most widely-used methods to mitigate catastrophic forgetting~\cite{aljundi2019gradient,bang2021rainbow,arani2021learning,wang2021memory}. The intuitive behind it is to leverage the small memory buffer containing previous data samples to reconstruct the task distribution and thereby alleviate catastrophic forgetting. Various works have focused on designing sampling methods to select informative samples that best represent the previous distributions. RWalk~\cite{chaudhry2018riemannian} selects exemplars with high entropy and near the decision boundary. InfoRS ~\cite{sun2021information} adopts the concept of information gain and proposes a selection criterion that ensures diversity while reducing the inclusion of outliers. Gradient Coreset Replay ~\cite{tiwari2022gcr} maintains a buffer to estimate the gradient of all the data encountered so far with respect to the current model parameters. 
In addition to storing real samples, another approach to memorize previously learned data distributions is generative replay. This involves utilizing generative models for data generation, along with task solvers for prediction. Generative models used in this context include generative adversarial networks~\cite{shin2017continual, chenshen2018memory}, variational autoencoders~\cite{kemker2018fearnet}, and diffusion models~\cite{gao2023ddgr}. However, generative models also face the challenge of catastrophic forgetting.
To address this, DGR~\cite{shin2017continual} learns new tasks with replaying the pseudo data, while MeRGAN~\cite{chenshen2018memory} enforces consistency between data sampled with random noise from old and new models.

\textbf{Regularization-based methods.} Regularization-based methods play a crucial role in addressing catastrophic forgetting by applying constraints to the model parameter update process. These constraints aim to consolidate previously acquired knowledge when learning new tasks. 
There are two main types of regularization-based methods.
Parameter regularization methods focus on preserving important weights that contain crucial knowledge. EWC~\cite{kirkpatrick2017overcoming} calculates an importance matrix using the Fisher information matrix and regularizes important parameters using an L2 loss. K-FAC~\cite{lee2020kfac} extends the Fisher information matrix approximation with the Kronecker factorization technique. In contrast, SI~\cite{zenke2017si} estimates importance in an online manner based on its contribution to the loss function. Another approach to imposing restrictions is by controlling the learning rate for different parameters. UCB~\cite{ebrahimi2019uncertainty} introduces the Bayesian Gradient, which adjusts the learning rate of mean weights in a Bayesian neural network in proportion to the uncertainty of the weights, quantified by their standard deviation.
Function regularization methods aim to preserve previously learned knowledge by introducing a penalty term that compares the output of old and new models. Knowledge distillation is a widely used technique in this category, where a model trained on old tasks acts as a teacher network, and the current model is the student network. ICaRL~\cite{rebuffi2017icarl} utilizes the sigmoid outputs of each class in previous tasks as the teacher's knowledge, while LwM~\cite{dhar2019lwm} exploits attention maps. PODNet~\cite{douillard2020podnet} minimizes the difference of pooled intermediate features instead of performing element-wise comparisons, which proves to be more effective for class incremental learning.

\textbf{Architecture-based Methods.} This approach address catastrophic forgetting by dedicating different model parameters to each task, ensuring that no forgetting occurs. 
When there are no constraints on the architecture size, dynamic networks can be employed, where new branches are added for new tasks while freezing the parameters related to previous tasks. 
This expansion of the neural architecture can be formulated as a reinforcement learning problem~\cite{XuZ18rcl} or achieved through neural architecture search~\cite{LiZWSX19l2grow}.
Additionally, DER~\cite{YanX021der} aggregates features using an enlarged fully connected layer and compresses them with a channel-level mask-based pruning strategy. FOSTER~\cite{WangZYZ22foster} tackles the learning process as a feature-boosting problem and incorporates an additional model compression step to reduce redundancy.
Alternatively, parameter isolation involves maintaining a static architecture with fixed parts allocated to each task.
HAT~\cite{SerraSMK18hat} explicitly optimizes a binary mask to select dedicated parameters for each task, while freezing (almost) all parameters related to the old tasks to prevent catastrophic forgetting. CCGN~\cite{AbatiTBCCB20CCGN} equips each convolutional layer with task-specific gating modules to select filters for a given input. During inference, a task predictor is used to determine which gating modules to use.


\subsection{Training Data Efficiency}
\subsubsection{Few-Labeled Data}

\textbf{\numberedParagraph{Active Learning}}

Many existing deep neural networks (DNNs) can only show excellent performance with a large amount of labeled data, the annotation of which is labor-intensive, time-consuming, and economically expensive. To address this challenge, Active Learning (AL) tries to select the optimal unlabeled samples for labeling and training under limited labeling resources or manpower to achieve the best model performance. Categorized by scenario, active learning can be divided into the following three types: Pool-Based scenario \cite{liu2021pool,kravberg2022active, holzmuller2023framework}, Stream-based scenario \cite{pham2022stream,aguiar2023active, bemporad2023active} and Query Synthesis scenario \cite{mayer2020adversarial, piedboeuf2022working}. For the Pool-Based scenario, AL selects samples from an unlabeled data pool for labeling with certain strategies, whereas for the Stream-based scenario, the data comes from a data stream, and AL needs to decide whether each incoming data needs to be annotated or not. As for Query Synthesis scenario, AL generates one or more new data points based on existing unlabeled data, which in turn are labeled for training.

Most of the existing work focuses on the classification tasks in pool-based scenario, which we can formally define in the following format \cite{sener2017active, zhan2022comparative}: suppose we have a labeled set $D_l=\{(x_j,y_j)\}_{j=1}^M$ and a much larger unlabeled data pool $D_u=\{x_i\}_{i=1}^N$, where $M \ll N$, $y_i\in\{1,\dots,C\}$ for a $C$ class classification problem. In each iteration of AL, we need to select a subset $D_q$ of samples with size $b$ from $D_u$ based on current model parameters $\theta$ and an acquisition function $\alpha(x;\theta)$. Getting the top-$b$ data samples $D^{\ast}_q=argmax_{x\in D_u}^b \alpha(x;\theta)$, we then update $D_l$ and $D_u$ after labeling, and the model $\theta$ is retrained on $D_l$ to get new parameters $\theta'$. We will keep this process until the model reaches the desired performance or the budget for annotation is exhausted.

Following the definition above, AL methods differ mainly in the query strategy, i.e., the acquisition function $\alpha(x;\theta)$, and we can roughly categorize them into three main branchs: Uncertainty-based, Diversity-based and Learn to Score.

\textbf{Uncertainty-based} Uncertainty-based strategies are the most common query strategies in AL, which tend to select samples that are the most uncertain for the current model. Uncertainty can be measured by various criterions, such as entorpy \cite{wu2021novel, wu2022entropy}, margin \cite{netzer2011reading, ducoffe2018adversarial}, disagreement \cite{beluch2018power, gao2020consistency}, least confidence \cite{wang2014new, agrawal2021active} and mean standard deviation \cite{kendall2015bayesian, kampffmeyer2016semantic}. In particular, the Disagreement-based query strategies require a group of models, called committees, which jointly decide the uncertainty of each unlabeled data, so this type of works are also called Query-By-Committee. In addition to measuring uncertainty with these criterions, there are also some other practical methods, for example, \cite{ducoffe2018adversarial} queries the unlabeled samples which are closest to their adversarial attacks; \cite{ash2019deep, wang2022boosting} utilize gradients to guide the selection of unlabeled data; \cite{yoo2019learning} trains a module for predicting loss on labeled data and selects the unlabeled samples with the highest predictive loss.

\textbf{Diversity-based} Diversity-based strategies take the representativeness of the data into account, which means it measures how much the labeled instances are aligned with the unlabeled instances in distribution. In other words, the diversity-based approaches expect the selected unlabeled samples to act as a surrogate for the entire dataset. The most intuitive way to achieve this goal is clustering: \cite{sener2017active, kim2022defense} selects a batch of representative samples based on a core set; \cite{venkatesh2020ask} utilizes K-means++ \cite{arthur2007k} on the learned gradient embeddings to select the query samples. Another common idea is to measure the alignment of the distribution between labeled and unlabeled data: \cite{du2015exploring} calculates the triple similarities that include the similarities between a query sample and the unlabeled set, between a query sample and the labeled set, and between any two candidate query samples to measure the representativeness; \cite{biyik2019batch} captures diversity by constructing a pair-wise (dis)similarity matrix and calculating its determinant; \cite{gissin2019discriminative, sinha2019variational, shui2020deep} trains a GAN and tries to make the discriminator distinguish between the distribution of labeled and unlabeled data.

\textbf{Learn to Score} Both Uncertainty-based and Diversity-based query strategies are based on heuristics approaches, while the same method may perform differently in various scenarios. Thus, there are some researchers trying to learn the appropriate query strategy directly in the sampling process. Such approaches generally take three perspectives: learn a score function \cite{yoo2019learning, li2021learning}, imitation learning \cite{liu2018learning, gonsior2021imital}, and reinforcement learning \cite{hsu2015active, fang2017learning, wang2020deep}. For example, \cite{li2021learning} adds an extra loss prediction model and calculates the ranking loss instead of the ground truth loss, which in turn directly provide a ranking of the unlabeled samples; \cite{wang2020deep} trains a RL network in an actor-critic way by taking the prediction probability of the whole unlabeled set as the state, the strategy to get a rank of the unlabeled set as the action, and the difference between the prediction value and true label of the selected instances as the reward.

In conclusion, active learning tries to find out the optimal unlabeled samples under limited labeling resources, which would in turn maximize the performance of the model. The three categories of active learning (Uncertainty-based, Diversity-based, and Learn to Score) are not independent of each other, and there may be some combinations or trade-offs between them.

\textbf{\numberedParagraph{In-context Learning}}

As the dimensions of the model expand, the conventional paradigm of pre-training and fine-tuning \cite{devlin2018bert} is no longer applicable to massively scaled models. In-context learning (ICL) \cite{min2022rethinking} aims to enable the large-scale pre-trained model to learn latent patterns embedded within the demonstrations, thereby making increasingly accurate predictions. Recently, there has been extensive research in the field of ICL\cite{dong2022survey}. As a new paradigm applied to large-scale models, ICL has many advantages. First, different from supervised learning, ICL does not require updating the model parameters, which saves training costs and makes it more environmentally friendly and low-carbon. Second, ICL requires manually constructing templates to communicate with the large model and guide it to generate correct outputs. This explicit injection of knowledge provides greater interpretability. Third, ICL resembles the human thinking process, where problems are solved through providing questions, hints, and feedback, allowing the model to learn and reason step by step. 
Through in-context learning, the larg e model can learn the maximum likelihood estimation of potential answers by leveraging a few examples.

Although ICL has shown multiple advantages, there is also one improvement method for ICL that we refer to as the warmup strategy. The main purpose of warmup is to provide appropriate initialization and preparation for the model before engaging in an ICL task through pre-training or an initial phase. Different from finetuning, warmup can enhance the overall capability of the model rather than its performance on specific tasks. There are mainly two methods of warmup, one is supervised in-context training, and the other is self-supervised in-context training. 

\textbf{Supervised In-context Training.} By constructing corresponding supervised data for in-context and performing multi-task training, followed by in-context fine-tuning, the gap between pre-training and downstream in-context learning (ICL) can be narrowed. Min et al. proposed a new framework called MetaICL\cite{min2022metaicl}. MetaICL enables LLM to undergo in-context training on a large number of training tasks, allowing the model to better understand new tasks through in-context learning. This improves the model's performance on new tasks by leveraging its ability to comprehend the context. Besides, there is a lot of research focusing on instruction tuning\cite{mishra2022cross,thoppilan2022lamda}. For example, Thoppilan et al.\cite{thoppilan2022lamda} tune LaMDA-PT by using instruction templates on over 60 NLP datasets.

\textbf{Self-supervised In-context Training.} There is a large amount of unlabeled data, and how to effectively utilize it is a worthwhile research direction. Chen et al.\cite{chen2022improving} transformed the raw text into several forms that align with pre-training tasks, which are as follows: next sentence generation, masked word prediction, last phrase prediction and classification. 

The performance of LLM models depends on the format of the demonstration instances\cite{zhao2021calibrate}. During inference, LLM models may produce different results for the same question when different examples are provided. There are limitations on the input length of language models, so it is crucial to select appropriate portions from numerous samples as examples. The process of selection can be broadly categorized into two methods: unsupervised and supervised. The unsupervised methods can be categorized into the following types: based on distance metrics methods\cite{liu2022makes,sorensen2022information}, based on mutual information methods\cite{tanwar2023multilingual}, and those that utilize the language model itself to generate demonstration examples\cite{kim2022self}. As for supervised methods, there are several types. The first involves using an unsupervised retriever to recall a number of similar samples, followed by scoring through a supervised learning-trained Efficient Prompt Retriever to select the most suitable sample. Additionally, there are methods based on prompt tuning and reinforcement learning for sample selection.

\textbf{\numberedParagraph{Meta Learning}}

Meta-learning refers to the iterative process of extracting knowledge from multiple learning episodes, which typically encompass a diverse range of related tasks\cite{hospedales2021meta, huisman2021survey}. This accumulated experience is then utilized to enhance the performance in few shot setting, where only a few training data are avaliable. Termed "learning-to-learn," meta-learning offers various advantages, including improved efficiency in handling data and computational resources. Moreover, it exhibits a closer alignment with the learning processes observed in humans and animals, where strategies for acquiring knowledge evolve and improve over an individual's lifetime and even across evolutionary timescales.

In meta-learning, a task consists of a support set and a query set, which correspond to the training set and test set in traditional machine learning. In each task, the objective is to optimize the model using the support set and make predictions on the query set. In meta-learning, the training and test sets are composed of various tasks. And the goal of meta-learning is to achieve good performance on the query set using only a small support set during testing. Meta Learning has been widely applied in few shot settings of various domains such as computer vision\cite{sung2018learning, finn2017model, snell2017prototypical}, natural language processing\cite{qian2019domain, madotto2019personalizing}, and recommendation systems\cite{lee2019melu,lu2020meta}. Meta Learning enables training with a small number of samples when faced with new tasks, reducing the reliance on large-scale datasets. Consequently, it reduces the energy consumption associated with data collection and storage. As meta learning models achieve satisfactory performance within fewer training iterations, they are capable of reducing training time and computational resource consumption. Meta Learning is commonly classified into three distinct categories\cite{hospedales2021meta, huisman2021survey}, including optimization-based methods, model-based methods, and metric-based methods.

\textbf{Optimization-based methods}
Optimization-based methods is a popular category of Meta-learning. In Optimization-based meta-learning, there are two levels of learning: the inner or base learning algorithm and the outer or meta learning algorithm. During base learning, the inner algorithm tackles a specific task, such as classifying images, using a given support set and objective. The goal of Optimization-based meta-learning is to update the inner algorithm through the outer algorithm, improving its performance on new task in test set. This objective could be related to generalization performance or the speed at which the inner algorithm learns.

An influential work is the Model Agnostic Meta Learning (MAML) approach\cite{finn2017model}. MAML focuses on acquiring an initialization of parameters of a machine learning model, which enables the model to achieve favorable performance on new tasks within only a few iterations of inner update. Apart from learning the initial parameter values, certain models endeavor to learn the length of step\cite{antoniou2018train} or learn a recurrent networks to replace conventional optimizers\cite{andrychowicz2016learning, ravi2016optimization, li2016learning}. These approaches aim to enhance the optimization process and potentially improve the overall performance in few shot setting. 

\textbf{Model-based methods}
Model-based meta learning aims to incorporate current task into the state of a model. This model is then used to make predictions for test data based on the embedded task. To elaborate further, in model-based meta-learning, when a model is confronted with a task, it processes the task's training data in a sequential manner. At each step, an input is fed into the network, modifying the model's state. This state acts as a storage for task-specific information, enabling the network to make predictions for new inputs by leveraging the accumulated knowledge. Typically, this state is stored in a memory component of the model. Because the internal reasoning process is not directly observable, model-based meta-learning is often called black-box meta-learning. The Neural Turing Machine (NTM) firstly introduced external memory to increase the capacity for storing information of Long Short-Term Memory (LSTM)\cite{graves2014neural}. Then, \cite{mishra2017simple} present a unique blend of temporal convolutions and soft attention. Temporal convolutions are employed to gather information from previous experiences, while soft attention is used to precisely identify specific pieces of information. Moreover, hypernetwork is often applied to model-based meta learning\cite{qiao2018few,gidaris2018dynamic}. In contrast to optimization-based approaches, model-based methods offer the advantage of simpler optimization that does not require second-order gradients. However, it has been observed that model-based approaches tend to have lower generalization capabilities when it comes to out-of-distribution tasks compared to optimization-based methods.

\textbf{Metric Learning}
Metric learning, which is also known as non-parametric algorithms, is another type of meta learning. In this approach, the goal is to learn a metric or distance function that can effectively compare and match validation data points with training data points. By leveraging the similarity between the validation and training points, these algorithms can make predictions based on the labels of the matching training points. This approach has shown promise in the few-shot learning setting, where the available labeled data for each specific task is limited. The Siamese network was the initial work that introduced the concept of making predictions by comparing inputs with the training data of each task\cite{koch2015siamese}. Subsequently, the Matching network was proposed, which was trained in a few-shot setting and utilized cosine similarity\cite{vinyals2016matching}. The Prototypical network further enhanced robustness by comparing each input with a class prototype instead of individual training data\cite{snell2017prototypical}. Relation networks took a step further by replacing the fixed similarity metrics with a neural network, which allow for learning a domain-specific similarity function\cite{sung2018learning}.

These metric-based techniques offer several key advantages. Firstly, the concept of similarity-based predictions is straightforward and intuitive. Secondly, these techniques can be computationally efficient during test-time, especially when dealing with small tasks, as the networks don't require task-specific adjustments. However, a limitation arises when tasks in the meta-test phase differ significantly from those in the meta-train phase. In such cases, metric-learning techniques struggle to incorporate new task information into the network weights, leading to potential performance degradation.

\textbf{\numberedParagraph{Data Augumentation}}

Data augmentation alleviates the issue of insufficient data in deep learning. It was first widely used in the image domain and later extended to the NLP field, achieving results in many tasks. A primary direction is to increase the diversity of training data, thereby enhancing the model's generalization capability. There are several common techniques for data augmentation in NLP. First, Synonym replacement is a simple and efficient method, which replacing words in a sentence with their synonyms while preserving the original meaning. It helps the model learn different ways to express the same concept. Second, back translation translates the text to a different language using a machine translation model. Then, the translated text is translated back to the original language using another machine translation model. Third, We can add noise to the original text for data augmentation. This method involves adding, deleting, or swapping words or characters in a sentence to create new examples. This helps the model learn to be more robust to noise and spelling errors. In addition, there are many other approaches, such as text paraphrasing and rule-based transformations.

In computer vision (CV), data augmentation techniques are also widely used to increase the amount and diversity of training data, which can help improve the performance and generalization of models. For example, images can be flipped horizontally or vertically to create a mirror image, simulating objects being viewed from different angles; Images can be scaled up or down, simulating objects of different sizes or captured at different distances; Random noise can be added to the images, simulating sensor noise or other disturbances in the image capturing process; the colors in the images are slightly altered to simulate variations in lighting and object color.

\textbf{\numberedParagraph{Data Sampling}}

Data Sampling is an important method in deep learning. Some common Data Sampling methods include: Random Sampling, Stratified Sampling, Batch Sampling, Oversampling, Undersampling, and Adaptive Sampling. Oversampling and Undersampling are primarily used to handle imbalanced data situations. Oversampling is typically achieved by randomly duplicating samples from the minority class, while Undersampling is achieved by randomly deleting samples from the majority class. SMOTE \cite{chawla2002smote} is a special type of oversampling method, which interpolates the samples of the less numerous classes, generating new samples to increase the number of samples. Adaptive Sampling is a method that dynamically adjusts Data Sampling strategies based on the current performance of the model or the characteristics of the data. 

Data Sampling has the following five advantages:
First, in real life, we often encounter large-scale datasets. Using all data for training requires a large amount of computational resources. Through Data Sampling, we can select a subset from the dataset for training, such as Random Sampling and Batch Sampling, which can significantly reduce computational burden and memory usage, making it possible to handle large-scale datasets.
Second, Data Sampling can handle imbalanced datasets, where the number of samples for certain categories far exceeds others. This could lead the model's predictions to be biased towards the larger categories. Through Oversampling or Undersampling, we can adjust the number of samples in each category, allowing the model to better learn the features of all categories.
Third, Random Sampling or Batch Sampling can randomly select a batch of samples for training in each training cycle, increasing the randomness during the model training process. This helps prevent model overfitting and improves the model's generalization ability.
Fourth, through Batch Sampling, vectorized operations and parallel computing can be utilized to improve training efficiency. At the same time, by using different batches in each training cycle, model parameters can be updated more stably, thereby improving the convergence speed of the model.
Fifth, Data Sampling can be used to implement specific training strategies. For example, when training object detection or recognition models, using adaptive sampling strategies, such as Hard Negative Mining, i.e., prioritizing the selection of negative samples that the model finds difficult to recognize for training, can further improve the model's performance.

These data sampling methods are not mutually exclusive and often multiple methods are combined in practical applications. For example, when dealing with imbalanced datasets, Oversampling may be carried out first, followed by Random Sampling or Batch Sampling to select training samples.

\subsubsection{No-Labeled Data}

\textbf{\numberedParagraph{Transfer Learning}}

Transfer learning, a machine learning technique, has proven to be a solution to issues related to scarcity of labeled data and limited training data. This paper explores the concept of transfer learning, its application, and its effectiveness in scenarios with no-labeled data or less training data. In traditional machine learning, models are trained from scratch, which requires a large amount of labeled data and extensive computational resources. However, obtaining a large quantity of labeled data can be challenging, time-consuming, and expensive. 

Transfer learning, first introduced by Pan and Yang \cite{pan2009survey}, offers a solution to this problem by leveraging the knowledge gained from solving one problem and applying it to a different but related problem. For instance, a model trained on ImageNet, a large dataset of general images \cite{deng2009imagenet}, can be repurposed for classifying specific types of images, such as dog breeds. The initial layers of the model, which have learned to detect edges, shapes, and textures from the general images, can be reused, and only the final layers need to be retrained on the new task. This significantly reduces the amount of labeled data and training time required. Transfer learning has been particularly successful in the field of deep learning. Pre-trained models like VGG16 \cite{simonyan2014very}, ResNet \cite{he2016deep}, or BERT \cite{devlin2018bert} have been used to achieve state-of-the-art results on a wide range of tasks. These models, trained on large datasets, have learned rich feature representations, which can be effectively transferred to new tasks with smaller datasets. 

In the context of no-labeled data or less training data, transfer learning can be a powerful tool. When we have no-labeled data, we often resort to unsupervised learning methods. However, these methods can be challenging and less effective than supervised learning methods, which use labeled data. By using a pre-trained model through transfer learning, we can leverage the knowledge the model has already gained from its original training on a large dataset, even if we don't have any labeled data for our specific task \cite{yosinski2014transferable}. Similarly, when we have less training data, a model trained from scratch might overfit to the training data, leading to poor performance on unseen data. However, by using transfer learning, we can avoid over-fitting, as the model has already learned general features from the large dataset it was originally trained on. We only need a little data to fine-tune the model for specific task \cite{shin2016deep}.

In conclusion, transfer learning is a powerful technique that can overcome the limitations of traditional machine learning methods in scenarios with no-labeled data or less training data. By leveraging the knowledge gained from pre-trained models, transfer learning allows us to build effective models with less data and computational resources, thereby opening up new possibilities for a wide range of applications.

\textbf{\numberedParagraph{Self-Supervised Learning}}

Self-supervised learning (SSL), as a branch of unsupervised learning, has received increasing attention in recent years due to its excellent performance. Generally speaking, self-supervised learning has the following two properties \cite{liu2021self}: 1) Obtain labels from the data itself; 2) Predict part of the data from other parts. In other words, SSL enables the model to learn complete information from parts of the data itself, which allows the model to be trained on large-scale unlabeled datasets. Due to this property, in computer vision field, SSL methods have been able to match and in some cases outperform supervised models, even in highly competitive benchmarks such as ImageNet \cite{he2020momentum, tomasev2022pushing}. Similarly, in natural language processing field, large language models (LLMs) \cite{brown2020language, ouyang2022training, zeng2022glm} have comprehensively outperformed traditional models, while their training are all based on SSL.

The intuition for self-supervised learning is to utilize the co-occurrence relationships inherent in the data as self-supervision. For example, if we randomly remove a small part of the image, it does not affect our understanding of the whole image since we can guess the vacant part by the remaining part, which is the basic idea of MAE \cite{he2022masked}. The mainstream SSL works can be summarized into two main categories: Generative and Contrastive.

\textbf{Generative} Generative methods attempt to train an encoder to obtain hidden features $z$ of the input $x$, while the original data $x$ can be recovered as closely as possible via a decoder based on features $z$. Auto-encoding (AE) Model is one of the main types in Generative methods, which focuses on the reconstruction of the original input from corrupted inputs. The well-known models VQ-VAE \cite{van2017neural}, BERT \cite{devlin2018bert}, ERNIE (Baidu) \cite{sun2019ernie} all belong to AE models. Another important branch is the Auto-regressive (AR) Model, which typically aims to maximize the likelihood under the forward autoregressive factorization \cite{yang2019xlnet}. Common AR models include GPT series \cite{radford2018improving, radford2019language,brown2020language}, PixelCNN \cite{van2016conditional}, GCPN \cite{you2018graph}, etc. There are also some efforts to try to combine the benefits of AR and AE. For example, XLNet \cite{yang2019xlnet} introduces Permutation Language Model (PLM), which implements bidirectional contexts learning by maximizing the expected likelihood of all permutations of the factorization order; GLM \cite{du2021glm} propose autoregressive blank infilling task with 2D positional encodings to enable an arbitrary order of predicting spans, which cleverly combines the structure of AE and AR.

\textbf{Contrastive} Contrastive methods belong to the discriminative model, which means they only need to train the encoder to encode the input $x$ into the hidden features $z$, and do not need to reconstruct the feature $z$ back to $x$. Contrastive methods learn by comparing the feature similarity of two inputs, so we can categorize them into global-local and instance-instance according to the type of inputs. The global-local contrastive methods focus on modeling the attribution between the local features of a sample and its global contextual representation. In computer vision, we can infer the relative position between two patches \cite{doersch2015unsupervised}, or disrupt the patches to rearrange them \cite{kim2018learning, noroozi2016unsupervised, wei2019iterative}. In natural language processing, similar tasks are Next Sentence Prediction (NSP) from BERT \cite{devlin2018bert} and Sentence Order Prediction (SOP) from ALBERT \cite{lan2019albert}. Unlike global-local contrastive, instance-instance contrastive methods compares the features of different samples. InstDisc \cite{wu2018unsupervised} proposes the instance discrimination task which considers each image as a special category and the goal of SSL is to find feature space that can distinguish all images; CMC \cite{tian2020contrastive} proposes to adopt multiple different views of an image as positive samples; MoCo \cite{he2020momentum} solves the memory limitation and substantially increases the number of negative samples in each batch; SimCLR \cite{chen2020simple} introduces 10 data augmentation (crop, rotate, cutout, etc.) methods to construct positive samples. Different from the previous work, BYOL \cite{grill2020bootstrap} discards all negative samples and only utilizes positive samples, and surprsingly they still can achieve considerable performance.

Overall, self-supervised learning is gradually becoming the cornerstone of deep learning, with its Generative approaches supporting the large language models of NLP, and its Contrastive approaches are leading the CV to get more and more breakthroughs.

\subsection{Hyper-Parameter Optimization}

%

This section aims to give a brief review of hyper-parameter optimization (HPO) from the aspects of search space, search methods, evaluation methods, and toolkits. For a more comprehensive review of HPO, one can refer to~\cite{HPOsurvey,BischlBLPRCTUBBDL23}. In this recap, we would like to emphasize the green issue of HPO.

\subsubsection{Search Space}
Given constraints on computational resources during hyper-parameter optimization, hyper-parameters that exert greater influence on model performance typically receive preferential treatment in the tuning process. According to prior work, hyper-parameters with a stronger effect on weight updating over the course of training tend to be more influential for neural network learning~\cite{Youngjun2019Hyperparameter}. However, quantitatively determining the relative significance of each hyper-parameter on final predictive accuracy remains challenging. Generally speaking, hyper-parameters perceived as more important based on previous empirical studies and practitioner experience are more thoroughly investigated in HPO research. Hyper-parameters can be roughly categorized into two groups - those related to the model architecture design (e.g., the number of layers) and those governing the training procedure (e.g., learning rate). A systematic understanding of hyper-parameter importance across different classes could guide more sample-efficient search strategies.
It is worth noting that we only outline some exemplary and influential hyper-parameters within two broad categories. However, HPO may involve more hyper-parameter types or customized hyper-parameters beyond what will be described below.


\textbf{\numberedParagraph{Hyper-Parameters of Model Architecture}}

The number of hidden layers is a critical hyper-parameter for determining the overall architecture of neural networks, as it directly impacts the representational capacity and final outputs~\cite{Geoffrey2006DBN}. Deeper learning models with additional hidden layers are more capable of fitting complex patterns in data and generally achieve higher predictive accuracy. A common approach to improving performance is to iteratively increase network depth during hyper-parameter tuning. In doing so, a baseline structure can be repeated to expand the receptive field in a computationally efficient manner. For instance, practitioners may choose from models like ResNet-18 to ResNet-200 based on their accuracy requirements given constrained resources~\cite{He2016ResNet}. 
The number of neurons in each hidden layer also requires thoughtful tuning. Insufficient nodes may result in underfitting as the model lacks representational power~\cite{salam2021effect}, while too many neurons could lead to overfitting and increased training time.
In~\cite{Tan2019EfficientNet}, it demonstrates that constructing a series of networks with systematic variations in depth and width during hyper-parameter optimization can yield high performance using limited parameters and float operations. The findings suggest that jointly tuning architectural hyper-parameters like depth and width, rather than single hyper-parameters independently, may assist the search in escaping shallow local optima and discovering neural architectures with accuracy-efficiency trade-offs. 


In contrast to increasing network depth or width, regularization~\cite{goodfellow2016deep} is commonly employed to reduce model complexity, especially for datasets with limited samples. Overfitting is prone to occur in deep neural architectures with many layers and nodes. Regularization helps address this by incorporating additional terms that constrain network weights during optimization. One widely-used approach is to add some kind of regularization penalty to the loss function in order to select more salient features and prevent overfitting. This drives weights towards smaller values to induce smoother decision boundaries. Compared to simply expanding network dimensions, regularization provides an effective means of controlling model capacity for improved generalization performance.

Dropout~\cite{goodfellow2016deep} is another widely-used regularization technique in deep learning. It randomly disables (or drops out) neurons from the neural network during training, preventing co-adaptation across neurons~\cite{Liang2021R-Drop}. As such, dropout helps reduce overfitting by making the model less reliant on specific node weights. At each training iteration, individual neurons are dropped out with some probability, resulting in a simplified network architecture. This has the effect of reducing complex co-adaptations in the original network. Generally, a dropout rate of 20-50\% is recommended, with smaller rates having less regularization effect and larger rates risking underfitting. 
Additionally, when using dropout, a higher initial learning rate accompanied by exponential decay is often beneficial, as are larger momentum values. This is because dropout effectively thins the network during training, meaning fewer parameter updates per batch on average. Larger stepsizes and momentum help counteract this thinning and maintain adequate weight updates. Together, dropout provides a straightforward and effective means of controlling model complexity without requiring changes to the basic model configuration or training algorithm.

Activation functions~\cite{goodfellow2016deep} play a critical role in deep learning by introducing non-linearity into the relationship between the input and output of neurons. Without an activation function, a neural network would simply perform linear learning and be unable to model complex, nonlinear patterns in data. Ideally, activation functions should be differentiable to support efficient computation of gradients during backpropagation. 
Some of the most widely-used activation functions include the sigmoid, hyperbolic tangent (tanh), rectified linear unit (ReLU)~\cite{Nair2010ReLU}, Maxout~\cite{Tseran2021Maxout}, and Swish~\cite{Mercioni2022Swish}. The sigmoid and tanh are historically significant but can potentially lead to the vanishing gradient problem. ReLU helps address this and has become a standard choice, but could cause dead neurons. Maxout and Swish aim to improve upon ReLU. 
Automatic search techniques have also been applied to optimize activation function structure and hyper-parameters empirically~\cite{Ramachandran2018AF}. Since activation functions play a fundamental role in modeling capabilities, choosing the most appropriate type for a given task remains an important component of neural architecture engineering and hyper-parameter optimization. Further research on adaptive, task-specific activation functions may help unlock even greater representational power in deep learning models.

\textbf{\numberedParagraph{Hyper-Parameters of Model Training}}

Learning rate~\cite{goodfellow2016deep} is a positive scalar which determines the step size during gradient descent. Learning rate often needs to be adjusted over the course of training to enhance model performance. 
It is common to implement a learning rate schedule that varies the rate dynamically. For example, learning rate may be decayed by a constant factor every few epochs~\cite{Bengio2012GBT}. Adaptive methods also aim to automatically adjust learning rate in response to training progress or model structure via specialized learning algorithms~\cite{Smith2017CLR}. Other schedule hyper-parameters include the decay floor and the number of epochs between decays. Traditionally, learning rate may halve every 10 epochs until a floor is reached~\cite{Tan2019EfficientNet}. More sophisticated schedules like exponential decay can yield better results, as demonstrated by models like EfficientNet~\cite{Tan2019EfficientNet} trained on large-scale datasets such as ImageNet~\cite{Krizhevsky2017ImageNet}.


The choice of optimizers strongly influences training efficiency and performance. In~\cite{sun2020opt}, the survey gives a comprehensive overview of optimizers for deep learning. To name a few, common choices include stochastic gradient descent, mini-batch gradient descent, gradient descent with momentum, AdaGrad, RMSprop and Adam, etc. 
The hyper-parameters of optimizers involve the type of algorithms, mini-batch size, using momentum or not, decay rates and the like. Given a task, searching a tailored configuration for optimizers has received increasing attention~\cite{AndrychowiczDCH16,ChenHCDLBF17}, and it is of great significance for green AI under the requirement of relatively low cost.

\subsubsection{Search Methods}



Once the search space is defined, the next step is to find a suitable optimizer to guide the search within the search space. HPO aims to automatically explore the hyper-parameter landscape using specific evaluation criteria to find the optimal combination of hyper-parameters. We roughly divide the search methods for HPO into two categories: learning-free HPO methods and learning-based ones. Learning-free optimization methods such as random search are simple to implement, but they could be hard to meet the efficiency and effectiveness requirements due to the complex search space and intricate mapping between the search space and the performance indicator. To address the complex search space and implicit mapping, the learning-based HPO methods by the way of learning to optimize~\cite{ChenCCH0WY22} are introduced. In general, most of the learning-based optimization methods begin with some sampled and evaluated solutions, and then follow the loop of explicitly or implicitly learning/updating a model from the sampled solutions as well as their objective function values and sampling solutions from the model.


\textbf{\numberedParagraph{Learning-Free HPO Methods}}

Manual Tuning. Central to the human tuning process is the practitioner's expertise. Through keen observation and nuanced analysis of the model's behavior, skilled individuals can intuitively adjust hyper-parameters, driving enhanced model performance. Once the learning problem is defined, one can determine the initial/iterative hyper-parameters based on their previous experience and domain knowledge. In order to achieve better learning performance, one attempts to train and evaluate the model using the current configuration. Based on the feedback from the model, adjustments to the hyper-parameters are made to improve the performance. The iterative process stops when a certain level of performance is achieved or when the computational budget is exhausted.

Grid Search~\cite{dean1999design}. Grid search systematically searches for the best combination of hyper-parameters within a predefined space. In grid search, it first specifies a finite set of values for each hyper-parameter. These values are combined to form a grid of hyper-parameter combinations, with each combination representing a point in the search space. Subsequently, the model is trained and evaluated using the Cartesian product of these sets. Ultimately, by traversing this grid, one can identify a good hyper-parameter combination under the given evaluation metric. The advantage of grid search is its simplicity and intuitiveness for HPO. It exhaustively examines all hyper-parameter combinations in the search space to find a globally optimal one. As a deterministic method, grid search is void of randomness, ensuring reproducible results. However, grid search suffers from the curse of dimensionality. The necessary evaluations increase exponentially with the increasing dimensionality of search space, and enhancing the resolution of discretization significantly escalates the number of required evaluations.

Random Search~\cite{james2012random}. As opposed to grid search, random search does not exhaustively explore all possible hyper-parameter combinations. Instead, it randomly selects hyper-parameter values, and trains and evaluates the model accordingly. The algorithm initiates from a random point, and progressively selects subsequent random candidates until it meets the optimization objective or the pre-defined stopping criterion. Random search is commonly used in handling high-dimensional search space~\cite{james2012random}. The drawback of random search is its randomness. Random search does not leverage the inherent structure within the hyper-parameter space, possibly oversampling in certain regions while paying insufficient attention to others. 




\textbf{\numberedParagraph{Learning-Based HPO Methods}}

Bayesian Optimization (BO)~\cite{shahriari2016taking}. Bayesian optimization is a sample-efficient optimization framework commonly used for global optimization of expensive black-box functions. BO has two key components: a surrogate model to learn the underlying objective function and an acquisition function to determine which solution should be sampled. Widely-used surrogate models include Gaussian processes~\cite{Jasper2012PracticalBO, Martin2018SGPT}, deep neural networks~\cite{Jasper2015DNGO}, trees~\cite{BergstraBBK2011nips} and random forests~\cite{Frank2011SMAC}, etc. Commonly-used acquisition functions include probability of improvement, expected improvement and upper confidence bound~\cite{shahriari2016taking}, etc. 
In each iteration, the surrogate model makes full use of the information contained in all the observed results of the objective function and tries to fit it. The acquisition function uses the predicted distribution of the probabilistic model (if Gaussian processes are utilized) to determine the utility of different candidate solutions, striking a balance between exploration and exploitation to identify the next solution to be sampled. The new sampling solution and its objective function value are added to the dataset, and the probabilistic model is updated. This process continues until the most promising set of hyper-parameters is found or the resources are exhausted. 
However, BO has certain limitations. Typically, it suffers from poor scalability in high-dimensional search space. Recently, this limitation has been mitigated by methods such as random embeddings~\cite{Ziyu2016REMBO,zhang2023pedbo}, additive models~\cite{Kirthevasan2015AddGPUCB}, and others~\cite{eriksson2019scalable}.

Bandit-Based Methods. HYPERBAND~\cite{Lisha2017Hyperband} is a representative bandit-based method used to accelerate stochastic search to efficiently optimize hyper-parameters of machine learning models. Via competitive training and screening, it is a favored method for hyper-parameter optimization. HYPERBAND strikes an effective balance between the number of hyper-parameter combinations and the resources each hyper-parameter group can access. It dynamically allocates more resources to high-performing hyper-parameter configurations, leveraging the successive halving~\cite{jamieson2016non} technique to achieve this. Since HYPERBAND relies on random search, it suffers from relatively slow convergence rate to the best configurations. To address this, BOHB~\cite{Stefan2018BOHB} integrates Bayesian optimization with HYPERBAND to achieve the best of both worlds, i.e., strong anytime performance and final performance. Specifically, BOHB substitutes a model-based search for the random selection of configurations in the early stage of each HYPERBAND iteration. Given the constructed model, Bayesian optimization is utilized to choose a new configuration.





Evolutionary Algorithms (EAs)~\cite{Thomas1996Evolutionary}. EAs are a kind of population-based optimization methods inspired by the process of natural evolution. They are capable of addressing complex optimization problems (e.g., non-differentiable functions) in machine learning, which yields evolutionary learning~\cite{Yao99,zhou19EL,Jin23EDL}. Representative EAs include but not limited to genetic algorithms, evolution strategies, evolutionary programming, particle swarm optimization and differential evolution. Although EAs have various implementations, most of them can be roughly summarized as follows. (a)~Randomly generating an initial population of solutions. (b)~Reproducing new solutions on the basis of the current population. (c)~Removing relatively low-quality solutions in the population. (d)~Repeating from Step (b) until a stop criterion is met. A population of solutions is maintained and evolved in the evolutionary process, and therein mutation and crossover are two widely-used variation operators for reproducing new solutions in Step (b). Mutation slightly alters a solution to generate a new solution. Crossover refers to the combination of two different solutions in some way to create a new solution. A fitness function on the basis of the objective function is used to guide Step (c) and push the evolutionary process forward. The population-based search nature of EAs makes them friendly to parallelization.

In addition to the aforementioned general-purpose search methods for HPO, other advanced learning techniques such as meta-learning have also been adopted to boost the methods of learning to optimize. Meta learning based HPO~\cite{RijnH18kdd,Zimmer2021AutoPytorch,FeurerEFLH2022jmlr} aims to search good configurations across different datasets or tasks. via meta-learning, knowledge can be transferred among datasets or tasks, and it is useful for obtaining priors for a good warm start. Furthermore, under the scenario of large-scale data, parallelization and distributed computing~\cite{jun2023anttune} is very practical to save HPO running time. 
Population-based training (PBT)~\cite{jaderberg2017population,Ang2019PBTFramework} is an asynchronous and decentralised framework that aims to simultaneously optimize a population of model parameters and their hyper-parameters. PBT bridges parallel and sequential search methods and is able to share information across concurrently running optimization processes. PBT possesses the merit of effectively utilizing a fixed computational budget and high efficiency. 
In the future, it is expected that more search methods tailored to different green HPO requirements will be proposed and developed.


\subsubsection{Evaluation Methods}

Given a learning task as well as a dataset, a direct way of evaluating a hyper-parameter configuration involves two steps. At first, a model is fully trained on the entire training set under this hyper-parameter configuration and the corresponding optimal parameters are learned. Then, the quality of this hyper-parameter configuration is assessed on the entire validation set under the aforementioned learned parameters through computing a certain performance indicator to estimate the generalization ability, e.g., accuracy on the entire validation set for a binary classification task. To obtain a low-variance estimator, these two steps are often accompanied by $K$-fold cross validation. Although accurate, this direct way is computationally expensive, time consuming and sometimes contravenes the principle of green AI when it encounters big learning model or big data. 

To reduce the cost of evaluation and follow the rule of green HPO, alternative ways are proposed in recent years. The hyper-parameter configuration quality evaluation process is essentially seeking a balance between accuracy and cost. When a model has less training time and thus is not fully trained, the accuracy of evaluation could decrease and variance could increase. A longer training time can result in a more well-prepared model, but it also significantly extends the training process and requires more resources. In practice, for large models, it is crucial to utilize fast evaluation methods that can save unnecessary training time and computational resources. Besides, to evaluate a certain hyper-parameter configuration, the sample size used for training and validation also plays a significant role in accurate assessment and efficiency trade-off. 

Multi-fidelity evaluation is one representative of the alternative ways. The high-fidelity evaluations are relatively accurate but costly, while low-fidelity ones are relatively cheap but noisy or not so accurate. To construct different fidelities, we can utilize subsets of a dataset with different sizes or train a learning model with different epochs. Since the function from hyper-parameter configurations to their performance usually cannot be expressed implicitly and this function becomes expensive to be evaluated with the increasing size of datasets and the growing complexity of models, black-box optimization is cooperated with multi-fidelity evaluation to substantially reduce its cost and boost efficiency while maintaining its efficacy, which yields multi-fidelity optimization~\cite{Kandasamy2016Multifidelity,Takeno2020Max-value,Li2021Batchmulti}. A well-known generic test suit of it can be found in~\cite{WangJD18tec}. 
Generally speaking, multi-fidelity optimization for HPO~\cite{Hu2019Multi,Zimmer2021AutoPytorch,qian2023greenDFO} uses cheap but relatively low fidelity function value evaluation proxies to approximate the true function value of the aforementioned black-box expensive function. To find the optimal hyper-parameter configuration under limited resources, low-fidelity evaluations are used for rapid exploration of the search space so as to rule out the low-quality configurations, wheres high-fidelity evaluations refine the search and mainly need to focus on the potential high-quality configurations. Switching mechanism among different fidelities is vital which needs to be elaborately designed, which enables that different fidelities perform their respective duties and cooperate with each other.

Other approaches~\cite{yao2018takinghumanout} to efficient evaluation include sub-sampling, early termination, parameter reuse, and data-driven offline optimization, etc. Sub-sampling method trains the parameters using only a subset of the training data~\cite{KleinFBHH17aistats}. Generally, the smaller the training data, the faster the evaluation speed. Early termination allows an evaluator to terminate training and report low performance when poor performance is observed early on, indicating that the candidate configuration's performance is not satisfactory~\cite{DomhanSH15ijcai}. While it can reduce running time, noise and bias are introduced, as poor early performance does not necessarily imply the impossibility of finding the optimal solution. Parameter reuse refers to utilizing previously evaluated model parameters to warm-start the current model. These model parameters could serve as a good starting point for training or may lead to convergence to a local optimum~\cite{sutskever2013NAGwithCM}. Data-driven offline optimization~\cite{JinWCGM19tec,TrabuccoKGL21icml,lu2023degradation} is tailored to the scenario where true evaluation is unavailable but offline historical data resources can be leveraged. Data-driven offline optimization can complete HPO tasks without incurring high model evaluation cost, thereby helping facilitate green HPO.

\subsubsection{HPO Toolkits}

Recently, various green HPO toolkits have been developed to reduce the burden on researchers and engineers of implementing HPO in their own training and inference process. To name a few, these toolkits include HEBO~\cite{cowen2022hebo}, ZOOpt~\cite{Liu2022zoopt}, FLAML~\cite{chi2021flaml} and Auto-sklearn 2.0~\cite{FeurerEFLH2022jmlr}, etc.
%


HEBO implements a series of heteroscedastic evolutionary Bayesian optimization methods for sample-efficient green HPO~\cite{cowen2022hebo}. Since even the simplest machine learning problems can raise heteroscedasticity and non-stationarity in HPO, HEBO facilitates the modeling of complex noise processes by input and output transformations~\cite{edward2003warped}. Through performing nonlinear input-output warping, HEBO enables exact marginal log-likelihood optimization while retaining robustness to the learned parameter values. Furthermore, individual acquisition functions used in Bayesian optimization may occasionally provide conflicting guidance, where local optima for one function occur at sub-optimal points for others. HEBO addresses this issue via a multi-objective formulation that seeks to discover the Pareto optimal front across objectives.

ZOOpt is an easy-to-use Python library that provides efficient HPO solvers based on model-based derivative-free optimization techniques~\cite{Liu2022zoopt}. Primal methods such as RACOS~\cite{yang2016derivative} employ classification models wherein hyper-parameters are classified as good or bad based on prior evaluations. This learned partition of search space guides future suggestions towards more promising regions, reducing the number of trials needed for green HPO. Besides, ZOOpt utilizes random embedding~\cite{Ziyu2016REMBO,hong2016derivative} to enable accelerated convergence for optimization in high-dimensional search space exhibiting low intrinsic dimensionality. To enable users to parallelize single-machine code, ZOOpt also adopts the Ray framework~\cite{philipp2018ray} and implements an efficient distributed optimization module.


FLAML is a lightweight, green and easy to customize toolkit to automatically choose learners and hyper-parameters~\cite{chi2021flaml}. It explores and utilizes the structure of search space to optimize a search order that well balances the trial cost and model error. As the optimization procedure proceeds, FLAML iteratively decides the learner, hyper-parameter, sample size and resampling strategy and, at the same time, makes full use of their compound influence on both computational cost and model error. In general, the optimization procedure is inclined to gradually move from cheap trials but inaccurate models to costly trials but accurate models.

Auto-sklearn 2.0~\cite{FeurerEFLH2022jmlr} is a hands-free automated machine learning (AutoML) toolbox. Auto-sklearn 2.0 includes two key parts, i.e., portfolio successive halving in PoSH Auto-sklearn and automating AutoML in Auto-sklearn 2.0. In the first part, instead of always using full budget to assess the quality of a configuration, the budget allocation strategy successive halving is also suggested as a complementary. The purpose behind this strategy is to allocate more resources to promising configurations. In the second part, to automatically search the best setting of the AutoML system for a given learning task as well as a dataset, it proposes a meta-learning method on the basis of algorithm selection. Overall, Auto-sklearn 2.0 can perform well on large-scale datasets given a limited time budget.


\section{Energy-Efficient Inference}
\label{Sec:energyEfficientInference}

\subsection{Model Pruning}

In order to achieve green AI, it is necessary to reduce the energy expenditure of neural networks.
Model pruning is an effective method that can reduce the complexity and computational requirements of neural networks, thus lowering energy consumption. By analyzing the connection weights between neurons and removing unimportant connections, pruning can significantly reduce the number of parameters and computational workload of the model. As a result, the computational resources required by the neural network during inference and training are greatly reduced, leading to reduced energy consumption.
Model pruning can also contribute to improved hardware efficiency. For neural networks running on edge devices or embedded systems, pruning can help reduce memory usage and computational requirements, enabling these devices to efficiently perform inference tasks while reducing energy consumption and heat generation.
Through model pruning, we can significantly reduce the energy expenditure of neural networks and achieve more environmentally friendly and sustainable green AI applications. The widespread application of this technique helps to drive the field of artificial intelligence towards a more sustainable and environmentally friendly direction.

It's important to note that the energy reduction achieved by pruning is highly dependent on the specific pruning method, the target pruning rate, and the hardware implementation. Different neural network architectures and datasets may exhibit varying levels of sensitivity to pruning. Therefore, it is recommended to evaluate the energy savings on a case-by-case basis, considering the specific circumstances and requirements of the given scenario.
Overall, while there is no fixed percentage in terms of energy reduction for model pruning, researches\cite{wang2023progressive, marino2023deep} have shown that significant energy savings can be achieved through proper pruning techniques, leading to more efficient and sustainable AI systems.

\subsubsection{Pruning Unit Types}

\begin{figure}[h]
	\centering
	\includegraphics[width=0.6\linewidth]{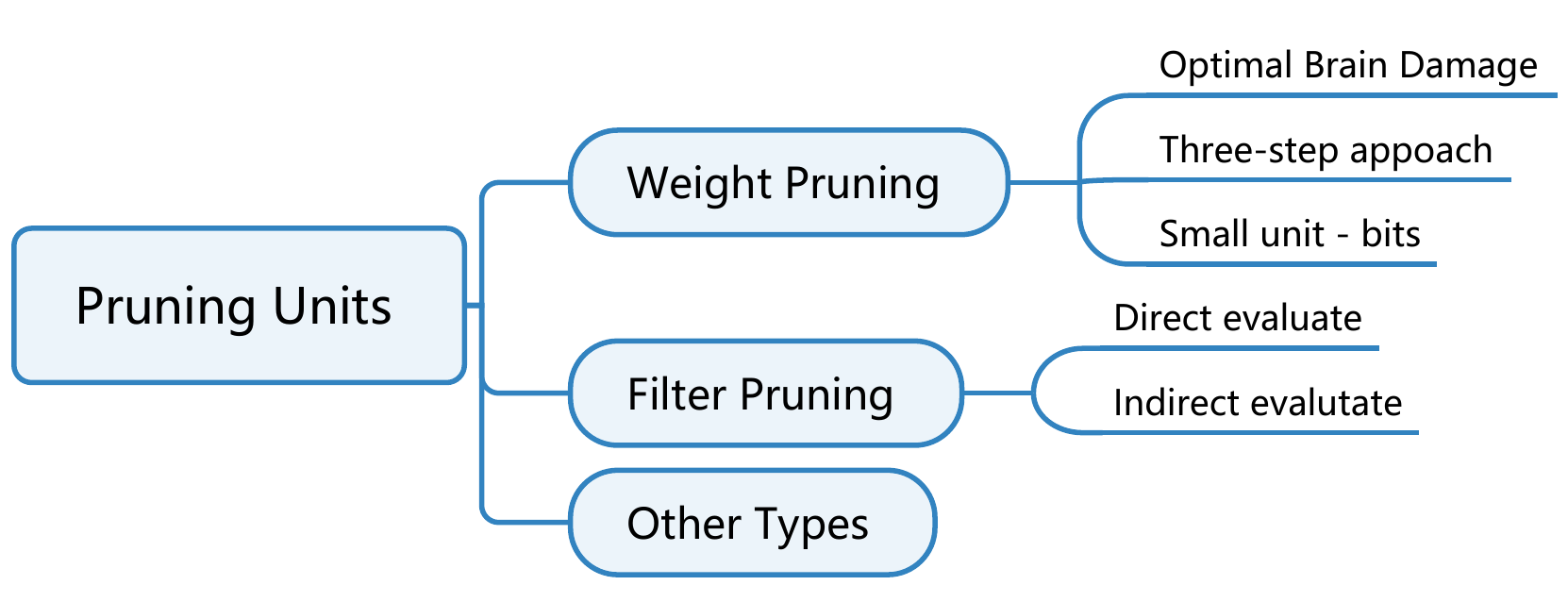}
	\caption{Types of model pruning units.}
	\label{fig:PruningUnits}
\end{figure}

Pruning units can be implemented at different granularities, including element-wise, row-wise, column-wise, filter-wise, or layer-wise. However, some related reviews \cite{cheng2018recent,he2023structured} provide excessive categorization of pruning units without thoroughly explaining the methods related to each subclass. To simplify the understanding, we adopts a simpler structure of development and focuses on three common forms of classification in Figure \ref{fig:PruningUnits}. It elaborates on the subclasses and key literature included in each pruning unit for easier and more intuitive understanding.

\textbf{\numberedParagraph{Weight pruning}} 

Weight pruning focuses on the basic weight components of Neural Networks. The concept of pruning individual weights was initially inspired by brain science and was first proposed by LeCun in 1990 \cite{lecun1990advances}, aiming to mimic the biological learning process in mammals. The Optimal Brain Damage (OBD) approach utilizes the second derivative (Hessian matrix) of the loss function. However, OBD has certain limitations due to its reliance on three default conditions: near-quadratic cost function, converged extremal pruning, and diagonal error by co-consequence. These conditions greatly restrict the pruning accuracy of the method. In contrast, subsequent optimization in OBS \cite{hassibi1993optimal} eliminates diagonal error by utilizing a non-diagonal Hessian matrix and calculates a significance score for each weight using backpropagation to compute the second-order differential information. This improvement allows for pruning ideas based on the update of the surgical response weights, resulting in improved accuracy of the XOR networks after pruning. However, it is vital to recognize that these methods only operate on the fully-connected layers and use Hessian-weighted distortion measure \cite{choi2016towards}.

Unlike the OBD idea, Srinivas et al. \cite{srinivas2015data} determined the optimal number of pruned neurons by minimizing the expected squared difference of logits and human observation of the pruning sensitivity curve. The pruning process is more rapid as it does not rely on training data and backpropagation to compute gradient information.

As the idea of measuring neuron connection weights gradually gained traction, Han et al. \cite{han2015learning} first proposed a three-step approach to removing pruning redundancy, i.e., analyzing the importance of neuron connections, remove unimportant connections, and retraining to fine-tune network weights. This staged pruning and retuning approach resulted in a 13-fold reduction of the VGG-16 network parameters on the ImageNet dataset without compromising accuracy.

Exploring each of these three stages is crucial for subsequent research. In the first stage of pruning, Hoang et al. \cite{hoang2022revisiting} argue that instead of the complete network, a sparse subnetwork can be found. They seek to establish the approximate relationship between the highly connected but sparse Ramanujan Graph and the subnetwork. Additionally, they accurately evaluate the naive randomness of the Ramanujan Graph location-found subnetwork using the Iterative Mean Difference of Bound (IMDB) assumed upper bound and normalized Random Coefficient (NaRC) assumed lower bound. The validity of Ramanujan Graph localization search for network subnets is demonstrated. This work takes into account the pseudo-randomness and irregular bi-graphs in practical sparse NNs for the first time.

In retraining step, Paul et al. \cite{paul2022unmasking} provide a thorough examination of iterative magnitude pruning (IMP) \cite{elesedy2020lottery,maene2021towards}, focusing on its impact on the loss geometry. The authors aim to improve our understanding of how pruned mini subnets are discovered through IMP. Unlike one-shot pruning methods, the IMP mask retains spatial information related to linearly connected modes, which have a low error barrier and yield similar sparse solutions. Furthermore, the study establishes a connection between the IMP mask and the Hessian eigenvalue spectrum, shedding light on the intricate relationship between these two components.

In addition to weight, some researchers also consider smaller units like bits. Maene et al. \cite{sekikawa2022bit} argue that the main issue to be considered for pruning is the multiplication (mult) in dot-product. They propose Bit-Pruning, which reformulates a dot-product between an integer weight and activation into an equivalent operation consisting of additions followed by bit-shifts. Bit-Pruning aims to reduce the energy consumption of additions followed by bit-shifts by removing unnecessary bits in each weight value during training. Compared to classical Weight-Pruning, Bit-Pruning demonstrates a better accuracy-energy trade-off. However, it requires appropriately optimized hardware and linear algebra library support.

Other scholarly endeavors investigate the pruning process through diverse lenses, as evidenced by the works of Wang et al. (2023) \cite{wang2023ntk}, Chen et al. (2023) \cite{chen2023unified}, and Peste et al. (2022) \cite{peste2022cram}. NTK-SAP \cite{wang2023ntk} delves into the realm of the neural tangent kernel (NTK) theory, focusing on the identification of connections that exert minimal influence on the spectral properties of the NTK. By contrast, Soft Threshold \cite{chen2023unified} introduces a score parameter associated with each weight, enabling the retention of weights with the highest sigmoid(score) values. CrAM \cite{peste2022cram} employs the inner maximization step derived from the Sharpness-aware minimization (SAM) technique to selectively retain stable parameters during the pruning process. Such interdisciplinary investigations contribute to a comprehensive understanding of pruning methodologies from various angles.

Weight pruning often involves unstructured pruning, which means deleting parameters at any position without changing the model structure. This type of pruning leads to increased sparsity, resulting in storage and computational overhead on the sparse matrix. In practical usage, deploying the pruned model on hardware becomes more complex and may not be applicable to general-purpose model architectures. It requires specific technical support to handle the computation of sparse matrices and specialized libraries like cuSPARSE.

\textbf{\numberedParagraph{Filter Pruning}}

To address the sparsity issue caused by non-structured pruning, researchers have proposed structured-based filter pruning methods. Filter pruning involves determining the contribution of filters, which are made up of multiple weights, to the current task. Filters with weak or no contribution are then removed. The evaluation of filter importance can be categorized as either direct or indirect.

The first instance of the filter-based pruning approach can be attributed to the Pruning Filter for Efficient ConvNets (PFEC) \cite{li2016pruning}, which employed the l1-norm method. The direct evaluation method assessed the significance of filters by computing the absolute values of all weights within the filters. The subsequent experiments conducted on the CIFAR-10 dataset revealed that a 34\% reduction in the number of filters resulted in a 0.75\% improvement in accuracy. However, it is crucial to note that assigning less importance to smaller norm values during pruning may not always be appropriate. 
Subsequently, Soft Filter Pruning (SFP) \cite{he2018soft} introduced the L2-Norm approach, which utilizes the square root of the sum of squares of individual elements in the weight vector. The L2 paradigm's smoothness and robustness render it insensitive to outliers, unlike L1. Additionally, a preference for dense solutions characterizes the L2 paradigm, wherein most elements remain non-zero. Furthermore, the L2 paradigm can measure vector similarity or distance.

He et al. \cite{he2019filter} examined a large number of the aforementioned approaches using L1 and L2 paradigm metrics, and determined that the filters' bigger nom deviation and lower minimum norm are the key causes for the algorithm's low efficiency. As a result, they suggested Filter Pruning by Geometric Median (FPGM) for filter assessment, which effectively distinguishes between redundant and low-importance filters.
ThiNet \cite{luo2017thinet} approaches the pruning formalism as an optimization issue, indicating that filter evaluation must be dependent on statistical information obtained from the next layer rather than the current layer. The strategy reduces the original VGG-16 model to 5.05 MB with just a 0.52 percent drop in accuracy.

Subsequent studies have indicated that merely calculating the geometric median of a single filter is insufficient to account for the replacement of filters with similar effects within the same layer \cite{liu2017learning, wang2019cop}. To address this limitation, Correlation-based Pruning (COP)\cite{wang2019cop} evaluates the variability of filters across layers. COP globally assesses the effectiveness of all filters in the network simultaneously by normalizing them on a layer-by-layer basis using the Pearson correlation test and scaling the different layers to a consistent range.

Some methods evaluate the activation values by computing the differences of the input data instead of directly evaluating filters with small weights. This approach allows them to determine the importance of all filters in the channel corresponding to the activation value for various domain tasks. This type of filter pruning, also known as channel pruning, not only impacts the number of activation maps generated in the current layer but also the number of filters in the subsequent layer.Average Percentage Of Zeros (APoZ) \cite{hu2016network} is a typical method used to determine the activation map obtained after filter computation in order to select channels suitable for pruning. Specifically, when the activation function of Rectified Linear Unit (ReLU) is used, the activation map produces a large number of zeros, which contributes to a high percentage of APoZ pruning.Wang et al. \cite{wang2022trainability} proposed trainability preserving pruning (TPP) to maintain the filters' trainability during pruning. TPP utilizes the gram matrix of filters and regularizes the batch normalization parameters. The authors extend the classical pruning stage by emphasizing the pruning of those unimportant filters while preserving trainability. As a result, TPP easily scales to large-scale networks.

In addition, altering the magnitude of the weight decay can indicate the significance of the filters in the network. Penalty-based pruning involves adding penalty terms or constraint functions to the error function during training. This causes the network weights to converge towards zero more quickly. It has been shown that pruning filters that rapidly approach zero is an effective approach. Hanson et al. \cite{hanson1988comparing} utilized hyperbolic and exponential bias terms during backpropagation to determine the degree of weight decay and identify filters for removal. Subsequent studies \cite{lebedev2016fast,wen2016learning,zhou2016less} introduced structured sparse terms or regularization restrictions to the loss function, using backpropagation gradient or splitting methods, to reduce filter redundancy.

Unlike direct sparsification filters, several studies \cite{liu2017learning,ye2018rethinking,marion2023less,grimaldi2023accelerating} achieve redundancy reduction by sparing other factors. Liu et al. \cite{liu2017learning} and Ye et al. \cite{ye2018rethinking} directly apply sparsity-induced regularization to the scaling factors in batch normalization layers. The introduced L1-paradigm sparsity restriction enables the identification of relatively weak channels during training. Huang et al. \cite{huang2018data} adds sparsity regularizations and modifies the stochastic Accelerated Proximal Gradient (APG) to achieve the effect of removing the structures corresponding to zero scaling factors, while adaptively adjusting the depth and width of the network.

Filter pruning, as a method for altering the hierarchical relationship of networks, often falls under the category of structured pruning. This approach involves removing parameters at specific constrained positions, thus modifying the model structure. As a result, the model does not experience an increase in sparsity and can seamlessly integrate into existing deep learning frameworks. However, this pruning method has more limited compression parameters compared to unstructured pruning. Additionally, the layer-by-layer fixed pruning approach requires manual judgment of the sensitivity of each layer and the setting of a controllable threshold. This process demands a significant amount of time to fine-tune the parameters, ultimately affecting the efficiency of offline pruning.

\textbf{\numberedParagraph{Others Types of Pruning}} 

Indeed, while mainstream pruning methods primarily concentrate on determining the significance of network units or neuron connections, certain researchers have explored alternative approaches to achieve pruning compression. For instance, some studies have explored pruning at the network level \cite{goel2020survey,ferbach2022general}, while others have investigated the fusion of different pruning methods \cite{liu2023sparsity,chmiel2022minimum}. These alternative directions offer valuable insights into the field of pruning and contribute to the overall advancement of compression techniques.

Goel et al. \cite{goel2020survey} demonstrated the effectiveness of combining pruning, quantization, and knowledge distillation to enhance performance. By employing these three techniques, they were able to reduce the size of the VGG-16 model to a mere 2\% of its original size. Notably, pruning simplifies the network structure and mitigates overfitting, leading to improved accuracy even with minimal pruning. However, the pruning process necessitates multiple iterations and meticulous fine-tuning, which consumes substantial computational resources and training time. Furthermore, when both pruning and quantization are utilized, the training time increases by a significant 600\%.

As the extension of dense networks and CNNs, general equivariant networks \cite{ferbach2022general} exhibit a certain symmetry in their operations in the strong lottery ticket hypothesis. Damien et al. \cite{ferbach2022general} introduces a unifying framework , which is evidence suggesting that for randomly (logarithmically) overparameterized networks with double the depth exists a high probability of winning tickets.

prove the existence with high probability of winning tickets for randomly (logarithmically) overparameterized networks with double the depth.

Chmiel et al. \cite{chmiel2022minimum} use N:M sparsity for the neural gradients by using a masking mechanism in the forward and backward phases, which the gradients are still unbiased and have minimum variance. The proposed approximate 2:4 and the exact 1:2 algorithms achieve minimal loss in final performance.

To investigate the impact of pruning on models trained on difficult tasks, Liu et al.  \cite{liu2023sparsity}  propose the 'Sparsity May Cry' Benchmark (SMC-Bench). This benchmark aims to address more complex scenarios and potentially overcome the limitations of the 'lazy regime' in large model training.

In the context of network pruning, three main types are identified: weight pruning, filter pruning, and other types. Weight pruning aims to reduce the size and computational effort of the model by eliminating unimportant weights. This approach offers advantages such as simplicity, ease of implementation, and interpretability. However, weight pruning may result in an irregular structure and uneven weight sparsity. On the other hand, filter pruning reduces the model's size and computation by removing entire filters. This method can achieve higher compression rates and is compatible with optimization techniques like accelerated hardware. Nevertheless, filter pruning can lead to information loss and a decrease in model performance. Other types of pruning methods are also considered, taking into account factors such as network sparsity or model fusion. These methods may be more suitable in specific scenarios. Overall, model pruning serves as an effective technique for compressing and accelerating models, significantly reducing redundancy and computation. Further research and improvement of these pruning methods will help enhance the efficiency and performance of models.

\subsubsection{When to Prune}
In this section, we categorize model pruning techniques based on the timing of their application. Specifically, we classify existing techniques into three categories: pruning before training, pruning during training, and pruning after training.

\textbf{\numberedParagraph{Pruning Before Training}}

Pruning before training \cite{lee2018snip,wang2019picking,tanaka2020pruning} typically involves removing connections based on randomly initialized weights before the training process. The primary motivation behind this approach is to bypass the computational overhead associated with pre-training. To elaborate, consider a model $f(\boldsymbol{x}; \boldsymbol{\theta}_0 \odot \boldsymbol{M})$ which utilizes a mask $\boldsymbol{M}$ to prune its randomly initialized weights $\boldsymbol{\theta}_0$. Commonly, we denote $s\in(0,1)$ as the pruning ratio; for instance, if $s$ is 0.2, it signifies a 20\% reduction in weights $\boldsymbol{\theta}_0$. Once pruned, the neural model is trained to convergence with the anticipation of attaining superior performance. As this method circumvents the pre-training phase, it often concurrently reduces computational expenses and shortens both training and testing durations. Several notable methods fall under this category.
Single-shot Network Pruning (SNIP) \cite{lee2018snip} prunes before training by examining the influence of individual weights on the loss function.
Gradient Signal Preservation (GraSp) \cite{wang2019picking} achieves pruning by evaluating the gradient's norm. By doing so, it ensures that pruned weights minimally affect gradient flow, allowing the model to regain performance following sparse training.
Synaptic Flow Pruning (SynFlow) \cite{tanaka2020pruning} identifies sparse sub-networks without training data. They introduce the concept of synaptic flow to mitigate the layer-collapse issue during sparse training.
The study \cite{gebhart2021unified} propose a unified pre-training pruning techniques based on Neural Tangent Kernel (NTK) \cite{jacot2018neural}.
Further studies \cite{wang2020pruning,liu2021unreasonable} have delved into the underlying reasons that enable the identification of effective sub-networks without training. For instance, Liu et al. \cite{liu2021unreasonable} demonstrate experimentally that both the network's size and appropriate pruning ratio play pivotal roles. Similarly, Wang et al. \cite{wang2020pruning} observe that direct pruning of randomly initialized weights can potentially foster the emergence of more diverse and efficient sub-networks.

\textbf{\numberedParagraph{Pruning During Training}}

Pruning during training \cite{huang2018data,evci2020rigging,zhao2019variational,he2018soft,liu2019metapruning} typically entails the direct training of a randomly pruned network while dynamically adjusting the sub-network's structure throughout the training process. Specifically, given a neural network $f(\boldsymbol{x}; \boldsymbol{\theta}_0 \odot \boldsymbol{M})$ with randomly pruned parameters $\boldsymbol{\theta}_0 \odot \boldsymbol{M}$, it's trained directly. Both weights $\boldsymbol{\theta}_0$ and masks $\boldsymbol{M}$ are concurrently updated during the training stage. Upon completion, a trained sub-network is obtained, without any further fine-tuning.
We summarize the following three pruning paradigms.

Dynamic Sparse Training \cite{mocanu2018scalable,mostafa2019parameter,evci2020rigging,liu2021sparse,lin2020dynamic,junjie2019dynamic} pivot on the notion of ``pruning'' and ``growing'' as the model undergoes its training cycle.
Sparse Evolutionary Training (SET) \cite{mocanu2018scalable} functions by eliminating the smallest weights (both most positive and most negative) and subsequently reintroducing new weights.
Dynamic Sparse Reparameterization (DSR) \cite{mostafa2019parameter} introduces a dynamically adaptive pruning threshold, fostering dynamic sparse learning.
For a randomly sparse networks, Rigged Lottery (RigL) \cite{evci2020rigging} grows new weights by analyzing the gradient of pruned weights throughout training.
Gradual Pruning with zero-cost Neuroregeneration (GraNet) \cite{liu2021sparse} prunes old weights and grows new weights based on weight magnitude and gradient.

Sparsity Regularization \cite{meng2020pruning,ding2019global,wen2016learning,lin2019toward,yuan2006model,gordon2018morphnet} attain sparse training through sparsity regularization. These methods initiate with a dense, unpruned neural network. The pruning process is facilitated by imposing a sparsification regularization on the weights.
Structured Sparsity Learning (SSL) \cite{wen2016learning} achieves network sparsity by introducing the group LASSO \cite{yuan2006model} regularization on model weights during the training stage.
MorphNet \cite{gordon2018morphnet} repurposes the parameters of Batch Normalization (BN) and enforces sparsity regularization on BN layers.

\textbf{\numberedParagraph{Pruning After Training}}

Post-training pruning \cite{liu2018rethinking,renda2019comparing,frankle2018lottery,chen2021lottery,chen2021unified} is the most prevalent type of neural network pruning, often credited for identifying the most effective sparse sub-networks. Although this approach tends to increase computational overhead due to the additional pre-training process, it typically results in superior model performance. This method mainly follows a ``pre-training-pruning-retraining'' paradigm. 
Specifically, given a randomly initialized dense neural network $f(\boldsymbol{x}; \boldsymbol{\theta}_0)$, it is trained to convergence after $T$ iterations, yielding $f(\boldsymbol{x}; \boldsymbol{\theta}_T)$. Subsequent to this, pruning units (e.g., filters, neurons or weights) are selected based on specific criteria, resulting in the pruned network $f(\boldsymbol{x}; \boldsymbol{\theta}_T\odot\boldsymbol{M})$. Directly utilizing $f(\boldsymbol{x}; \boldsymbol{\theta}_T\odot\boldsymbol{M})$ for inference tends to underperform. To recover the performance, $f(\boldsymbol{x}; \boldsymbol{\theta}_T\odot\boldsymbol{M})$ typically undergoes fine-tuning the model weights $\boldsymbol{\theta}_T$ on the original dataset.

The Lottery Ticket Hypothesis (LTH) \cite{frankle2018lottery,frankle2020pruning} stands out as a cornerstone hypothesis in network pruning. It posits that within a randomly initialized dense network, there exists a sparse sub-network (referred to as the ``winning ticket'') that, when trained from scratch, can achieve the dense network's performance. To validate this hypothesis, LTH employs an Iterative Magnitude Pruning (IMP) strategy, wherein a pre-trained model iteratively prunes weights based on magnitude. Post-training, the remaining weights are reset to their original initilization. This hypothesis has inspired numerous research endeavors \cite{malach2020proving,orseau2020logarithmic}. Some probe into stronger propositions like Multi-Prize Tickets \cite{diffenderfer2020multi}. 
Several studies \cite{chen2021unified,morcos2019one,mehta2019sparse,gan2022playing} have investigated the transferability of the winning tickets, such as its application in self-supervised pre-training models \cite{chen2021lottery} or across datasets \cite{morcos2019one}.
Furthermore, beyond image classification with Convolutional Neural Networks (CNN) backbones, the existence of lottery tickets has been explored in diverse model architectures and tasks, including BERT \cite{chen2020lottery}, Transformers \cite{chen2021chasing}, Generative Adversarial Networks (GANs) \cite{chen2020gans}, Graph Neural Networks (GNNs) \cite{chen2021unified,sui2021inductive}, and recommendation systems \cite{wang2022exploring}. 
These studies greatly broaden the potential applicability of LTH.
However, the LTH's reliance on the IMP strategy, which necessitates multiple iterative training rounds, can exponentially amplify training costs. As a consequence, research has sought ways to pinpoint lottery tickets more efficiently. For instance, Early-Bird (EB) ticket \cite{you2019drawing} suggests that lottery tickets can be discerned early during training. Similarly, EarlyBERT \cite{chen2020earlybert} extend this observations within the context of the BERT pre-training model. PrAC \cite{zhang2021efficient} efficiently finds winning tickets from the perspective of training data.

\subsubsection{Pruning Criteria}

In the pruning process, deciding which network components to delete (e.g., filters, neurons or weights) based on specific criteria is crucial. These criteria, thus, play a pivotal role in effective pruning. This section provides a brief overview of some widely-adopted pruning criteria, including magnitude, norm and loss sensitivity.

\textbf{\numberedParagraph{Magnitude}}

Intuitively, weights in a neural network with magnitudes closer to 0 exert minimal influence on the model's overall behavior. Hence, the weight's magnitude can be perceived as an indicator of the connection's importance within the network. The magnitude-based pruning criteria can be defined as:
\begin{equation}
  M_{i}=\left\{
\begin{aligned}
1 & : if \, ||\theta_i||_1 \geq a \\
0 & : if \, ||\theta_i||_1 < a
\end{aligned}
\right.,
\end{equation}
where $M_i$ and $\theta_i$ denote the $i$-th element in the mask $\boldsymbol{M}$ and the model weight $\boldsymbol{\theta}$; $a$ is the pre-defined pruning theshold.
Hanson et al. \cite{hanson1988comparing} introduce magnitude-based pruning aimed at reducing hidden neurons.
Han et al. \cite{han2015deep} adopt magnitude-based pruning strategies tailored for deep neural networks.
Magnitude-based pruning can be seamlessly integrated into both non-structural and structural pruning paradigms. For instance, the IMP in LTH uses the magnitude-based pruning criteria. Early-Bird (EB) ticket \cite{you2019drawing} employs weight magnitudes within the Batch Normalization (BN) layer for structural channel pruning.

\textbf{\numberedParagraph{Norm}}

Norm-based pruning is a broader manifestation of magnitude-based pruning. It evaluates the aggregated importance of specific components (e.g., filters, neurons or weights). In structured pruning contexts, the importance of filters \cite{he2018soft}, for example, can typically be ascertained using the following expression:
\begin{equation}
  ||\mathcal{F}_{i,j}||_p=\sqrt[p]{\sum\limits_{n=1}^{N_i}\sum\limits_{k_1=1}^{K}\sum\limits_{k_2=1}^{K}|\mathcal{F}_{i,j}(n,k_1,k_2)|^p},
\end{equation}
where $N_i$ denotes the number of input channels for the $i$-th convolution layer; $K$ is the kernel size.
According to this criteria, a filter $\mathcal{F}_{i,j}$ with a smaller $\ell_p$ norm indicates that it is less important and more likely to be pruned.

\textbf{\numberedParagraph{Loss Sensitivity}}

Sensitivity evaluates the significance of components (e.g., network weights or filters) concerning the final model's loss. Thus, it offers a lens into each component's importance from a training perspective. For instance, SNIP \cite{lee2018snip} employs a sensitivity metric termed ``connection sensitivity'', expressed as:
\begin{equation}
  s_i=\frac{|g_i(\boldsymbol{\theta}; \mathcal{D})|}{\sum_{k=1}^m|g_k(\boldsymbol{\theta};\mathcal{D})|},
\end{equation}
where $\mathcal{D}$ is the dataset, $s_i$ is the sensitivity of the weight $\theta_i$; $g_i$ is the derivative of the loss w.r.t the mask ${M}_i$.
Greater weight sensitivity denotes heightened importance. 
Concerning the loss function's variability, some methodologies \cite{you2019gate,nonnenmacher2021sosp,liu2021group} leverage the first-order or second-order Taylor expansion to gauge a component's impact on the loss function. 

\subsection{Low-Rank Factorization}
\label{Sec:lowrank2}
Section~\ref{Sec:lowrank1} primarily discusses the construction of a new network for training by replacing original modules with low-rank modules. However, if we already have an existing network, we can also employ low-rank decomposition to inherit the parameters of the original network. Most of the work in Section 3.1 is based on random initialization rather than decomposing parameters from a pretrained network. Therefore, in this section, we place more emphasis on cases where we already have a pretrained neural network and use decomposition algorithms, not just decomposition formats, to adjust the original network for inference purposes. Of course, the low-rank neural network in Section 3.1 generally performs better than the original networks during inference time.

Xue et al.~\cite{DBLP:conf/icassp/XueLYSG14} introduced the application of SVD decomposition to compress fully-connected neural networks. Rigamonti et al.~\cite{DBLP:conf/cvpr/RigamontiSLF13} suggested approximating trained CNNs with low-rank filters. Additionally, Denton et al.~\cite{DBLP:conf/nips/DentonZBLF14} extended this idea by leveraging the inherent linear structure within convolutional filters.
Tensor decomposition and matrix factorization algorithms can be directly applied to the parameters, and after decomposition, training can also be conducted. For instance, 
the matrix ALS~\cite{lu2021numerical} can be employed to identify low-rank neural networks, thereby reducing the memory footprint of the neural networks while simultaneously enhancing their performance.
TensorGPT~\cite{xu2023tensorgpt} efficiently compresses the pretrained embedding layer within LLMs through Tensor-Train Decomposition (TTD). By representing each token embedding as a Matrix Product State (MPS), the embedding layer can achieve compression rates of up to 38.40 times, all while maintaining or even enhancing the model's performance when compared to the original LLM. All matrix decomposition and tensor decomposition can be employed to approximate parameters, but these decomposition algorithms are always combined with the model design discussed in Section~\ref{Sec:lowrank1}. Thus, this Section does not delve into more detailed discussions.

\subsection{Quantization}

Large language or vision models are pre-trained on high precision data type such as floating point precision which requires high computation and memory resources, which hinder the application of large model. Quantization aims to accelerate the inference and reduce the storage of model by using lower bid widths for storage and computation, while minimizing the accuracy drop as much as possible. 
The key of quantization is defining a quantizer that maps a real valued neural network weight or the activation to a lower precision quantized one:
\begin{equation}
  c(x) = X_i\text{, if }x \in [\Delta_i, \Delta_{i+1})
\end{equation}
where the $X_i$ represents the low precision quantized value and $\Delta_i$ is the quantization steps.
The simplest and most popular uniform quantization function is:
\begin{equation}
  c(x)=Round(x/S)-Z
\label{easy_quant}
\end{equation}
where $S$ is real valued scaling factor, $Z$ is integer zero point that ensure zero is quantized with no error and $Round(*)$ function maps a real value to nearest integer value. We will give a brief introduction to quantization methods based on Equation ~\ref{easy_quant}.
\subsubsection{Quantization Parameter}
In Eq.~\ref{easy_quant}, the scale factor can be calculated as following:
\begin{equation}
  S=\frac{x_{max}-x_{min}}{2^b-1}
\end{equation}
where $b$ is the bitwidth after quantization, $x_{max}$ and $x_{min}$ are the maximum and minimum value of quantizer inputs, and the process of choosing $x_{max}$ and $x_{min}$ is called calibration. According to the time of calibration process, quantization methods can be classified to \textbf{static quantization} and \textbf{dynamic quantization} ~\cite{2023-aaai-dynamic-1, 2022-cvpr-dynamic-2, 2022-emnlp-dynamic-3, 2022-eccv-dynamic-4}. For static quantization methods, they first run the model with a set of calibration data and record the value of activations, then the quantization parameters can be computed as constants for any other inputs data. In dynamic quantization, the quantization parameters of model weights are pre-computed, while the activations are dynamic quantized in inference. According to the value of weights and activations, there are several metrics to determining clipping thresholds, such as min-max~\cite{2018-cvpr-min-max}, mean squared error~\cite{2019-eccv-mse}, percentile~\cite{2018-arxiv-percentile} and cross-entropy. Furthermore, the quantization parameters could be also trained jointly with neural networks~\cite{2013-arxiv-learn-param-1, 2018-arxiv-learn-param-2, 2020-ICLR-learn-param-3, 2020-MLSys-learn-param-4, 2018-eccv-lq-nets}. Due to dynamic quantization methods compute quantization parameters for each input, the cost of performance improvement is an increase in inference latency, which is a trade-off between accuracy and latency.

According to the whether the center value of the quantized range $x_{max}$ and $x_{min}$ equals zero, i.e., $x_{max}+x_{min}=0$, quantization methods can be classified to \textbf{symmetric} quantization and \textbf{asymmentric} quantization, and in symmetric mode, the zero point $Z$ is set to 0. The main considering of asymmetric is two folds, first, the quantized range is fully utilized in asymmetric mode, this is because the min and max value of input range can exactly map to the min and max value of quantized range. Moreover, symmetric methods are inappropriate when the input range is biased towards one side, such as the activations of ReLU output, using symmetric methods in this case would waste much quantized space in the sparse side. In most cases, symmetric quantization is used for model weights because model weights are usually zero centered distributed. Asymmetric quantization, no the other hand, is often used for quantizing activations. 

\subsubsection{Quantizer Design}
In Equation ~\ref{easy_quant}, the quantized spaces $\Delta_i - \Delta_{i-1}$ between each quantization levels are equal, which is call \textbf{uniform} quantization. Conversely, if the quantization space between quantization levels is not the same, it is called \textbf{non-uniform} quantization. Due to distribution of weights and activations are not uniform, a well-designed non-uniform quantization methods always achieve higher accuracy because it can better capture the underlying distribution. Non-uniform quantization methods can be divided into three sub-categories~\cite{a_survey_of_quanti}: rule-based, optimization-based and clustering-base. Logarithmic distribution and power law distribution are often utilized in rule-based quantization, QLORA~\cite{QLORA} proposes a new data type which is efficient for quantization model weights that approximate normal distribution. Opimization-based quantization methods optimize model weights and learnable quantization functions simultaneously ~\cite{2021-cvpr-learnable, 2018-iclr-model_compression, 2018-eccv-lq-nets}. Clutering-based methods utilize k-means~\cite{2014-arxiv-comp-deep, 2016-cvpr-quant-conv, 2015-arxiv-deep-comp} to set the quantization steps and levels to alleviate the information loss. Although non-uniform quantization achieve better performance than uniform quantization, it is inefficient to deploy non-uniform quantization on general hardware~\cite{2021-cvpr-learnable}, such as GPU and CPU. 

In Equation ~\ref{easy_quant}, $Round()$ function is utilized to convert floating point value to integers, which is a deterministic transformation. Inspired by dropout, some works ~\cite{2015-nips-binary-stochastic-1, SDQ, 2023-cvpr-nipq-stochastic-3},introduce noise in quantization, called \textbf{stochastic} quantization. The intuitive idea is that the noise introduced in stochastic quantization can improve the model's generalization and robustness. A simplest quantizer close to Equantion ~\ref{easy_quant} can be given as:
\begin{equation}
  c(x) = Round(x/S+\epsilon) - Z
\end{equation}
\begin{equation}
  \epsilon \sim Uniform(-1/2, 1/2)
\end{equation}
where $\epsilon$ is a noise that follows uniform distribution. In addition to model weights and activations, some works introduce randomness in other modules. For example, Quant-Noise ~\cite{quant_noise} randomly selects a subset of weights and only quantize the selected weights in each training forward, which allows unbiased gradients for other weights. SDQ~\cite{SDQ} presents the choice of discrete ditwidths as a set of Differentiable Bitwidth Parameters(DBPs), which is used as probability factors in choosing quantization bidthwidth. QDrop~\cite{QDrop} randommly drops quantization of activation during post-training quantization.

Furthermore, according to the granularity in convolution layer, quantization can be classified to \textbf{layer(tensor)-wise} quantization and \textbf{channel-wise} quantization. In layer-wise quantization, the hyper-parameters such as scaling factor $S$ and zero point $Z$ are same within each layer and different across different layers. However, layer-wise quantization is sub-optimal, especially for convolution filters with tight distribution, which waste a lot of quantization space and lead to decreased performance. Channel-wise quantization overcomes this drawback by adapting the quantizer parameters for each filter in a layer. Generally, the bitwidth is an uncomputed hyper-parameter, and due to the large search space for exploring different channel bitwidths, it is usually set to be same for all channels. While RDO-Q~\cite{RDO_Q} argues that different channels have unique reactions to quantization and suggests that assigning unequal bit widths to channels can yield higher precision. It treats quantization as a rate-distortion optimization problem and apllies classic coding theories to address the complexity challenge.

\subsubsection{Fine-tuning Mode}
According to whether combine quantization with retraining, quantization can be classified to ~\textbf{post training quantization}(PTQ) and ~\textbf{quantization-aware training}(QAT). As PTQ does not involve a retraining stage, it only requires a minimal amount of calibration data to calculate the static quantizer parameters of weights and activations or to directly perform dynamic quantization during inference. This makes PTQ is a simple and first go-to tool in quantization. However, PTQ leads to substantial disparities between model training and inference, and despite the inherent resilience of neural networks to quantization, it still makes a performance drop, particularly when targeting low-bit quantization. Therefore, QAT are proposed to address the problem. QAT describes the technique of inserting quantization operations into the neural network in the retraining stage to adapt the model from floating point to quantized weights and activations. During the forward pass, QAT simulates quantization operations on both weights and activations. However, a notable issue with QAT during backporpagation is that quantization is a non-differentiable discrete function, rendering gradient descent methods unsuitable for optimization. One simple but effective approach is Straight Through Estimator(STE)~\cite{2013-arxiv-learn-param-1}, which ignores the rounding operation and approximates it with an identity function. However, the inconsistency of STE during forward and backward may result in unstable gradients, especially in the case of low-bit quantization such as binarization. Sajad et al.~\cite{2018-arxiv-regularized} thinks that STE fails to learn weights near the border and proposes a new activation to alleviate this issue. ReSTE ~\cite{ReSTE} designs two indicators to quantitatively demonstrate the equilibrium phenomenon, and the power function-based estimator proposed by ReSTE effectively enhances the original straight-through estimator by achieving a balanced optimization between estimator error and gradient stability. Despite its superior performance compared to PTQ, QAT also has the following limitations: QAT requires pre-training data for retrainig, which may be difficult to obtain due to privacy concerns. Additionally, the retraining stage of QAT is time-consuming and is difficult to converge, resulting in a significant increase in expenses. ~\textbf{Data-free}~\cite{2022-eccv-patch-datafree-1, 2022-Arxiv-AdaDFQ, 2022-cvpr-PNMQ} is another category quantization that does not require pre-training data as in QAT or calibration data as in PTQ. It effectively avoids potential privacy issues. ~\cite{2022-cvpr-PNMQ} employs Parametric Non-uniform Mixed precision quantization to generate a quantized network without data, and ~\cite{2022-Arxiv-AdaDFQ} utilizes the fake samples generated by a generator that learnt from full-precision network and alleviate over-and-under fitting issues.

\begin{figure}[!tbp]
  \setlength{\belowcaptionskip}{-0.5cm}  
  \centerline{\includegraphics[width=3.28in]{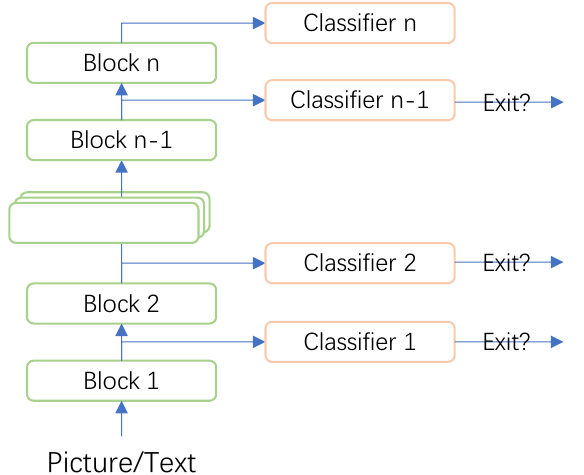}}
  \caption{An overview of early exit.}
  \label{fig:early_exit}
\end{figure}

\subsection{Distillation}

\textbf{\numberedParagraph{Logits Distillation}}

Logits distillation refers to using the prediction results of a teacher model to guide the training of a student model.

In recent years, \cite{logits_dis:wang2022gradient} proposed to use higher-order information such as the gradient in the teacher model, to help the learning of student model. This approach, known as Gradient Knowledge Distillation (GKD), has been experimentally shown to outperform previous methods of knowledge distillation, significantly improving interpretability.
Though KL Divergence is one of common objectives of KD, it may lead to poor performance in complex distribution situations such as text generation in LLM. So \cite{logits_dis:gu2023knowledge} proposed to use reverse KL divergence instead to prevent student from learning too many long-tail variants of the teacher white-box LLM. And extensive experiments demonstrating that improved performance across various tasks.

Many applications of logits distillation have been proposed across various fields. For example, \cite{logits_dis:zhou2023universalner} employed Mission-Focused Instruction Tuning with logits distillation for named entity recognition. \cite{logits_dis:sanh2019distilbert} utilized knowledge distillation during the pre-training stage to reduce the model size by 40$\%$ and speed up by 60$\%$, while maintaining 97$\%$ performance of the origin model.
As for the context of large language models, \cite{logits_dis:hsieh2023distilling} proposed a method that take few-shot CoT Prompt as input of the teacher LLM, which generate corresponding output rationales for logits distillation training, to decrease the reliance on labeled data.
\cite{logits_dis:ma2023sci} leveraged distillation in training two smaller student language models specifically for the generation of rationales and answers, respectively.

\textbf{\numberedParagraph{Feature Distillation}}

Feature distillation (FD) refers to the transfer of knowledge to a student model by passing the intermediate layer features of the teacher model. This technique has found numerous applications in Computer Vision. 

For instance, \cite{feature_dis:zhang2022dtfd} applied FD in the classification of histopathology whole slide images, while
\cite{feature_dis:wei2022contrastive} proposed the use of feature distillation to convert the subpar representations generated by previous prevalent pre-training methods (such as image classification, instance contrastive learning, etc.) into new representations that possess desirable properties similar to those produced by Masked image modeling.
\cite{feature_dis:gao2022feature} introduced a lightweight yet efficient Feature Distillation Interaction Weighted Network, which uses a specially designed backbone consisting of Feature Shuffle Weighted Groups, as well as novel mutual Wide-residual Distillation Interaction Blocks. 
\cite{feature_dis:li2022self} proposed a teacher-free feature distillation framework, that transfers knowledge from salient feature to redundant feature within the same layer (intra-layer distillation) and from deeper layer to shallow layers (inter-layer distillation) within the student model, eliminating the need for extra teacher models.
Additional, \cite{feature_dis:yang2022masked} proposed Masked Generative Distillation, which enhances students' representation capability by masking pixels of the student's feature randomly and compelling it to reconstruct the full feature through a projector neural network.

\textbf{\numberedParagraph{Relation Distillation}}

This part aims to provide a review of recent advancements in relation distillation techniques.

\cite{relation_dis:liu2019knowledge} proposed a method called Instance Relationship Graph for knowledge distillation. This approach constructs a graph to represent the distilled knowledge of a specific layer in a network. The graph is built by considering instance features and instance relationships as vertices and edges, respectively.
\cite{relation_dis:wang2020minilmv2} proposed to utilize self-attention relation knowledge contained in self-attention module to train the student model for distillation of the pretrained transformers.
\cite{relation_dis:deng2019relation, relation_dis:hu2021dense} adopted relation distillation techniques and achieved promising results in object detection.

\subsection{Early Exit}

With the gradient vanishing problem in deep neural networks solved by residual structures, neural networks have become increasingly deeper, such there are more than one hundred convolution layers in ResNet. However, deep neural networks have high requirements for computational resources and energy. The serial structure of deep neural networks has a linear relationship between inference latency and the number of layers, which hinders their deployment in real-time and energy-sensitive scenarios. Through continuous research on deep neural networks, it has been discovered that not all inputs need to pass through the entire deep neural network to obtain correct outputs, shallow neural networks are capable of handling simple inputs. Therefore, early-exit methods have been proposed to accelerate deep neural networks inference, in which if the output of the shallow network is reliable enough, it can be taken as the final result of the entire network. Generally, early exit methods can be classified into two categories based on exit strategies: entropy-based early exit and learning-based early exit methods. In this section, we will briefly introduce early exit methods based on this classification.
\subsubsection{Entropy-based early exit}
Entropy-based early exit is the most commonly used method~\cite{BranchyNet}. Taking image classification as an example, a classifier is placed after each layer of the deep neural network. During training, the classifiers after each layer are jointly trained:
~\begin{equation}
  L(x,y)= \sum_{i=1}^NL_{CE}(y, \hat{y_i})
\end{equation}
where $x$, $y$, $y_i$ and $L_{CE}$ represents the input, label, output of i-th classifier and cross-entropy loss respectively. During inference, after obtaining the output of each layer's classifier, the entropy of the distribution is calculated and compared with a predefined threshold. If the entropy is greater than the threshold, it is considered that the shallow network has enough capability to classify the input sample and early exit the deep neural network, using the result of the current layer as the output.Confidense-based method similar to entropy-based methods ~\cite{not_all_pixels}: If the probability value corresponding to the classifier predicting category is larger than the predefined threshold, the current shallow networks is considered to be sufficiently reliable.

Hao et al.~\cite{improved_techniques_for_training} argues that the loss function may lead to the issue of gradient imbalance due to the reusability of the lower-level layers. Specifically, when considering training a k-exit adaptive network using the sum of cross-entropy losses from all classifiers, the backward graph can be described as a binary tree with a depth of k, where gradients come from the left nodes and propagate from child nodes to parent nodes. This can result in gradient explosion in the lower-level networks, making the model training difficult. Therefore, they proposed a gradient normalization method to balance the gradients of the lower-level networks. 
Yigitcan et al.~\cite{shallow_deep_networks_understanding} considers another perspective on the necessity of early exit: overthinking, where correct predictions in intermediate layers may turn into incorrect predictions in the final layer. Additionally, based on the idea of pre-trained models, two training methods are proposed: only training the classifiers based on pre-trained models or training the entire network.

With the widespread use of the Transformer in pre-trained language models, Transformer-based neural language models are also becoming increasingly deep. 
DeeBERT~\cite{deebert} applies early exit to the BERT model and achieves good results on sentence-level classification tasks. 
FastBERT~\cite{fastbert} also applies early exit to the BERT model. In addition, it proposes self-distillation for branches, where the output of the top Transformer layer is used as the teacher and other Transformer layers are students, and distillation loss is calculated using KL-divergence. 
PABEE~\cite{bert_loses_patience} analyzes the overthinking phenomenon in BERT and considers multiple consecutive classifiers. It stops early only when the entropy values from multiple consecutive layers exceed a predefined threshold, which avoids incorrect outputs from shallower layers. 
The early exit methods based on pre-trained language models mentioned earlier are primarily designed for sentence-level tasks. Furthermore, Xiaonan et al. ~\cite{accelerating_bert_inference} extends the early exit method to token-level sequence labeling tasks, allowing certain tokens to exit at different layers. Considering the local dependencies in token-level tasks, the authors design a window-based criterion to determine whether or not a token should exit. Additionally, the authors propose a Halt-and-Copy strategy to update the representations of tokens that have already exited. 
Kaiyuan et al.~\cite{a_global_past_future} believe that relying solely on predictions from lower layers inevitably leads to the loss of higher-level features from future layers, resulting in sub-optimal performance. To address this issue, the authors propose a novel past-future approach based on the perspective of distillation learning, which enables comprehensive predictions from a global perspective.
DeeCap~\cite{deecap} applies early exit to image captioning. Through preliminary experiments, it concludes that internal classifiers are not reliable. Therefore, it proposes an approach based on the imitation learning mechanism to simulate top-layer representations using bottom-layer representations. Additionally, it designs a Multi-Level Representations Fusion module to incorporate predictions from all available layers' representations. 
Shengkun et al.~\cite{you_need_multiple_exiting} argues that previous strategies cannot be applied to the encoder in the widely-used unified architecture with both encoder and decoder due to the difficulty of estimating output confidence in the encoder layers. The use of early exit on the encoder can further optimize the inference speed. For the task of image-text matching, the authors decompose the modeling process of the image and text modalities and make exit decisions based on the similarity between the intermediate layers of the two modalities.

\subsubsection{Learning-based early exit}
In learning-based early exit, whether to make an exit decision at the current layer or at which layer to make an exit decision is determined through end-to-end learning. Maha et al. ~\cite{depth_adaptive_transformer} applies the early exit method to sequence generation tasks and proposes different adaptive depth estimation approaches for sequence-specific tasks and token-specific tasks. For sequence-specific tasks, the average of all hiddens is taken as the input for the layer predictor. The predictor's labels are obtained through Likelihood-based and Correctness-based methods. For token-specific tasks, there are two layer estimation methods: Multinomial and Geometric-like. Additionally, an additional confidence-based method is used to obtain the labels for the layer classifier. 
Yijin et al.~\cite{fast_depth_adaptive_transformer} believes that previous works~\cite{depth_adaptive_transformer} often build a halting unit to decide whether computation should continue or stop at each layer. Due to a lack of specific supervision for depth selection, the halting unit is sub-optimal and inaccurate, leading to unstable performance in modeling sentences, and they propose two un-supervised estimations-Mutual Information Based Estimation and Reconstruction Loss Based Estimation. Berxit~\cite{berxit} is also a learning-based method, where the ground truth certainty level is determined by whether the classifier makes the correct prediction.
HASHEE~\cite{hashee} differs from general learning-based methods as it replaces the learning modules with hash functions to assign each token to a fixed exiting layer. Unlike previous methods, HASHEE does not require internal classifiers or extra parameters, making it more efficient.

\section{Green Computing Systems}
\subsection{Resource Optimization}

\subsubsection{Cluster resource scheduling}
For computation-intensive, long-running jobs (e.g., deep learning training jobs) in the cluster, expensive hardware devices (eg. GPUs or TPUs) are often demanded for distributed execution aiming to complete the jobs within reasonable time. An appropriate scheduler can play a significant role on improving cluster utilization, resource fairness, and job completion times. 

Earlier works\cite{gu2019tiresias,xiao2018gandiva} adopt inelastic schedulers, i.e., agnostic to the performance scalability of jobs with respect to the amount of allocated resources. These schedulers require users to specify the resource amounts and improve resource utilization by strategies such as time-sharing and job packing. Recent studies\cite{mahajan2020themis,qiao2021pollux,gu2023elasticflow} propose to schedule the resources elastically, i.e., developers don't need to explicitly declare the required resources, thus alleviating their pressure of low-level resource management. Generally, the key steps of scheduling the resources are as follows.






\textbf{Load estimation}. Load estimation means predicting the arrival or duration of the jobs based on the historical information, which helps make better resource decisions. The estimation can rely on the recurrent jobs\cite{weng2022mlaas} or job structure knowledge\cite{venkataraman2016ernest}, or, for more general cases, learning from the historical information of relevant jobs. For example, MLaaS\cite{weng2022mlaas} observes the phenomenon of repetitive jobs in Alibaba and uses decision tree models to predict jobs' execution time, achieving less than 25\% prediction error for 78\% instances. In addition, Lucid\cite{hu2023lucid} utilizes job-specific profiled characteristics  to improve prediction accuracy while keeping the prediction model explainable.

\textbf{Resource allocation}. At the core of cluster management is resource allocation, either for homogeneous resources\cite{hu2023lucid} or heterogeneous resources (e.g., Gavel\cite{narayanan2020heterogeneity} improves resource utilization by leveraging the heterogeneity of GPU generations). Narayanan et al.\cite{narayanan2021solving} demonstrate that numerous resource allocation problems within computer systems are characterized by granularity, i.e., each client requests only a small portion of the total amount of resources, and partitions the large-scale problem into more tractable sub-problems to improve scheduling efficiency. Considered elastic resource allocation, Lucid\cite{hu2023lucid} leverages an indolent packing approach to mitigate interference and schedules resources according to estimated job priorities and sharing scores to efficiently scheduling GPUs. Cilantro\cite{bhardwaj2023cilantro} argues that resource allocation systems should directly account for real-world performance and the varied allocation objectives of users, therefore employing an online learning mechanism to estimate the resource-to-performance mappings and load shifts.

\textbf{Job placement}. Besides the decision of jobs' resource allocation, how to properly place the jobs distributed among different nodes is also challenging. Harmony\cite{bao2019harmony} proposes a deep reinforcement learning based framework for efficient job placement, where the reward is maintained through a neural network learned from limited historical traces. ElasticFlow\cite{gu2023elasticflow} further considers the effect of workers topology, which may impact the  parameter synchronization overhead, and adopts buddy allocation strategies to eliminate the effect of topology.

\subsubsection{Server resource partitioning}
Workload consolidation has become a widely used approach in data centers to enhance the performance of co-located interactive jobs. Resource partitioning is a class of finer-grained, multi-resource optimization strategies relying on the hardware resource isolation techniques (e.g., Intel's CAT\cite{intel_cat} for last level cache isolation), which aims to mitigate the interference and resource contention among co-located jobs. The main challenge of reasonably partitioning the resources lies in the following aspects (summarized in Table \ref{tab:part_solutions}). 

\begin{table}[htbp]
  \centering
  \caption{Main resource partitioning solutions.}
    \begin{tabular}{|p{5em}|p{11em}|p{6em}|p{10em}|p{8.5em}|}
    \toprule
    \textbf{Solutions} & \textbf{Main proposition} & \textbf{Prior\newline{}knowledge} & \textbf{Partitioned resources} & \textbf{Supported\newline{}objectives} \\
    \midrule
    \textbf{dCAT}\cite{xu2018dcat} & dynamic cache management & not required & Last Level Cache (LLC)   & throuput \\
    \midrule
    \textbf{CoPart}\cite{park2019copart} & coordinated partitioning of cache and memory bandwidth & not required & LLC, memory bandwidth & fainess \\
    \midrule
    \textbf{PARTIES}\cite{chen2019parties} & QoS-aware resource management while maximizing throughput for the machine & not required & hyperthread, CPU cores, power, LLC capacity, LLC bandwidth, memory bandwidth, memory capacity, disk bandwidth and network bandwidth & QoS guarantee for latency-critical jobs and throuput for the machine \\
    \midrule
    \textbf{CLITE}\cite{patel2020clite} & Bayesian Optimization-\newline{}based multi-resource partitioning & not required & CPU cores, LLC, memory bandwidth, memory capacity, disk bandwidth and network bandwidth & QoS guarantee for latency-critical jobs and throuput for the machine \\
    \midrule
    \textbf{DRLPart}\cite{chen2021drlpart} & deep reinforcement\newline{}learning-based resource partitioning & requiring\newline{}offline training & CPU cores, LLC and memory bandwidth & throuput \\
    \midrule
    \textbf{OSML}\cite{liu2023intelligent} & avoidance of resource cliffs using many learning models & not required & CPU cores, LLC and memory bandwidth & QoS guarantee for latency-critical jobs \\
    \midrule
    \textbf{OLPart}\cite{chen2023olpart} & intelligent search indicated by runtime performance counters & not required & CPU cores, LLC and memory bandwidth & QoS guarantee for latency-critical jobs and throuput for the machine \\
    \midrule
    \textbf{SATORI}\cite{roy2021satori} & tradeoff  between fairness and throuput & not required & CPU cores, LLC and memory bandwidth & throuput and fairness \\
    \midrule
    \textbf{Orchid}\cite{chen2023orchid} & awareness about runtime system status with multiple objectives & not required & CPU cores, LLC and memory bandwidth & throuput and fairness \\
    \bottomrule
    \end{tabular}%
  \label{tab:part_solutions}%
\end{table}%

\textbf{Large and multi-dimensional search space}. Considering the multi-resource demand of complex job co-locations, it is highly profitable to simultaneously partition multi-dimensional resources. The early work on resource partitioning, dCAT\cite{xu2018dcat}, leverages Intel’s CAT to partition only LLC among co-located applications. CoPart\cite{park2019copart} further proposes to partition both LLC and memory bandwidth while guaranteeing the fairness of jobs. These methods are limited to only one or two supported resources, thus lacking exploration efficiency.

PARTIES\cite{chen2019parties} considers QoS-aware resource partitioning for many latency-critical services while optimizing the throughput of best-effort jobs. The partitioning strategy of PARTIES relies on the observation of "resource fungibility", i.e., different resources can be traded for each other to arrive to similar system performance. After that, CLITE argues that optimizing one resource at a time is not efficient enough, and a Bayesian Optimization based solution is proposed to improve the partitioning efficiency. Further, facing the complex exploration space, OSML\cite{liu2023intelligent} discovers the phenomenon of “resource cliffs” near the resource configuration of QoS violations, and then employs multiple machine learning methods collaboratively to avoid resource cliffs and make better resource partitioning choices.


\textbf{Unpredictable workload interference}. It is usually hard to obtain prior knowledge about complex varying workload in practical systems, thus accurately modeling the interference of jobs can contribute to high-performance decisions a lot. Previous works\cite{sanchez2011vantage} usually establish dedicated analytical performance model to guide the exploration. Nevertheless, these methods typically depend on extensive domain knowledge. More autonomous techniques, OLPart\cite{chen2023olpart} and Orchid\cite{chen2023orchid}, propose to collect real-time system status to indicate the resource sensitivity of workloads, which is then input as the contextual information of online learning models to make better partitioning choices. DRLPart\cite{chen2021drlpart} also adopts the performance counters of the system, yet leveraging deep reinforcement learning to decide final partitioning schemes.

\textbf{Competing optimizing objectives}. Previous studies considers optimizing either the fairness\cite{park2019copart}, or the application performance\cite{chen2019parties,chen2023olpart,chen2023orchid}. SATORI\cite{roy2021satori}, however, finds that there exits the opportunity to optimize the two competing objectives, and uses a dynamic weighting method to trade off fairness and throughput. Then, Orchid\cite{chen2023orchid} suggests the maintenance of many distinct multi-armed bandit (MAB) models for the concurrent learning of the two objectives, affording the flexibility to trade off between these two objectives as per the decision maker's aspirations. 

\textbf{Other optimizations}. Other studies related to resource partitioning generally consider incorporating more optimizing objects. JointOpt\cite{chen2023jointly} proposes a joint optimization framework with job assignment. CoTuner\cite{yang2023cotuner} coordinately optimizes resource partitioning and parameter tuning. To overcome the interdependency between configurations and huge exploration space, CoTuner constructs a hierarchical architecture collaborating many sub-models to work together.

\subsection{Energy-Efficient Data Storage}


With the rapid expansion of data, the imperative for energy-conscious storage options, termed "green storage," is more pressing than ever. Green storage endeavors to enhance performance and simultaneously reduce energy use, harmonizing the needs of today with the principles of sustainability. This notion can be segmented into three primary domains: \textbf{Energy Efficiency Storage, Hardware/ Software Co-Design and Distributed Resource Management}. Each domain presents distinct strategies to realize the vision of sustainable data storage.

\subsubsection{Energy Efficient Storage}
In this section we introduce the methods that directly contribute to reducing energy consumption and enhancing energy efficiency in data storage.

ResTune\cite{restune}focuses on optimizing resource utilization in cloud databases. The meta-learning approach aids in efficient tuning, indirectly contributing to energy savings.
DimmStore\cite{memory_power} addresses memory power consumption in database servers, aiming to optimize power usage in scenarios where servers aren't fully utilized. 
Kissinger et al.\cite{imdb_energycontrol} investigates energy consumption in in-memory database systems, emphasizing the optimization of power consumers like processors and main memory. Furthering this exploration, Kissinger et al.\cite{imdb_energy_utility} proposes energy-utility functions to improve energy efficiency in scalable in-memory database systems.
GreenDB\cite{greendb} specifically targets energy efficiency by optimizing prefetching and caching in database clusters.
Ranjbari et al.\cite{cloud_storage} concentrate on enhancing the energy efficiency of virtual machine consolidation within cloud data centers, simultaneously addressing energy consumption and service level agreement considerations.
Hassan et al.\cite{hybrid_storage1} discusses energy-efficient data management on hybrid 
main memory systems, optimizing both performance and power usage.

\subsubsection{Hardware/Software Co-Design}
In this section, we review the technique which emphasizes the synergy between hardware and software, aiming to achieve optimal energy efficiency through co-design.

Polynesia et al.\cite{Polynesia} proposes a hardware-software co-designed system for in-memory HTAP databases, aiming to avoid throughput losses and enhance energy efficiency.
HAMS\cite{nvm1} proposes a hardware solution for memory-over-storage, aiming to optimize the benefits of persistent memories like NVDIMM.
Automatic-SSD\cite{autoSSD} advocates for full hardware automation for storage, emphasizing the energy efficiency and high performance of new memory-based storage.
Yoon et al.\cite{PCM} introduces a PCM-based memory storage architecture with adaptive data filtering, aiming for energy efficiency in embedded devices.

\subsubsection{Distributed Resource Management}
In this section, we introduce the management and allocation of resources, ensuring that systems operate at optimal energy efficiency even in parallel processing scenarios.
Dominico et al.\cite{data_allocation} discusses the elastic allocation of multi-cores in database systems, indirectly contributing to energy efficiency by optimizing resource usage.
Zhang et al.\cite{heterogeneous_storage} focuses on a distributed system operating on a heterogeneous CPU-GPU cluster, optimizing resource allocation for energy efficiency.
Anna\cite{anna} discusses the auto scaling of cloud storage, which can lead to energy savings by optimizing storage resources based on demand.
Caribou\cite{Caribou} focuses on intelligent distributed storage, indirectly contributing to energy efficiency through optimized storage operations.

\subsection{Energy-Efficient Data Management}

Data management encompasses the practices and tools used to ensure high-quality, accessible, and timely data. It addresses challenges in storage optimization, cloud configuration, and performance tuning, among others. In the following sections, we review three key design components in data management system: \textbf{System Design, Cloud Configuration and Performance Tuning}.

\subsubsection{System Design}
The data storage optimization and management focuses on enhancing the efficiency and performance of data storage, retrieval, and maintenance. This category delves into the intricacies of storage hierarchies, structures, and techniques to ensure optimal data handling.

Appuswamy et al.\cite{fiveminute} revisits the five-minute rule in the context of storage hierarchies, emphasizing its historical significance and relevance in the modern era of DRAM and HDD.
Chen et al.\cite{hybridLSM} addresses the write amplification issue in LSM-trees and introduces a method leveraging NVM to optimize the balance between update and search efficiency.
Proteus\cite{Proteus} is an adaptive distributed database system tailored for mixed workloads, dynamically adjusting its storage layout for optimal performance.
ADOC\cite{ADOC_LSM} tackles the write stall challenges in LSM-KV systems and introduces Automatic Data Overflow Control, a tuning framework that harmonizes data flow to mitigate data overflow issues.

\subsubsection{Cloud Configuration}
Cloud configuration and optimization revolve around the strategies and techniques to enhance the performance, cost-efficiency, and adaptability of cloud-based systems. This section explores the challenges and solutions in tailoring cloud environments to specific workloads and objectives.

Bilal et al.\cite{balckboxconfig} evaluates various black-box optimization algorithms, highlighting their effectiveness in cloud configuration tasks.
Cosine\cite{cosine} is a key-value storage engine that is self-designed with cloud-cost optimization. It can dynamically adapt its architecture to workloads, cloud budgets, and performance goals.
Moneyball\cite{azure_moneyball} discusses proactive auto-scaling in Azure SQL Database, emphasizing strategies to predict and optimize resource allocation in response to varying workloads.
CDBTune\cite{cdbtune} stands as an efficient system for automatic cloud database tuning, employing deep reinforcement learning techniques. It can autonomously tune cloud database configurations, ensuring optimal performance across diverse scenarios.

\subsubsection{Performance Tuning}
Performance tuning is centered on the methodologies and tools that allow systems to self-adjust and optimize their operations based on prevailing conditions and workloads. This section underscores the importance of resilience and adaptability in data management systems.

Endure \cite{robust_tuning} presents a robust tuning approach for LSM trees when faced with workload uncertainty, emphasizing strategies to optimize performance even under uncertain and fluctuating workloads.
QTune\cite{QTune}, a system employing deep reinforcement learning to efficiently tune database configurations, addressing the need for fine-grained, query-level tuning in diverse environments. By considering rich features of SQL queries, it achieves better performance. 
CGPTuner\cite{CGPTuner} introduces a novel approach to DBMS configuration auto-tuning, focusing on optimizing the entire IT stack to enhance performance and manage costs effectively. The method swiftly identifies and adapts well-performing configurations to workload variations without depending on a knowledge base, addressing the challenges posed by the multitude of tunable parameters and their inter-dependencies across different layers. 

\subsection{Energy-Efficient Data Analysis}

As data storage and management solutions evolve to embrace energy-efficient practices, it is imperative that the same principles extend to data analysis processes. Energy-efficient data analysis not only aligns with sustainability goals but also contributes to cost savings and reduced environmental impact. This section explores various strategies and techniques for achieving energy efficiency in data analysis while building upon the foundations laid in the preceding sections.

\subsubsection{Learning based Query Optimization}
Learned optimizers represent an advanced approach to achieve energy-efficient data analysis. These learned optimizers adapt to changing workloads and data distributions, selecting the most energy-efficient processing paths.
Additionally, specialized hardware designed for specific data analysis tasks can significantly boost processing speed while maintaining energy efficiency. 

LEON\cite{ChenCLLWZSZ23} introduces a framework for ML-aided query optimization, enhancing expert query optimizers through machine learning and fundamental knowledge. It employs a pairwise ranking objective for ML model training and employs a ranking and uncertainty-based exploration strategy. LOGER\cite{ChenGCT23} presents a learned optimizer that utilizes deep reinforcement learning for efficient and robust query plan generation. It employs a Graph Transformer to capture table and predicate relationships, optimizing the search space and restricting specific operators. BASE\cite{ChenWLLZDZSZ23} introduces a two-stage reinforcement learning-based framework to optimize queries by bridging the gap between cost and latency considerations. This approach transfers the reward function to achieve superior performance compared to traditional DBMS. Lero\cite{ZhuCDCPWZ23} introduces a learning-to-rank query optimizer that builds upon native query optimizers and continuously improves optimization performance. 

Hybrid Query Optimization has been widely studied in the field of database management systems (DBMS). It refers to the combination of various query optimization techniques to achieve efficient query execution. The goal is to minimize the time and resources required to retrieve data from a database.
Yu et al.\cite{hybrid_query_opt} proposes a method which introduces a hybrid query optimizer that combines the strengths of traditional cost-based and learning-based optimizers, emphasizing energy-efficient query plan selection. This approach generates high-quality candidate plans by leveraging learning-based hints and supplements them with cost-based methods. Queryformer\cite{queryformer} is a tree-structured Transformer model, addresses limitations in existing query plan representation methods by incorporating database statistics and effectively modeling information flow within query plans, resulting in significant performance improvements in various database optimization tasks.

\subsubsection{In-Memory and Hardware Acceleration}
In-memory data processing and hardware acceleration are pivotal strategies for enhancing energy efficiency in data analysis. Storing frequently accessed data in memory reduces the need for disk I/O operations, resulting in faster query execution and lower energy consumption. Additionally, specialized hardware accelerators can dramatically boost processing speed while optimizing power usage.

NVQuery\cite{data_allocation2} presents NVQuery, a nonvolatile memory-based query accelerator that efficiently performs various basic query functions in memory, utilizing the analog properties of nonvolatile memory. It achieves a remarkable performance speedup and energy savings compared to traditional processors, with further energy-efficient gains through configurable approximation, making it a significant improvement over state-of-the-art query accelerators. ReSQM\cite{ReSQM} introduces ReSQM, a novel ReRAM-based accelerator that leverages in-situ computing with nonvolatile memory to dramatically reduce response times for database operations, achieving substantial efficiency improvements ranging compared to traditional processors across various query types. Additionally, it outperforms state-of-the-art CAM, GPU, FPGA, NDP, and PIM solutions with speedups ranging. The Pliops Extreme Data Processor (XDP)\cite{hardware_processor} is introduced as a customized hardware-based storage engine. XDP aims to optimize various cost metrics, overcoming previous limitations in storage space, recovery time, and performance penalties for database operations.

\subsubsection{Quantum-Inspired Optimization and Sketch-Based Techniques}

QQuantum-inspired optimization techniques and sketch-based methods are emerging as innovative ways to achieve energy-efficient data analysis. Drawing inspiration from quantum computing principles, these approaches employ probabilistic algorithms and sketch data structures to approximate query results with remarkable efficiency. By reducing the computational load and data transfer requirements, they contribute to substantial energy savings in data analysis processes.

M.Sch{\"{o}}nberger et al.\cite{Schonberger22} proposes a method which assesses the feasibility of applying quantum computing to key database query optimization problems, exploring the potential of gate-based quantum systems and quantum annealers for energy-efficient optimization. COMPASS\cite{IzenovDRS21} introduces a novel query optimization approach for in-memory databases using Fast-AGMS sketches. This technique addresses issues with traditional cost-based optimization by incrementally composing sketches over the query join graph.

SEAL\cite{FeiL000J21} presents a novel data compression approach tailored for causality analysis in enterprise logs, achieving lossless compression and near real-time retrieval of historic events. It optimizes causality graph-based compression and opportunistic decompression, resulting in a data size reduction, with queries performing faster on the compressed dataset while maintaining query result consistency.

\subsubsection{Streaming and Specific Query Processing}
Stream query optimization and data processing strategies are important for real-time and energy-efficient data analysis. In scenarios where data arrives continuously, such as IoT applications and streaming platforms, optimizing query execution becomes imperative to minimize energy consumption.

TiLT\cite{JayarajanZSP23} introduces an intermediate representation (IR) called TiLT that offers an expressive temporal query language for efficient query optimization and parallelization in stream processing engines (SPEs). In contrast to contemporary state-of-the-art SPEs, TiLT attains notably enhanced throughput by rectifying the constraints within prevailing SPE design choices. It demonstrates improvements of up to 326x in various real-world streaming analytics applications.

CompressDB\cite{0007WZZCL022} introduces a storage engine that supports data processing in databases without the need for decompression. It enables operation pushdown to storage for efficient data query and manipulation. Chukonu et al.\cite{YuFCLSWZZC21} effectively mitigate integration overhead and surpass the performance of existing pure native big data frameworks. They employ optimization techniques such as operator fusion, compaction, and vectorization to minimize integration overhead.

 Zhang et al.\cite{ZhangIM0GLFHPJ22} proposes a method which discusses the deployment of QO-Advisor, a system for steering a query optimizer towards better plans tailored to specific analytical workloads. QO-Advisor is currently enabled by default for production SCOPE workloads at Microsoft, offering improved query optimization for complex and heterogeneous analytical scenarios.
 
Incorporating energy efficiency into the data analysis phase of the data lifecycle completes the journey toward greener data management. By synergizing the principles of green storage and efficient data management with adaptive analysis algorithms and cloud-based energy optimization, organizations can reduce their carbon footprint, lower operational costs, and contribute to a more sustainable future while deriving valuable insights from their data.

\subsection{Conclusion}
The rapid growth of data has made it crucial to consider both performance and sustainability in data storage and management. Energy-efficient solutions are being developed to address these dual needs. In this section we briefly introduce the solutions span various aspects of data storage, from optimizing resource utilization in cloud databases to hardware-software co-design for energy efficiency. The same principles are being extended to data analysis, where techniques like hybrid query optimization and machine learning are being employed to make the process more energy-efficient. By adopting these energy-conscious practices, organizations can not only improve performance but also contribute to sustainability, reducing both operational costs and environmental impact.

\section{Green Large Language Models}

\subsection{Training Optimization}

\subsubsection{Parameter-efficient Training}
\label{peft}
Using large language models as initialization has become the dominant paradigm in natural language processing, demonstrating impressive performance on various tasks. Typically, researchers adapt general-purpose large language models to specific target tasks by fine-tuning all model parameters, a process known as full fine-tuning. However, the large number of parameters also contributes to costly adaptation wall-clock time. Moreover, this leads to separate copies of full fine-tuned model parameters for each task, resulting in high storage costs when the intelligence system serves a large number of tasks. As current parameter scale of the model ranges from hundreds of millions~\cite{kenton2019bert} to tens of billions~\cite{radford2018gpt} or even trillions~\cite{brown2020language}, these challenges become more severe.
To address this, methods for parameter-efficient fine-tuning have become a current research focus.They aim to strike a balance between model performance and computational costs by updating only a subset of the model parameters~\cite{GuoRK20Diffprun, FuYSLBC23Effective} or small modules injected into the layers~\cite{HoulsbyGJMLGAG19Adapter, LesterAC21Prompt}. They provide effective alternatives to full fine-tuning, enabling efficient adaptation of large language models for specific tasks. In the following discussion, we summarize the latest advancements in parameter-efficient fine-tuning.

Given the training data $\mathcal{D}$ and a pre-trained model $\Theta=\left\{w_1, w_2, \ldots, w_N\right\}$, the objective of model adaptation is to produce a new model $\Theta^{\prime}=\left\{w_1^{\prime}, w_2^{\prime}, \ldots, w_M^{\prime}\right\}$ which minimizing the loss function $f$ on $\mathcal{D}$. Define $\Delta \Theta=\Theta^{\prime}-\Theta$ as the parameter change from original model $\Theta$. In full fine-tuning, all parameters in $\Theta$ are updated and formulated as $N=M$ and $\Delta \Theta=\nabla f_{\Theta}(\mathcal{D})$. But in parameter-efficient tuning, $\Delta \Theta$ specifically refers to the modification of a small number of parameters, such as the inclusion of an adapter~\cite{HoulsbyGJMLGAG19Adapter} or prompt embedding~\cite{LesterAC21Prompt}. Usually $|\Delta \Theta| \ll|\Theta|$, where $|\cdot|$ indicates the number of parameters involved. Based on the categorization proposed in ~\cite{abs-2203-06904survey}, parameter-efficient methods can be classified into three types based on the type of delta parameters: addition-based, reparameterization-based, and specification-based approaches.

\textbf{Addition-based methods}. They incorporate additional trainable neural modules or parameters upon the original model. The size of delta parameters is determined by the additional structure. We will discuss two representative branches of addition-based methods: adapter-based tuning and prompt-based tuning.

\textit{Adapters-based Tuning.} Adapter-based methods involve adding small, task-specific neural networks (adapters) between the pre-trained layers of the language model, allowing adaptation without the need for extensive retraining of the entire model. In the classic adapter module~\cite{HoulsbyGJMLGAG19Adapter}, an input $\boldsymbol{h}$ is projected to space with a lower dimension using a down-projection matrix $\boldsymbol{W}_d \in \mathbb{R}^{d \times r}$, followed by an activation function $f(\cdot)$, and then projected back to the original dimension using an up-projection matrix $\boldsymbol{W}_u \in \mathbb{R}^{r \times d}$. The adapter is combined with a residual connection, resulting in the following form:
$$
\boldsymbol{h} = f\left(\boldsymbol{h W}_d\right) \boldsymbol{W}_u+\boldsymbol{h} .
$$
Houlsby et al.\cite{HoulsbyGJMLGAG19Adapter} inserted two adapters within each layer of the model, one after the multi-head self-attention and another after the feed forward network (FFN). In contrast, Pfeiffer et al.\cite{PfeifferKRCG21Adapterfusion} introduced a more efficient variant of adapters that is inserted solely after the "add \& layer norm" modules, achieving similar performance with fewer parameter overheads. By using adapter tuning, the number of tunable parameters per layer is reduced to $0.5 \% \sim 8 \%$ of the whole model during tuning process.

Although adapters have used significantly fewer tunable parameters compared to full fine-tuning, some studies attempts to further improve its efficiency on multi-task setting by modifying the structure of the adapter layer.
Compacter~\cite{MahabadiHR21Compacter} proposes a combination of adapters and low-rank optimization using hypercomplex multiplication and parameter sharing. Specifically, Compacter parameterizes the linear layer as the sum of the Kronecker product of two small matrices, reducing the parameter cost of multitask learning.
AdapterFusion~\cite{PfeifferKRCG21Adapterfusion}, on the other hand, introduces the vector representation of pre-trained task-specific adapters, and thus increases the transfer of knowledge across tasks. 
HyperFormer~\cite{MahabadiR0H20Hnet} generates adapter parameters using a shared hypernetwork. These hypernetworks learn adapter parameters for all layers and tasks conditioned on tasks, adapter positions, and layer IDs. It shares knowledge among tasks through the hypernetwork while adapting the model to each individual task through task-specific adapters.
Adapter tuning has also been successfully used in fields such as vision~\cite{ChenGTWSWL22AdaptFormer}, vision-language~\cite{Sung0B22VLadapter}, and image-video~\cite{PanLZS022STadapter}.

\textit{Prompt-based method}. Instead of introducing neural modules into the Transformer model, prompt-based methods incorporate additional context by wrapping the original input. These methods train the model to understand and generate responses based on specific prompts or instructions, and have been utilized in various natural language processing tasks~\cite{LiuYFJHN23prompt}.
One influential contribution in this research area is prefix-tuning~\cite{LiL20Prefixtuning}, which prepends trainable continuous vectors (prefixes) to the keys and values of the multi-head attention. Each prefix is initialized as a trainable parameter matrix $\boldsymbol{P}$. The pre-trained key $\boldsymbol{K}$ and value $\boldsymbol{V}$ are concatenated with two corresponding prefix matrices $\boldsymbol{P}_k, \boldsymbol{P}_v \in \mathbb{R}^{l \times d}$. Specifically, the output of multi-head attention layer changes to:
$$
\boldsymbol{h} =\operatorname{MSA}\left(\boldsymbol{h}_q, \left[\boldsymbol{P}_k; \boldsymbol{h}_k\right], \left[\boldsymbol{P}_v; \boldsymbol{h}_v \right]\right).
$$

While prefix-tuning adds tunable matrices to every intermediate Transformer layer, prompt tuning~\cite{LesterAC21Prompt} proposes a simpler approach for incorporating prompts into the input data:
$$
\boldsymbol{h} = \operatorname{LM} \left(\left[\boldsymbol{P}; \boldsymbol{X}_k\right]\right).
$$
Throughout the training process, the parameters of the prompts are updated through gradient descent while the model itself remains unchanged. Notably, as the model size expands, the performance disparity between prompt tuning and full parameter fine-tuning gradually decreases~\cite{LesterAC21Prompt}. Related work in this area includes P-tuning~\cite{LiuJFTDY022Ptuning}. Additionally, prompt tuning has demonstrated transferability across tasks through techniques such as hypernetwork~\cite{HeZTGDAZLCMCC22Hprompt}, decomposition~\cite{WangPKF0K23Mprompt}, and attention mechanisms~\cite{AsaiSPH22Attempt}. By using prompt-based tuning, the number of tunable parameters is reduced to less than $0.1 \%$ of the whole model during tuning process, but may sometime facing the optimizing difficulty~\cite{abs-2203-06904survey}.

\textbf{Reparameterization-based methods}. Another set of methods adopts a reparameterization approach to transform existing parameters into a parameter-efficient form. This approach is motivated by the observation that a relatively low-dimensional matrix can achieve performance similar to fine-tuning in large-scale models~\cite{AghajanyanGZ20InstrDim}. Therefore, it is reasonable to optimize only the compressed parameters, thereby reducing computational and memory costs.

One well-known method in this category is LoRA~\cite{HuSWALWWC22Lora}, which hypothesizes that the changes during model tuning exhibit a low intrinsic rank. LoRA adapts to new tasks by optimizing lower-rank matrices in the self-attention modules, which aims to capture the most significant changes while discarding the less important ones. 
Specifically, for a original weight matrix $\boldsymbol{W} \in \mathbb{R}^{d \times k}$, LoRA approximated its change during tuning using a low-rank decomposition:
$$
\Delta W= \boldsymbol{W}_{d} \boldsymbol{W}_{u}.
$$ 
Here, $\boldsymbol{W}_d \in \mathbb{R}^{d \times r}, \boldsymbol{W}_u \in$ $\mathbb{R}^{r \times k}$ are low-rank tunable parameters. LoRA applies this update to the query and value projection matrices $\left(\boldsymbol{W}_q, \boldsymbol{W}_v\right)$ in the multi-head attention sub-layer. By doing so, LoRA achieves fine-tuning performance comparable to that of the GLUE benchmark, while reducing the number of tunable parameters to just 1\% of the original count. The effectiveness of the LoRA method is demonstrated on various scales and architectures of pre-trained language models. To account for the significance of different weight parameters, AdaLoRa~\cite{zhang2023adaptive} dynamically assigns the parameter budget to decomposing matrices based on their importance scores.
In contrast, Aghajanyan et al.~\cite{AghajanyanGZ20InstrDim} optimize directly in a low-dimensional intrinsic subspace of the complete model for each task., while Qin et al.~\cite{qin2022exploring} further explore learning multiple tasks within a unified low-dimensional intrinsic subspace.

\textbf{Specification-based methods}. 
In these approaches, only a specialized subset of the pre-trained model's parameters are fine-tuned, while keeping the majority of the parameters frozen. 
It aims to optimize a small number of internal task-specific parameters to solve tasks without altering the internal structure of the model, i.e., $|\Theta| = |\Theta^{\prime}|$.

BitFit~\cite{ben-zaken-etal-2022-bitfit} optimizes exclusively the bias terms within the model while keeping other parameters fixed and yields impressive performance surpassing 95\% on diverse benchmarks.
Diff pruning~\cite{GuoRK20Diffprun} introduces a parameter selection mask modeled as a Bernoulli random variable and optimizes this variable using a reparametrization method. 
$($IA$)^3$~\cite{liu2022fewshot} scaling activations by learned vectors to achieve sparse fine-tuning and maintain stronger performance while updating up to 10,000$\times$ fewer parameters. 
Fu et al.~\cite{FuYSLBC23Effective} propose a novel Second-order Approximation Method (SAM) to better select tunable parameters. SAM approximates the selection problem with an analytically solvable optimization function and determines the tunable parameters by directly optimizing the approximation function.
Other works like ROME~\cite{meng2022locating} and MEMIT~\cite{meng2023mass} explore updating neurons with respect to specific error samples while leaving other outputs unchanged, using causal tracing to identify relevant FFN layers. ROME focuses on editing the top1 FFN layer, while MEMIT edits multiple layers.

\subsubsection{Continual learning}

AAs the LLM advanced from the large scale of model parameter and huge amount of training corpora, the computational resources required to retrain them from scratch have become prohibitively expensive. Additionally, accessing previously learned unlabeled data is not always feasible due to high memory requirements and data privacy concerns. Continual learning enables pre-trained models to be trained solely on new incoming data, allowing them to enhance their ability to handle evolving language or emerging domains without forgetting the knowledge acquired in the past in an efficient way.

Previous approaches to continual learning have primarily focused on models that are relatively small and trained from scratch. However, with LLMs possessing excellent generalization ability and robust representation power, their emergence brings new opportunities and challenges for continual learning. To this end, several continuous learning methods specifically targeting LLMs have emerged on top of conventional continual learning methods.

\textbf{Generative replay methods.} 
As an LLM is intrinsically a text generator with exceptional generative capabilities, it can solve downstream tasks while generating pseudo-samples of the previous tasks used for memory replay. The generated pseudo-samples remind the current model with prior task knowledge and alleviate forgetting. 
Sun et al.~\cite{sun2019lamol} first implement this method and call it LAMOL. In contrast to previous approaches, LAMOL optimizes a single model without the need for an additional generator, utilizing both data generation loss $\mathcal{L}^{\operatorname{lm}}$ and task optimization $\mathcal{L}^{\text {task}}$:
$$
\mathcal{L}^{\text {task }}=-\sum_{i=1}^n \log p\left(\boldsymbol{Y}_i \mid [\left[\text{TASK}; \boldsymbol{X}_i\right], \theta\right),
$$
$$
\mathcal{L}^{\operatorname{lm}}=-\sum_{i=1}^n \log p\left(\left[\text{GEN}; \boldsymbol{X}_i; \boldsymbol{Y}_i\right] \mid t, \theta\right).
$$
Here, 'TASK' and 'GEN' are special tokens append before the input to promote the model with different behavior. The LM trains on the mixture of current task and pseudo data generated give the token 'GEN'. It demonstrates comparable result with multitask learning and requires no extra memory or model capacity. With the advance in simplity and effectiveness, several works combine it with other type of continual learning methods and achieve better results. Rational LAMOL~\cite{kanwatchara2021rational} adopt critical freezing when update model to prevent forgetting, i.e., first identify the important block for old task and freeze when training. L2KD~\cite{chuang2020lifelong} utilize knowledge distillation to prevent forgetting of data generator. Except distillation, LFPT5~\cite{qin2021lfpt5} uses the soft prompt to enhance the memorization of data distribution for each task.

\textbf{Parameter-efficient tuning methods.} Parameter-efficient continual fine-tuning methods are the most popular approaches for continual learning of LLMs. These methods typically rely on additional-based parameter-efficient tuning techniques, such as Adapters and Prompts. These lightweight modules leverage the high generalization ability provided by pre-trained models, effectively reducing the cost of transferring to downstream tasks and enabling adaptation while freezing the model backbone to retain pre-trained knowledge and prevent forgetting. However, Adapters and Prompts are primarily designed for task-specific adaptation, resulting in a linear increase in memory cost with the number of tasks and no knowledge sharing across tasks. The objective of parameter-efficient continual fine-tuning methods is to address these challenges and mitigate forgetting when learning new tasks.

Madotto et al.~\cite{madotto2021continual} independently learn adapter for each task in a sequential manner to avoid forgetting and select the most confident adapter with the lowest perplexity during inference to generate the output. ELM~\cite{jang2023adapterensemble} finds that merging separately trained adapters showcased compositional capabilities, which were then applied to continual instruction tuning. To enhance efficiency, ADA~\cite{ermis2022ADA} limits the size of adapter during sequential learning. They train a new adapter for coming task, but distill it with one of the adapters in the pool that possessed transferable information, reducing the number of adapters to be stored. Zhang et al.~\cite{zhang2022continual} use the neural architecture search technology to automatically add and compact adapters in each layer when learning a new task. Conversely, Ke et al.~\cite{ke2021achieving, ke2021adapting} build a new adapter structure for continual learning,incorporating a knowledge sharing module and a task-specific module. The approach utilized a masking mechanism to prevent forgetting in important neurons and transferred knowledge through a task router in the sharing module. In terms of prompt-based methods, L2P~\cite{wang2022learning} and DualPrompt~\cite{wang2022dualprompt}) establish a pool of prompts that can be selected for insertion into the model to perform task-specific ability. They achieve this by creating a matching mapping between input data and prompts using a clustering-like optimization approach. These methods keep the language model parameters frozen and rely on prompts to capture task information, avoiding forgetting during continual learning and enabling generalization to multiple tasks. However, as these methods depend on a key and query system to select prompt indices from the pool, they cannot be optimized in an end-to-end manner. In contrast, CodaPrompt~\cite{smith2023coda} introduces an attention-based end-to-end mapping scheme, where a set of prompt components is learned and then weighted and combined using input-conditioned weights.

\subsection{Decoding Optimization}

Accuracy-lossless acceleration has recently been proposed as a solution to the inefficiency of the auto-regressive decoding strategy. In their work, \cite{blockdecoding} proposed a blockwise parallel decoding strategy consisting of three steps: predict, verify, and accept. In the prediction step, a modified transformer model is used to predict the next tokens for each subsequent position. This is achieved by inserting a multi-output feedforward layer with residual connections after the original decoder output layer. In the verification step, the output tokens of the original decoder are compared with the proposals from the previous steps, and the longest matching prediction is selected. In the acceptation step, multiple tokens are generated in parallel, and new hypotheses from the verification step are added to the decoding input. \cite{blockdecoding} also proposed fine-tuning and distillation strategies to train the extra feedforward layers.

While the blockwise parallel decoding strategy offers remarkable acceleration, it requires model training to obtain the extra feedforward layer. Furthermore, the prediction of tokens after the next token only uses information from the input tokens, which can lead to inaccuracies and frequent failures in the verification step. To address these drawbacks, speculative decoding \cite{xia2023speculative} has been proposed. This approach uses a small and fast model for prediction. The small model can be an off-the-shelf model, e.g., the Bloom model \cite{workshop2023bloom} of size 7.1B can be served as the small model for the 176B model. However, the small model may increase inference time significantly. To mitigate this, parallel decoding can be used for the small model to achieve further acceleration \cite{ge2022lossless}. If no available small models can generate similar sentence pieces as a large model, model-free prediction strategies are proposed \cite{ge2022lossless, Yang2023InferenceWR}. Specifically, \cite{ge2022lossless} presents an input-guided method that copies content from the input sentence using prefix matching, while \cite{Yang2023InferenceWR} utilizes a prefix matching strategy to retrieve content from the input sentence or a document database. These methods are used in greedy decoding without sacrificing accuracy, and they can achieve better performance gains through approximate inference, where the criterion used during verification is relaxed by accepting predictions within the top-k scores \cite{blockdecoding}.

For models that employ non-greedy decoding strategies, \cite{chen2023accelerating} proposes a modified rejection sampling scheme that preserves the distribution of the target model. Instead of auto-regressive decoding, another approach for fast inference is non-autoregression strategy, which achieves comparable accuracy. This strategy, known as non-autoregressive translation (NAT) \cite{gu2018nonautoregressive}, is primarily used in translation tasks \cite{kasai2021deep, Huang2021NonAutoregressiveTW, saharia-etal-2020-non}. However, there are significant differences between translation tasks and general language model (LLM) scenarios, which may result in poor performance of these methods for LLM decoding. Therefore, we provide only a brief introduction to the related works. \cite{Huang2021NonAutoregressiveTW} introduces a layer-wise iterative method, where each layer uses the decoding results and embeddings of the previous layers. Each layer is trained using maximum likelihood estimation for the prediction of every decoding layer. \cite{santilli2023accelerating} formalizes the standard greedy autoregressive decoding strategy with a parallel Jacobi and Gauss-Seidel fixed-point iteration. It initializes the next tokens with special tokens and performs iterative decoding until convergence.

\section{Applications of Green Computing}

\subsection{Green Computing For Environment}

Green computing emphasises that AI should not only be energy efficient in its own development and operation to achieve green technologies, but should also play an active role in a variety of green application areas to address environmental and sustainability challenges. With its ability to handle large and complex data, AI has increased the efficiency of data analysis, modelling and prediction, and improved productivity in various fields. In environmental governance, it often involves large amounts of monitoring data \cite{kibirige2023using}, remote sensing data \cite{dupuis2020can,hanan2020satellites}, meteorological data \cite{muthukumar2022pm2}, and so on. Clearly, AI can effectively extract useful information from these sources, identify trends, anomalies and patterns to make predictions and provide guidance for decisions and actions in areas such as air pollution monitoring \cite{yao2022spatiotemporal}, carbon sequestration estimation \cite{santoro2020global,sannigrahi2020examining,philipp2021trends}, carbon price forecasting and many others \cite{zhang2022novel}.

\subsubsection{Air Pollution Emission Monitoring}

\textbf{\numberedParagraph{Introduction}}

Air pollution is one of the major issues facing the world today, with serious impacts on human health\cite{lelieveld2019cardiovascular}, ecosystems and climate change\cite{masson2018global}. According to data published by the World Health Organization (WHO), about 7 million people die each year from air pollution-related diseases, including cardiovascular diseases, respiratory diseases, and lung cancer\cite{world2021global}. In addition, air pollution can lead to respiratory problems such as premature births in children, reduced lung function, chronic bronchitis and asthma. As for the ecosystem, high concentrations of air pollutants, such as ozone (O3) and sulphur dioxide (SO2), negatively affect plant growth and vegetation cover, while deposition of pollutants into the soil and water bodies can cause soil acidification and toxicity to aquatic organisms, disrupting the ecological balance. Considering that the emission of greenhouse gases contributes significantly to climate change, the presence of gases such as carbon dioxide (CO2) and methane (CH4) in air pollution further intensifies the issue of global warming. 

According to a report by the United Nations Environment Program\cite{christensen2019emissions}, an increase in greenhouse gas emissions will lead to a global average temperature increase of more than 3 degrees Celsius. Air pollution not only poses serious risks to human health, but also negatively affects ecosystems and climate change. Accurate knowledge and calculation of air pollutant emissions is therefore a critical step in the development of response measures and policies. However, traditional pollutant emission calculation methods usually estimate and predict pollutant emissions based on known source information, emission concentration formulas, and other relevant factors combined with a priori knowledge of pollutant transport and transformation. The calculation process of traditional methods usually requires manual involvement, including collecting source data, consulting formulas and guidelines, and performing calculations manually. The main disadvantages of this method are that it is cumbersome, time-consuming and prone to errors. Since air pollutant emissions are affected by a variety of factors, including meteorological conditions, human activities, national policies, etc., it is difficult for traditional calculation methods to take into account and synthesize these complex factors.

Over the recent period, the evolution of Artificial Intelligence (AI) has unveiled fresh avenues for computations related to air pollution emissions. Bakay et al. \cite{bakay2021electricity} used a dataset from the Turkish power production industry and, by applying AI algorithms such as deep learning, predicted the emissions of greenhouse gases. The results showed that all algorithms provided satisfactory predictions and the rRMSE values were less than 10\%. Mao et al. \cite{mao2021hybrid} developed a GT-LSTM model, combining a graph convolutional network with a time-sliding LSTM, to forecast air pollutant levels. Results showed its ability to extract spatio-temporal features and achieve high prediction accuracy and stability. Using machine learning and deep learning algorithms, Artificial Intelligence (AI) models can automatically learn from large amounts of sample data, capture these complex correlations hidden in the data, and calculate air pollutant emissions. Remote sensing tools provide the ability to measure air pollution emissions over large areas, while the spatial distribution of on-site monitoring stations is usually fragmented, with monitoring stations kept separated from each other, resulting in missing data over this space. 

In order to study the impact of the COVID-19 closure measure on emission reduction intensity and chemical sensitivity in eastern Asia, Ghahremanloo et al. \cite{ghahremanloo2021impact} analyzed the concentration and aerosol optical thickness of four major pollutants (nitrogen dioxide (NO2), formaldehyde (Formaldehyde), sulphur dioxide (SO2), and carbon monoxide (CO)) from satellite data. Ghahremanloo et al. also used the Global Land Data Assimilation System (GLDAS) to obtain meteorological parameters, which were compared and analyzed with satellite imagery, demonstrating the feasibility and validity of satellite remote sensing data in the monitoring of air pollutant emissions. Satellite remote sensing technology, as a means of remote monitoring, can provide atmospheric environmental data on a global scale, bringing new opportunities for air pollution emission calculations. Using machine learning and deep learning algorithms, Green Environment AI can process and analyze large amounts of satellite remote sensing data to efficiently extract information about pollutant emissions. This application not only helps to reduce energy consumption and carbon emissions, but also realizes real-time monitoring of air pollutant concentrations, meteorological conditions, etc., which provides support for pollution early warning and emergency response, and is an important development direction for future air pollutant emission calculations.

\textbf{\numberedParagraph{Air Pollution Monitoring Satellite}}

\textbf{Aura} which signifies "air" in Latin, was launched into orbit on July 15, 2004. It is a scientific observation satellite developed collaboratively by multiple national aerospace agencies. Following Terra and Aqua (which carry the MODIS sensor), Aura is another significant satellite in the Earth Observing System (EOS). Its primary mission is to conduct observations and research related to Earth's ozone layer, air quality, and climate change. Aura is in a near-polar, sun-synchronous orbit with a design lifespan of 6 years. It completes approximately one orbit around Earth in about 100 minutes, resulting in a repeat observation cycle of 16 days. The orbital inclination is approximately 98.2 degrees, with a local equator-crossing time of 1:45 PM. It completes 14 to 15 orbits around Earth per day. 

\textbf{METOP} In 2006, the GOME-2 instrument was successfully launched aboard the METOP-A satellite, followed by another launch in 2012 aboard the METOP-B satellite, both of which are part of the European Space Agency's meteorological satellite program. GOME-2 functions within the ultraviolet, visible, and near-infrared spectral ranges, covering wavelengths from 240 to 790 nanometers, and offers a spectral resolution ranging from 0.2 to 0.4 nm. It has a swath width of 1920 km and provides a ground resolution of 80×40 km² at nadir. The local equator-crossing time is 9:30 AM, enabling global coverage in just one day.

\textbf{GOSAT} GOSAT, On January 23, 2009, Japan's GOSAT project took flight. It is the first satellite specifically designed to measure concentrations of CO2 and CH4 near the Earth's surface with high sensitivity. Its 666-kilometer orbit enables it to complete a full orbit roughly every 100 minutes, resulting in global coverage in approximately three days. On GOSAT, the TANSO-FTS instrument employs three SWIR bands at 0.76, 1.6, and 2.0 $\upmu$m to deliver column measurements of carbon dioxide and methane with high sensitivity near the Earth's surface, while its fourth infrared (TIR) band spans from 5.5 to 14.3 $\upmu$m.

GOSAT-2, the Greenhouse Gases Observing Satellite-2, represents a significant leap forward in our ability to monitor and analyze Earth's environment. Positioned in a sun-synchronous orbit at an altitude of approximately 613 kilometers, GOSAT-2 offers comprehensive global coverage. Scientists and researchers worldwide rely on GOSAT-2's data to gain insights into Earth's carbon cycle, support climate agreements, and advance our efforts to combat climate change.

\textbf{OCO} OCO-2, the Orbiting Carbon Observatory-2 satellite, represents a groundbreaking asset in the realm of Earth observation. Launched on July 2, 2014, OCO-2 is equipped with cutting-edge technology and advanced instruments, positioning it as a vital player in the monitoring of Earth's carbon cycle. With a revisit period of 16 days, it provides narrow coverage down to 10.3 kilometers. The spatial resolution for each measurement is less than 1.29 kilometers by 2.25 kilometers, with a horizontal offset of approximately 150 kilometers between adjacent revisit orbits.

Following the successful launch of the OCO-3 satellite on May 4, 2019, it embarked on its mission to observe and analyze carbon dioxide levels in Earth's atmosphere. The primary spectrometer in OCO-3 is a spare unit originally designed for OCO-2. Unlike OCO-2, the International Space Station (ISS) does not traverse specific latitudes at a consistent local time daily; instead, it progressively covers the entire day from sunrise to sunset, shifting over 20 minutes each day. OCO-3's data include both the sunrise and sunset nodes, covering approximately 6 hours from local noon, with a revisit period of about 16 days.

\textbf{Sentinel-5P} As part of the European Copernicus program, the Sentinel-5P satellite was launched in October 2017. It serves as a dedicated satellite for the monitoring of atmospheric chemical composition within the context of global environmental and security initiatives. The satellite is equipped with the Tropospheric Monitoring Instrument (TROPOMI), which boasts higher spatial resolution (7x7 km2) and a broader spectral range (270-2385 nm) compared to existing instruments. It also features improved signal-to-noise ratios and a scanning width of 2600 km. These enhancements allow for more accurate measurements of atmospheric components such as O3, NO2, SO2, and others, and open up broader application prospects for studying urban-scale issues.

\textbf{\numberedParagraph{Satellite-Based Prediction of Gaseous Pollutant Concentrations}}

This section takes the concentration prediction of three gaseous pollutants, namely NO2, SO2, and O3, as examples to illustrate the significant role of satellite remote sensing and artificial intelligence in estimating gaseous pollutant concentrations.

\textbf{NO2 concentration prediction} Currently, multiple techniques are in use to measure and estimate ground-level nitrogen dioxide (NO2) concentrations. Ground monitoring stations offer precise and continuous data for specific locations, but their coverage is limited and may not capture broader NO2 concentration changes. In contrast, remote sensing technologies, like satellite imagery and drones, provide extensive spatial coverage and have been widely employed for high-resolution, spatiotemporal NO2 monitoring. This approach relies on measuring data reflecting atmospheric reflections or emissions, which can be used to infer NO2 concentrations. In recent times, there has been a surge in enthusiasm for employing deep learning and artificial intelligence methodologies to gauge NO2 concentrations, highlighting their unique merits. These approaches excel in their capacity to harmoniously blend an array of data sources, encompassing satellite data, Monitoring station data, and meteorological data, resulting in significantly improved precision and dependability when estimating NO2 concentrations.

Li et al. \cite{li2021spatiotemporal} used a complete residual deep network to estimate missing satellite NO2 data (Aura satellite OMI) and reliably estimate high spatial resolution (1 km) ground-level NO2 concentrations, generating daily ground-level NO2 concentration products for most of mainland China. On the other hand, Scheibenreif et al. \cite{scheibenreif2022toward} proposed an innovative deep learning model that combined Sentinel-2, Sentinel-5P satellite data, and EEA ground station data. Using a deep learning data fusion approach, they successfully estimated high spatial resolution NO2 concentrations globally and quantified the model's uncertainty, thereby improving NO2 estimation in different geographical regions. Additionally, Liu et al. \cite{liu2023high} utilized the GTNNWR model, which combines neural networks' learning capabilities with the ability to consider local spatial interpretability through spatial weights. This GNNWR model has proven effective in tackling spatial heterogeneity and capturing intricate non-linear relationships in regression analyses. They used a variety of remote sensing data and ground observations to reconstruct daily NO2 concentrations at 500-meter resolution.

\textbf{SO2 concentration prediction} Ground monitoring and satellite remote sensing are commonly used methods for atmospheric sulfur dioxide (SO2) monitoring. While ground monitoring can provide accurate surface-level SO2 measurements, monitoring stations are relatively sparse and concentrated in urban areas. Moreover, developing countries like China often lack long-term monitoring data. Satellite remote sensing serves as an additional data resource, offering extensive and enduring information. Satellite retrieval enables Systematic, frequent tracking of atmospheric SO2 column density on a broad scale, but climatic and topographic conditions can affect satellite retrieval, introducing high spatiotemporal uncertainty. Many researchers have employed machine learning models or deep learning models, leveraging auxiliary variables, to estimate missing SO2 concentrations, achieving high-precision, high-resolution seamless mapping of SO2.

Zhang et al. \cite{zhang2019spatiotemporal} developed a Random Forest-Spatiotemporal Kriging model, utilizing Aura satellite Level 3 OMI SO2 products and ground monitoring data to estimate daily SO2 concentrations in China for the years 2013-2016, with a spatial resolution of 0.1°. This model combines remote sensing and ground monitoring data, providing a powerful approach for accurate estimation of atmospheric pollutants. Zhang et al. \cite{zhang2022data} proposed a Robust Backward Estimation with Data Augmentation (RBE-DA) method for bias correction, based on ground observation data, OMI inversion data, and various geographic factors, to estimate daily surface SO2 concentrations in northern China for the years 2005-2019. This method not only enhances data accuracy but also contributes to a more comprehensive understanding of spatiotemporal variations in surface SO2 concentrations. Wei et al. \cite{wei2023ground} employed the spatiotemporal Extra Trees machine learning model, using input variables such as ground monitoring data, OMI satellite inversion data, and meteorological data, to generate daily 10-kilometer concentration products of NO2, SO2, and CO for China from 2013 to 2020. This integrated use of multiple data sources offers a new approach for accurate estimation of atmospheric pollutants and provides high-resolution data in both time and space.

\textbf{O3 concentration prediction} Ozone (O3) is a toxic gas that can be harmful to living organisms due to its high oxidizing potential. Since the mid-20th century, many countries worldwide have conducted observations of tropospheric and ground-level O3 concentrations. Nevertheless, the establishment and upkeep of ground-based networks demand substantial human and financial investments, resulting in limited distribution of monitoring stations. Satellite remote sensing can complement this gap by delivering uninterrupted atmospheric O3 data with broad spatial coverage. Existing space technologies primarily offer total O3 columns, tropospheric O3, O3 profiles at different vertical ranges, while near-surface O3 usually represents only a fraction of total column O3. In some situations, data on tropospheric total column provides assistance in comprehending global and regional characteristics. However, it remains a challenging task to measure O3 values within the planetary boundary layer, particularly at exposure heights (~2m). Hence, extracting O3 concentrations from satellite measurements, especially those near the surface, is particularly challenging.

Recently, three primary approaches have been used to estimate O3 concentrations near the surface. Chemical/numerical methods typically offer extensive spatial and temporal coverage but require significant computational resources. Predictions based on any chemical mechanism have nonlinear effects on emissions and meteorology. Statistical models, chosen for their speed and simplicity, are susceptible to outliers and the impact of collinear variables, resulting in less accurate estimations. In recent years, artificial intelligence has gained significant popularity thanks to its potent data mining capabilities. When applying artificial intelligence models, spatiotemporal heterogeneity in air pollution, such as the case with O3, needs to be considered to achieve high-precision O3 concentration estimation.

Wei et al. \cite{wei2022full} extended the Spatiotemporal Extreme Trees (STET) model to estimate ground-level O3 concentrations, covering daily 10-kilometer resolution data in China from 2013 to 2020. Through time series analysis, they delved into daily and multi-year O3 pollution variations in China, providing important insights into the spatiotemporal dynamics of O3 in the atmosphere. Han et al. \cite{han2023rebuilding} adopted an ensemble approach by combining a machine learning model (XGBoost) with an air quality model (WRF-Chem) and proposed the WRFC-XGB model. The model produced a highly accurate dataset of ground-level O3 concentrations on an hourly basis. Utilizing this dataset, they assessed the impact of O3 pollution on crop yields. This study not only aids in assessing the potential risks of O3 pollution to agriculture but also provides robust support for environmental policies and agricultural decisions. Additionally, Zeng et al.'s \cite{zeng2023estimation} research employed an improved U-Net and LSTM to construct a spatiotemporal feature extraction module. Subsequently, they employed data from ground monitoring stations and satellite remote sensing imagery to construct a hybrid spatiotemporal model known as MixNet, combining point and plane data. They successfully estimated daily average O3 concentrations from 2020 to 2021 with a spatial resolution of up to 0.05°. This research offers an effective method for obtaining high-resolution O3 concentration data and is expected to further advance research on atmospheric pollution and environmental management.

These studies demonstrate the significant role of data fusion from multiple sources and artificial intelligence in the estimation of atmospheric pollutants.

\textbf{\numberedParagraph{Satellite-based prediction of particulate matter concentration}}

Particulate matter (PM) is also a type of air pollutant, which refers to tiny particles suspended in the air, usually including both PM10 and PM2.5. Among them, PM10 refers to particulate matter with a diameter less than or equal to 10 micrometers, which mainly originates from industrial emissions, road dust, building construction, etc. PM2.5 is even smaller, with a diameter less than or equal to 2.5 micrometers, which mainly originates from vehicle exhaust, coal combustion, industrial emissions and natural dust, etc. Given its minute size, PM2.5 can stay aloft for extended durations, enabling it to disperse over a greater range, and it can also penetrate the alveoli and enter the human body, with the potential to affect respiratory, cardiovascular, and neurological health \cite{calderon2016air}. Therefore, we need to monitor and control PM2.5 emissions and take effective air quality management measures to ensure a clean and healthy air supply.

Airborne particulate matter (PM) profoundly affects both climate and human health \cite{jumaah2021development}, making the precise depiction of PM's spatial and temporal spread crucial. Typically, the spatial and temporal distribution data of PM can be acquired via terrestrial sampling devices or satellite-based remote sensing. In China, several ground-based monitoring stations have been established since 2013 to monitor PM concentrations. Narkhede et al. \cite{narkhede2023deep} constructed a deep learning model to predict hourly PM2.5 concentrations using monitoring data from 2013 stations, and it can predict PM2.5 concentrations in the next 2 hours. However, it is difficult to achieve large-scale monitoring coverage over the entire region because most stations are located in urban areas and sparsely distributed \cite{gui2020construction}. In contrast, satellite remote sensing is characterized by spatio-temporal continuity and comprehensive coverage, and can achieve complementary spatio-temporal information in surface PM estimation.

In addition, chemical transport modeling (CTM) is widely used for air quality monitoring. Li et al. \cite{li2020retrieval} used GEOS-Chem to simulate PM2.5 concentrations in North China, and Park et al. \cite{park2022implementation} improved on the Community Multiscale Air Quality Model by integrating a Kalman filter for assimilating data related to ground-level PM2.5. CTM uses emission inventory data to model and estimate pollutant distributions, but these data can be incomplete or subject to error. Inaccuracies in emission inventory data can affect the accurate estimation of pollutant concentrations and lead to distorted results \cite{HoulsbyGJMLGAG19Adapter}. Due to the limitations of computational resources and model complexity, CTM is usually simulated at a coarser spatial resolution, which may lead to insufficient capture of local details \cite{yan2016improved}. Especially in areas with dense pollution sources, such as cities, the spatial resolution of CTM may not be sufficient for accurate estimation of PM2.5 concentrations.

In recent years, many studies have utilized machine learning models to simulate and estimate PM concentrations. \cite{mamic2023developing} employed Sentinel-5P along with other open-source remote sensing data accessible on the Google Earth Engine (GEE) platform to gauge the levels of PM2.5 and PM10 during both heating and non-heating periods in the Republic of Croatia , and the average R2 reached 0.7 in different seasons. Chen and his team\cite{chen2022estimation} utilized top-of-atmosphere reflectance (TOAR) data from China's FY-4A, a second-generation geostationary meteorological satellite, along with hourly PM10 atmospheric observations to establish the correlation between the TOAR data and PM10 by a specific deep learning model\cite{zhou2019deep} which resembles the structure of a deep neural network, replacing the DNN neurons with a decision tree model. 

By comprehensively analyzing and summarizing several related studies, it can be understood that the combination of satellite remote sensing and machine learning has brought about a major breakthrough in PM emission detection. Satellite remote sensing data offers high spatial and temporal resolution, providing a broad range of spatial coverage. This capability enhances monitoring capabilities, while machine learning algorithms provide an efficient and effective tool for processing and analyzing data accurately. By effectively combining these two technologies, real-time and accurate monitoring and assessment of PM emissions in different regions and at different scales can be realized. However, there are some problems and challenges in the current research. For example, aspects such as the assurance of data quality and accuracy\cite{stafoggia2019estimation} and the generalization ability of the model\cite{hu2017estimating} still need to be further explored and addressed. In addition, due to the continuous development and advancement of satellite remote sensing and machine learning technologies, the model performance can be further optimized, the monitoring range can be expanded, and it can be applied to the monitoring of other environmental pollutants in the future.

\textbf{\numberedParagraph{Conclusion}}

The study of air pollutant emissions calculation holds significant importance. Firstly, it helps us assess the potential threats of various pollutants to the atmospheric environment and human health. By understanding the emission levels of different types and sources of pollutants, we can better grasp their impact on air quality, enabling us to take appropriate mitigation measures. Secondly, accurate pollutant emission calculations are crucial for setting emission reduction targets and regulatory policies. Only by accurately understanding the sources and distribution of pollutants can we formulate targeted control measures and assess their effectiveness.

In the realm of air pollutant emissions calculation, satellite remote sensing technology plays a crucial role. Satellite remote sensing can estimate pollutant emissions by observing and analyzing gas concentrations, aerosols, and other information in the atmosphere. This technology offers advantages such as global coverage, real-time monitoring, and non-contact observation, providing extensive data over a large geographical area without geographical limitations. Through satellite remote sensing technology, we can obtain air pollutant emission data for different regions and time periods, serving as a vital reference for environmental management and protection.

The advent of intelligent AI has transformed the way we research pollutant emissions. Compared to traditional chemical transport models, the integration of advanced AI techniques with satellite remote sensing data allows for a more comprehensive and accurate understanding of pollution sources, pollutant distribution, and their environmental impacts. By utilizing advanced AI algorithms and satellite data, pollution concentrations across different seasons can be effectively estimated and monitored.

However, there are challenges in the process of using satellite remote sensing to calculate air pollutant emissions. For instance, data resolution and accuracy need further improvement to provide finer and more accurate monitoring results. Additionally, inversion models for remote sensing parameters need continuous enhancement to increase the accuracy and reliability of calculations. Furthermore, the integration of satellite remote sensing data with ground-based monitoring data to obtain a more comprehensive and holistic picture of pollutant emissions is a crucial task.

Therefore, future research efforts should focus on enhancing the precision and reliability of satellite remote sensing technology and further refining related calculation methods and models. Additionally, strengthening the fusion of satellite remote sensing data with other monitoring methods can provide a more comprehensive understanding and assessment of air pollutant emissions. By continually improving and applying satellite remote sensing technology, we are poised to gain better insights into global air pollution issues, reduce pollutant emissions, and promote sustainable development for both human health and the environment.

\subsubsection{Carbon Sequestration estimation}

The process of carbon sink involves the absorption of carbon dioxide from the atmosphere via plant photosynthesis, storing it in vegetation and soil, and thereby mitigating the concentration of greenhouse gases in the atmosphere. Carbon sink calculations require consideration of two scenarios: urban and forest scale.

The main principle of forest carbon sink calculation with remote sensing is based on the reflectance spectral characterization of plants. Different growth and development stages of plants have different reflectance spectral curve patterns and characteristics. Then the forest biomass can be indirectly estimated to monitor the carbon sink. Alternatively, LiDAR can actively acquire three-dimensional coordinate information in the area to estimate forest structure information quantitatively. Li et al.\cite{li2023soil} devised techniques for fusing data from Visible Near-Infrared Reflectance Spectroscopy (VNIR) and Hyperspectral Images (HSI) to enhance predictions of soil carbon content. However, due to the simplicity of the data characteristics, the limited amount of data available, and the high interpretive demands, the majority of applications favor machine learning over deep learning.

Additionally, remote sensing data can be utilized for correlation analysis to acquire the spatial and temporal distribution, as well as to Raine et al.\cite{raine2023image} proposed the use of weakly supervised learning with only image labels for coarse seagrass segmentation, allowing for the computation of seagrass carbon content by deep learning without prior knowledge. analyze land use change's effects on carbon emissions. Huang et al.\cite{huang2022assessment} utilized time series forecasting to predict the forest and rubber plantation distribution, addressing the impact of growing rubber plantation areas in Southeast Asia on carbon storage. Reiersen et al.\cite{reiersen2022reforestree} presented the ReforesTree dataset, which includes aerial images and carbon storage labels for tropical forests in Ecuador. This demonstrates the potential for processing remote sensing data of carbon sinks, extracting characteristic coefficients, and fostering the development of scalable, reliable, low-cost, and accurate artificial intelligence models.

Compared to forest ecosystems, the spatial distribution and functional requirements of urban green spaces are more intricate. It is challenging to accurately monitor changes in urban biomass with just one source of remote sensing information. However, by utilizing multiple data sources, such as remote sensing, road networks, climate data, etc., we can more efficiently detect and forecast the carbon storage capacity at an urban-scale. Traditional methods for estimating carbon storage, including the plot inventory, model inversion, and flux observation methods, have limitations when applied to urban areas. Urban vegetation is influenced by both urban planning and meteorological factors, and its distribution is irregular while it exhibits distinct growth cycles\cite{guo2023novel}. This creates challenges, such as acquiring data and applying predefined metrics, as well as high costs in terms of time and money, which limit their widespread use. Mou et al.\cite{chao2023spatio} developed a neural network learning system for estimating urban-scale carbon storage capacity. Mou's model addresses these challenges by generating remote sensing data to overcome the shortage of high-quality remote sensing data. It integrates temporal and road network information to consider the impact of intricate urban topography on carbon storage capacity. This technique resolves problems associated with costly data collection, ultimately leading to reduced computational expenses and more precise estimations. Yang et al.\cite{yang2020modeling} developed a neural network ensemble model to estimate carbon emissions using satellite imagery captured during nighttime illumination.

Artificial intelligence uses remote sensing data to calculate vegetation indices, timber volumes, and other factors. Based on these calculations, it estimates metrics such as carbon stock, biomass, and carbon storage within a specific region to determine the final carbon sequestration amount. While existing studies can accurately estimate carbon sequestration in a cost-effective manner, it is difficult to establish a clear relationship between variables and output data. As a result, model transferability across various scales and regions remains challenging. Furthermore, remote sensing data updates rapidly, and it possesses characteristics such as spatiotemporal correlations and multi-source data, requiring substantial computational capabilities. The data processing and inferential capabilities of artificial intelligence require further improvement.

\subsubsection{Carbon price forecasting}

In the context of climate change and declining air quality, carbon emissions are highly constrained. Carbon emissions have a cost, which is reflected in the price of carbon in the city's emissions trading system. But predicting the price of carbon is a complex and challenging problem. There is no clear definition of the factors that influence the variability of the carbon price. It is almost impossible to predict accurately over time. Therefore, most of the work has focused on exploring the correlation between carbon prices and various factors. Wang et al.\cite{wang2022novel} combined historical data, influencing factors, and unstructured data such as search and sentiment to provide more complete information and thus reduce uncertainty. Pan et al.\cite{pan2023carbon} improved carbon price forecasting by better incorporating investor attention as a predictive factor. Carbon prices also experience large shocks due to a variety of unpredictable factors. This noisy data can reduce the accuracy of the AI's learning. Zhang et al.\cite{zhang2023ensemble} improved carbon price prediction by handling non-stationarity, adapting to changing relationships between price and factors, and preventing model structure problems such as overfitting. In addition, zhou et al.\cite{zhou2022carbon} proposed a hybrid CEEMDAN-LSTM framework combined with Variational Modal Decomposition (VMD). After optimizing the network structure, the prediction results become more accurate and stable.

Based on precise and objective carbon sequestration data, carbon trading systems are becoming more advanced. By analyzing fluctuations in various carbon sequestration quantities, artificial intelligence can predict changes in carbon prices. However, there are various factors influencing price changes, and their correlations are frequently unknown. Current efforts primarily rely on factors such as energy structure, market regulations, and policy influences to make predictions. Over time, the impact of each factor on prices shifts, resulting in reduced accuracy of the model. Accurately predicting carbon price trends at any given time becomes challenging. Artificial intelligence must address the problem of catastrophic forgetting by constantly learning the newest features of carbon price data to ensure prediction accuracy.

\subsubsection{Future trend}

Future research trends in artificial intelligence in green environments will be diverse and cutting-edge.

\textbf{Deeper merging of artificial intelligence with remote sensing data.} The application of deep learning techniques to remote sensing data processing will continue to deepen, including more complex neural network architectures, migration learning, semi-supervised learning, and so on. Meanwhile, the fusion of multi-source remote sensing data, such as optical, radar, hyperspectral, etc., will become an important research direction to improve the information content and accuracy of remote sensing data. Self-supervised learning and weakly supervised learning methods can reduce the dependence on large amounts of labeled data by automatically generating labels or using auxiliary information to train the model to adapt to the characteristics and diversity of remote sensing data. For the dynamic process of environmental change and resource use, time series analysis will continue to evolve, including finer time series prediction, spatio-temporal modeling, anomaly detection, and other methods to support more accurate dynamic monitoring.

\textbf{Expanding research areas.} In addition to detecting carbon sequestration, monitoring air pollution, and tracking carbon prices, AI is increasingly being applied to various areas of green environmental protection. It analyzes water quality and predicts supply and demand in water resource management. It facilitates waste sorting and recycling through image recognition. AI monitors wildlife activity for endangered species conservation and assesses soil health by analyzing soil data. It also drives sustainable architecture and urban planning, monitors marine pollution, aids in vegetation restoration, manages supply chains, and predicts extreme weather and disasters. At the same time, it generates engaging educational content to raise environmental awareness. These applications underscore AI's diverse contributions to promoting ecological balance, sustainable resource use, and environmental protection.

\textbf{Addressing Interpretability and privacy issues.} As the number of deep learning applications increases, researchers will pay more attention to model interpretability and develop methods that can explain the basis of model decisions. In addition, uncertainty handling will become a research focus, especially for applications with incomplete data and complex scenarios. As AI applications increase, privacy and ethical issues will become more important. Future research will explore how to balance data use and privacy in remote sensing AI applications.

In conclusion, the application of artificial intelligence in green computing showcases tremendous potential. The continuous evolution of this technology will further drive the realization of green technology and sustainable development, creating a cleaner and healthier environment.

\subsection{Green Computing For Engineering}

\subsubsection{Database Security by Green CryptoGraphy}

\textbf{\numberedParagraph{Introduction}}

New development trends such as the Internet of Things, intelligent healthcare, digital enterprise management, and the blockchain are at the cusp. Due to the rapid development of the Internet, the keep growing of data storage pressure has driven the rapid expansion of the entire storage technology. For providing efficient and secure remote storage and management of data, cloud storage systems have been researched carefully and become an indispensable part of the computer research field. In recent years, governments, enterprises, and personal users have been Actively using cloud storage services to gain convenience. Such a large amount of data can drive a lot of wealth. However, cloud storage also brings issues such as illegal access, data breaches, confidential information leaks, and personal information breaches.

Focusing on database and cloud storage scenarios, our project will research and develop green encryption algorithms that can realize data privacy according to database security requirements such as the privacy and integrity of stored data, improve the efficiency of encryption algorithms, and further design supporting data storage proof protocols to ensure intact data preservation and finally realize database security.

The attack on the database mainly refers to the tampering and disclosure of the data stored in the database, namely the destruction of the integrity, availability and confidentiality of the data. Integrity means that the data remains in its original state and is not illegally modified. Availability means that the original legitimate users can still access the data they are authorized to access. Confidentiality means that the data stored in the database is not illegally accessed by unauthorized users. Therefore, this chapter will introduce the cryptographic algorithms and concepts involved in the two core aspects of Database Security by Green CryptoGraphy, the confidentiality and integrity of cloud storage, and present a review on the relevant research results in recent years. In Section \ref{sec:conf}, we focus on the database encryption algorithms that guarantee database confidentiality. And in Section \ref{sec:int}, a review on the research of data integrity is presented.

\textbf{\numberedParagraph{Confidentiality}}
\label{sec:conf}

\medskip\textit{A. Data Encryption} 

With the continuous development of computer technology, database systems are increasingly being applied to various fields. Because the database stores large amounts of data information, some of which involve confidential data, and because the characteristics of resource sharing database, its data and information from a wide range of security has been huge threat. Database security has become the focus of attention, and foreign computer experts have also increased the security of the database but have also published a number of domestic and international protection of computer information security standards. Protection of database security technology, many such as: auditing functions, and database encryption is an important means to ensure database security.

The security of the database system is mainly guaranteed by the security measures provided by the operation system and the database management system themselves. However, there are still some security problems in relational databases. For some important sensitive data, users strongly hope to transmit and store it in an encrypted way. Although other users who do not know the key can access the database, they cannot crack the ciphertext data, so confidential data will not be leaked, which provides a background for studying the encryption of data in databases. Sensitive data is transmitted on a public channel after data encryption. Encrypted data is in an incomprehensible format, so agents decrypting without an authorized key cannot understand the data even if they access it. Encryption algorithms can be classified into {\it symmetric} encryption algorithm and {\it public} encryption algorithm. 
For symmetric encryption algorithm, the keys used for encryption and decryption are the same, and its encryption and decryption speeds are typically very fast. It is more suitable for the occasion of encrypting large amounts of data information.
In contrast, the asymmetric encryption and decryption algorithms usually involve heavy mathematical operations, its encryption and decryption speeds are relatively slow, and it is only suitable for encrypting a small amount of data. Therefore, for database encryption, it is more appropriate to use symmetric encryption algorithm to encrypt the data stored in it.

In this section, we give an overview of the encryption algorithms used in the database. 
More precisely, in Subsection \ref{subsec:enc} we briefly introduce two basic database encryption algorithms, AES algorithm for symmetric encryption and ElGamal algorithm for public-key encryption. And in Subsection \ref{subsec:hyb}, we present a review on the promising hybrid encryption algorithms in recent years.

\medskip\textit{B. Encryption algorithms}
\label{subsec:enc} 

\textbf{AES algorithm} Symmetric key cryptography \cite{alenezi2020symmetric} is also known as single key cryptography. It encrypts and decrypts using the same key, or although the keys used are different but if one key can be easily derived from the other key. The model for symmetric encryption is shown below.

\begin{figure}
  \centering
  \includegraphics[width=10cm]{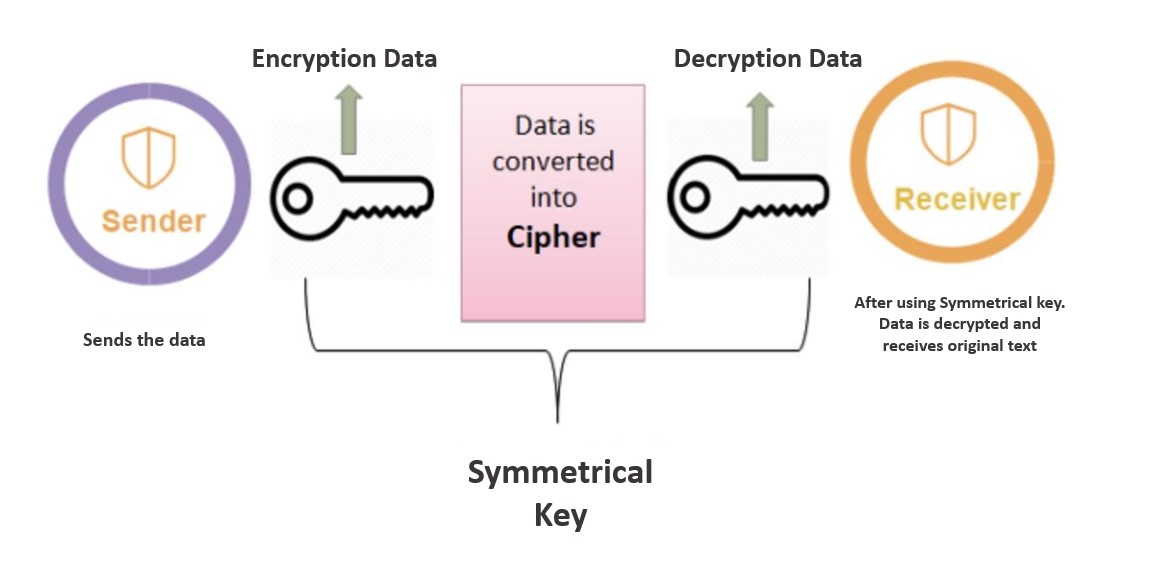}
  \caption{The Model for Symmetric Encryption}
  \label{fig:symm}
\end{figure}

Symmetric encryption algorithms are developed by simple substitution and iteration operations, and have been rapidly developed after the Data Encryption Standard (DES) published in 1977 in the United States. The encryption and decryption speeds of this method are very fast, so it is usually applied to the occasions where mass of data need to be encrypted, and it is widely adopted at present. For example, DES, 3DES and AES are typical symmetric encryption algorithms. We will introduce the AES algorithm in detail.

AES algorithm is an encryption algorithm picked from 15 candidate algorithms by the NIST in October 2000, and is used as a new key encryption standard. Rijndael was chosen to be the future AES and was founded in 1999 by Joan Daemen and Vincent Rijmen \cite{daemen1999aes}. Then, NIST developed the new Advanced Encryption Standard specification in 2002 \cite{akkar2001implementation}. 
AES algorithm is an algorithm designed to replace DES, which achieved a higher security level, based on permutation and substitution. 

\begin{figure}
  \centering
  \includegraphics[width=7cm]{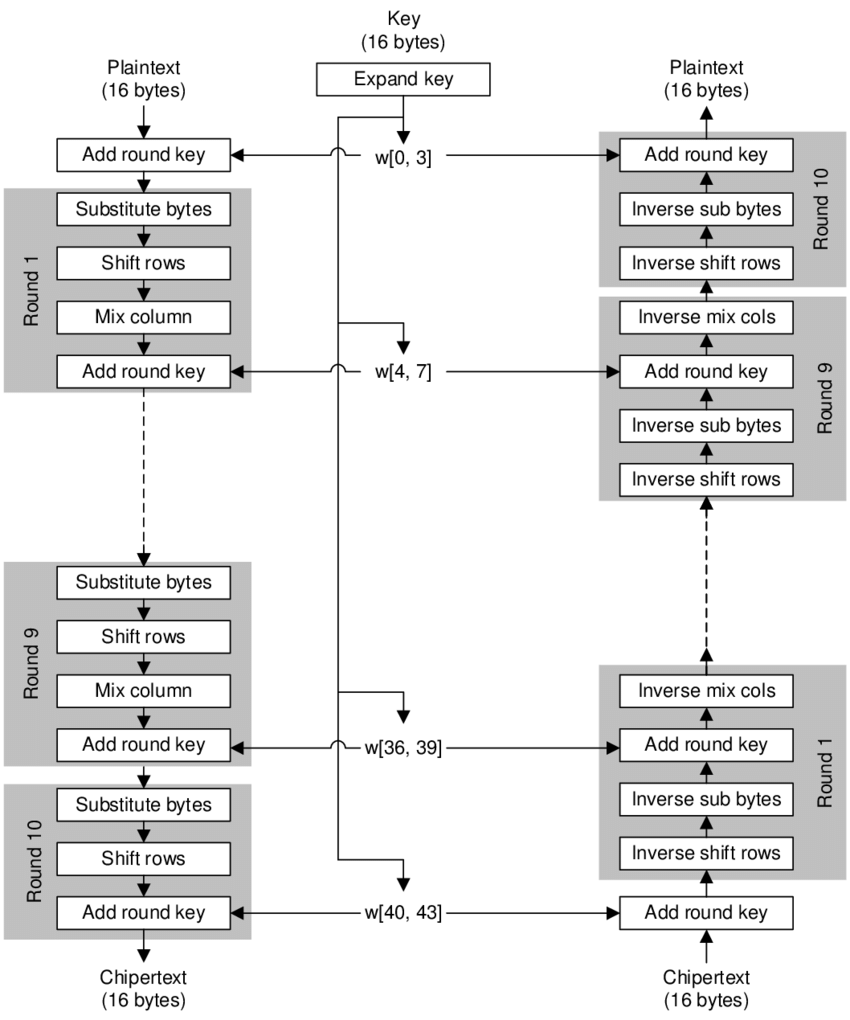}
  \caption{AES Encryption and Decryption Process Block Diagram}
  \label{fig:aes}
\end{figure}

AES algorithm is an iterative symmetric block encryption algorithm, the key length can be 128, 192, 256 bits. Iteration is a cyclic process in which data is repeatedly arranged and replaced. It has the advantage of simple design, suitable for various platforms, high operation speed, longer key length, stronger resistance to attack. The AES 128 bit key is much more safer than the DES 56 bit key. In the AES algorithm, the plaintext is changed into ciphertext output after S-box transformation, row transformation, column transformation, and XOR operation with the key for many times. The decryption process is similar, and the specific encryption process is shown in Fig. \ref{fig:aes}.

\textbf{ElGamal algorithm} In 1976, Diffie and Hellman proposed a key exchange protocol in Cryptography, which allows data to be transmitted confidentially and securely between two parties in an insecure communication channel. On the basis of this new idea, the public key encryption system appears. In the public key encryption system, the encryption key can be disclosed,through the communication channel, allowing it to be known and used by a third party. However, the key used in decryption is known only by the decrypter himself, which is confidential and cannot be disclosed. The key used for encryption is called the public key, and private key is used to decrypt the cipher text. Public and private keys can only be used in pairs, and plaintext encrypted with a person's public key can only be decrypted with its paired private key. 

\begin{figure}
  \centering
  \includegraphics[width=10cm]{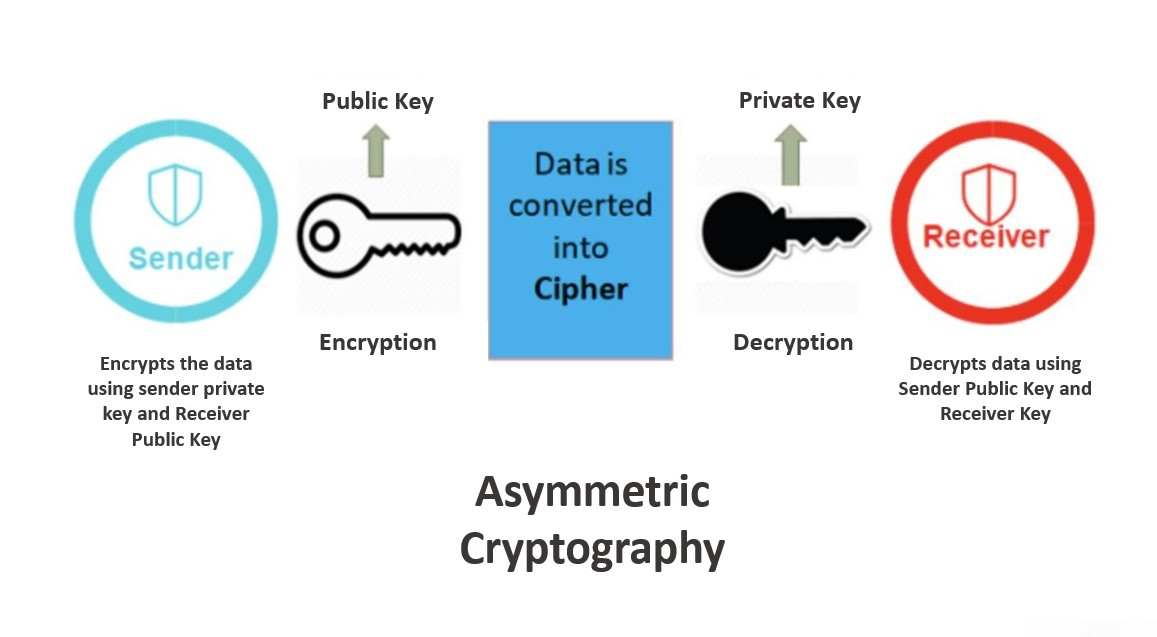}
  \caption{The Model for Public-Key Encryption}
  \label{fig:public key}
\end{figure}

The speed of encryption and decryption of asymmetric cryptography is very slow, even can only reach one thousandth of the symmetric encryption algorithm, but it has stronger security performance. In particular, we will introduce the ElGamal algorithm in detail.

The ElGamal algorithm, proposed by Tather ElGamal in 1985 \cite{elgamal1985public}, is an public cryptographic system. The security of ELGamal algorithm is based on the difficulty of discrete logarithm problem, whose idea can be applied in digital signatures. Compared with RSA algorithm, Encryption of ElGamal algorithm is not unique, which means that different ciphertext can be encrypted from the same plaintext using the same private key. The ELGamal algorithm can effectively preventing possible replay attacks in the network. 

\smallskip
\textbf{Definition(ElGamal Public-Key Encryption Scheme)}
The ElGamal asymmetric encryption algorithm can be described as follows:

\begin{itemize}
  \item \textbf{Key Generation algorithm $\mathbf{Gen}$}: $(PK,SK) \gets Gen(1^{\lambda})$
  \begin{enumerate}
    \item Select a cyclic group $G$ with order $p$, and generator $g$.
    \item Choose $s \gets \mathbb{Z}_p$ uniformly, and compute $h:=g^s$.
    \item Output $PK=(G, p, g, h), SK=s$
  \end{enumerate}

  \item \textbf{The encryption algorithm $\mathbf{Enc}$}: $C \gets Enc(PK, M):$ The message space is $\mathcal{M}=G$
  \begin{enumerate}
    \item Choose $r \gets \mathbb{Z}_p$ uniformly.
    \item Compute $C_1:=g^r$.
    \item Compute $C_2:=h^r \cdot M$.
    \item Output $C:=(C_1, C_2)$.
  \end{enumerate}

  \item \textbf{The decryption algorithm $\mathbf{Dec}$}: 
  $M^{'} \gets Dec(SK,C=(C_1,C_2))$:
  \begin{enumerate}
    \item Compute and output $M^{'}:=C_2 / C_1^S$.
  \end{enumerate}
\end{itemize}

In practical applications, such as in database system, the ElGamal encryption is often used in hybrid cryptosystems. For example, symmetric encryption is used to encrypt the message, and then ElGamal encryption algorithm is used to pass the key. This is because, as an asymmetric cryptographic system, ElGamal is generally slower than symmetric cryptographic schemes at the same level of security. The key of a symmetric the encryption algorithm is usually much shorter than the message to be delivered, so it is faster to use an ElGamal encryption key and then use symmetric encryption to encrypt message of any length.

\medskip\textit{C. Hybrid Encryption Algorithms Review}
\label{subsec:hyb}

As we introduced before, cryptography is very important in database system (Cloud Computing). Because public key encryption takes more time, and the security of symmetric encryption is worse, so in order to adapt to the cloud computing environment with high requirements for efficiency and security, the database encryption in recent years mostly uses hybrid encryption algorithm.

As shown by \cite{kuswaha2015data}, they firstly propose a hybrid encryption scheme to compensate for the shortcomings of common symmetric encryption and public key encryption algorithms. Their algorithm is an integration of AES and RSA encryption schemes. This new hybrid encryption technology is designed to improve security and open ciphertext integrity. Compared with the conventional AES algorithm, the hybrid model performs better in nonlinear performance and has better diffusion characteristics after combining with RSA, which greatly increases the difficulty of algebraic attack on their encryption model. The hybrid encryption algorithm provided by \cite{chauhan2016implemented} uses the ECC scheme and MD5 hash generation to to have better performance in efficiency and security. In this article, the ECC algorithm's private key generation is based on a self-generated key, and this key is combined with a cypher in secure file exchange. Beyond that, the algorithm can extract the key from the given ciphertext and has the ability to cross-check the validity of the ciphertext. Bhole et al. \cite{alkady2013new} have also proposed a security protocol that uses mixed encryption. The plaintext is divided into small blocks in their algorithm, and the key is protected by ECC algorithm, which has quite high security in the public key algorithm.

Soman and Natrajan \cite{soman2017enhanced} propose an optimized cloud hybrid data security scheme to ensure the protection of data security and integrity in the cloud. Their solution uses an encryption algorithm derived form AES, SHA256 and to transfer data in the cloud. For message file uploading, cloud users are required to ensure the security of data messages or files before sending them to the server. At first, the user generates the ECDSA with the digest of data generated by SHA256 and the signature locally, then uses AES to encrypt the data containing the public key. Then users send the encrypted file to the cloud service provider and stores it to the server. s for the download of files, the users of the cloud firstly send a request to inform the cloud server that they want to download the stored files or data packets. The cloud service provider then check the hash value of the request message. Only when the hash value matches, the cloud server will use the private AES key of the requesting cloud user to decrypt the file. Parallel encryption scheme proposed in \cite{chauhan2017novel}, which mixes and transforms the MD5 and Blowfish encryption schemes, creating a hybrid MD5-Blowfish cipher computation that improves security while improving security. The disadvantages of symmetric block cipher and hash generation schemes are overcome.

Mehul \cite{batra2018secure} proposed an algorithm consisting of four encryption schemes including AES, DES, RC4 and Steganography. Their goal of hybrid encryption is to make cloud storage systems robust and ensure data privacy by mixing different encryption algorithms. In the encryption process, data is divided into three parts: the first part uses RC4 encryption, the second part uses DES encryption, and the third part uses AES encryption. Finally, Steganography will be used to hide the key. Steganography is a technique of information hiding so that no one other than the intended recipient knows the events of the transmission of information (not just the content of the information). In \cite{biswas2019efficient}, the authors also propose algorithms using AES and RSA. To increase the level of security, the symmetric key used for message encryption is also used as the RSA public key. a hash value generated on the message will be encrypted again using the RSA algorithm to generate a digital signature. the digital signature of the ciphertext will help the receiver to verify the integrity of the data. The authors claim that the resistance of their proposed scheme to attack has been guaranteed.

Combining cloud storage technology with user local computing power to reduce costs and increase efficiency is an inevitable trend of future development. The cloud technology brings lot of benefits, which makes it inevitable that it will occupy a place in the IT field. Cryptography plays a important role in addressing security difficulties in the cloud environment. Hybrid cryptography schemes have created a variety of new research directions for new researchers trying to break through the various limitations of traditional encryption algorithms.

\textbf{\numberedParagraph{Integrity}}
\label{sec:int}

\medskip\textit{A. Introduction of Integrity Auditing}

By using cloud storage and cloud computing services, cloud users can easily access their private data anytime and anywhere without using local storage space, which is especially suitable for clients with restricted hardware condition to cache and manipulate data. By addressing users' local storage limitations, cloud storage is receiving attention\cite{yan2017context}. However, as the amount of file stored in the server increases, the security of data storage becomes particularly important \cite{yu2017survey}. Users lose control of their data when they store it in the Cloud, and the Cloud Service Provider (CSP) is not fully trusted. In order to reduce costs and save storage resources, the CSP may intentionally delete some data that is not frequently accessed by users. In addition, since the CSP provides storage services to multiple users at the same time, storage resources are relatively concentrated. When the cloud storage system suffers from unrecoverable events such as hardware and software failures or malicious attacks, the data stored in the cloud may be damaged, but users cannot detect the infrequently accessed data in a timely manner. Therefore, users are required to perform data integrity auditing on the file in the server storage to ensure data integrity and availability \cite{yu2019verifiable}.

The cloud data integrity audit was first proposed in 2003 in the article "Remote Integrity Check \cite{deswarte2004remote}". Subsequently, in 2007, a data integrity verification scheme called Provable Data Possession (PDP) \cite{ateniese2007provable} was introduced, which also initiated the research direction of PDP. At the same time, another integrity audit scheme called Proof of Retrievability (POR) \cite{juels2007pors} was introduced in 2007. Different from PDP, POR put more focus on the recovery process when data is verified to be falsified. The initial PDP and POR studies were focused on data integrity audits in single-cloud environments where the data of user is only stored in one cloud server. With the continuous research of cloud storage technology and the emergence of distributed cloud architecture, PDP and POR for the new distributed cloud environment appear variants on the basis of the two, and the research of integrity audit scheme in multi-cloud environment has also attracted wide attention.

In the following section, we present an overview of the integrity auditing. We review both the PDP and the POR researches from the beginning architecture to multiple clouds. In addition to introducing the basic models, we also investigated the improvement directions of PDP and POR integrity auditing schemes in recent years, such as public auditing and dynamic data operation. 
The rest of this section is organized as follows. 
In Subsection \ref{subsec:pre}, we present preliminaries of the integrity auditing and related concepts. And in Subsection \ref{subsec:audit}, we review and compare various PDP and POR schemes in detail 

\medskip\textit{B. Preliminaries}
\label{subsec:pre}

\textbf{Basic models} In the basic single-cloud storage model, users can only store one transcript in the cloud server, and the local data is deleted. Retrieve data from a single cloud server when the user needs it. However, when users use cloud storage in this way, they also lose complete control over their personal data. Cloud storage servers may modify or delete data stored by users to save storage space. Therefore, users must have the capability to verify the integrity of file copies they gave to the server, which can be defined as the integrity auditing. The audit can be conducted by the user himself, but when the user is unable to complete the audit on account of hardware limitation, the audit should be able to be proceeded by a third-party auditor with the permission of the user.

\begin{figure}
  \centering
  \includegraphics[width=7cm]{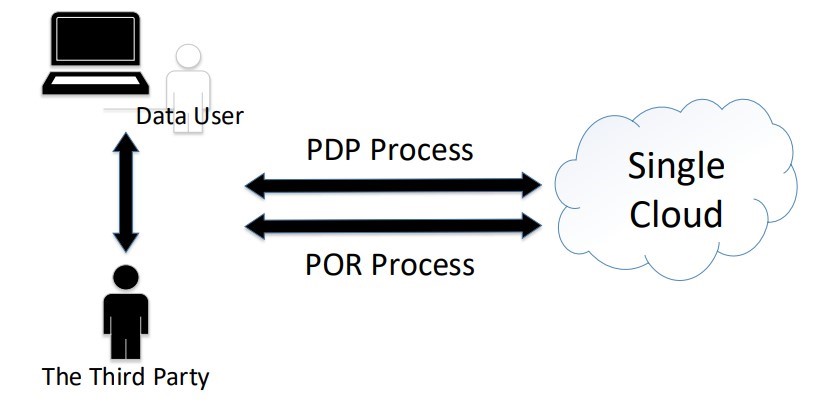}
  \caption{Single Copy Single Cloud Auditing Model.}
  \label{fig:1}
\end{figure}

The current data integrity audit research is mainly divided into two directions: Provable data possession (PDP) and proof of retrievability (POR). In the beginning cloud architecture, clients deliver the file transcript to the cloud, then remove local file possibly on account of hardware limitations, and then perform all operations of the file in the cloud. To guarantee the data integrity, the intuitive scheme is for users to download all the stored data from the server, check the data integrity and upload it again. However, this approach is not suitable for massive data environments that manage massive amounts of user data, often requiring significant communication overhead due to the interaction between users and the server. The solution based on this consideration that the user checks the integrity of the data without the retrieving it. The PDP scheme is shown in the figure. The user challenges the cloud service provider, and the cloud sends the integrity certificate to be verified by the user. The PDP scheme can evaluate the integrity of file, but cannot recover the corrupted data. However, the Proof of Recoverability (POR) scheme ensures that compromised data can be recovered. We summarized the audit flow pass in \ref{fig:1}.

\begin{figure}
  \centering
  \includegraphics[width=7cm]{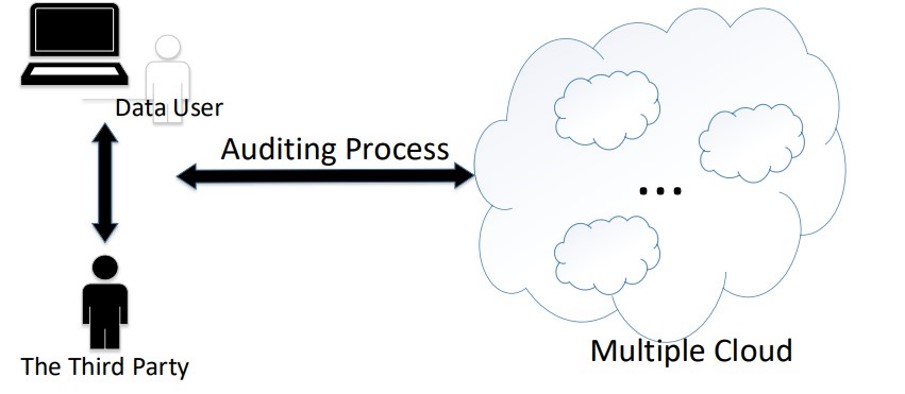}
  \caption{Multiple Cloud Model.}
  \label{fig:2}
\end{figure}

However, even POR schemes can only recover data when the loss of data is not serious. For instance, when a cloud service provider encounters some irreversible problem, such as devices destruction, the file can never be recovered, even under POR. The current distributed multi-cloud server architecture provides a solution to this limitation, in a distributed server cluster, users can store multiple transcripts of a file on multiple distributed devices. In this case, even if a transcript of the file on some servers are corrupted, the user can recover the file through the other transcripts. The presence of multiple servers also prevents some servers from storing fake files. In order to store and operate files, users may need to communicate with multiple servers, which places demands on the user's resources, such as the need to increase storage and computing power. Similarly, when users run low on computing storage resources, a trusted agency can be utilized to reduce the computing overhead of the user. The distributed model is shown in the figure. \ref{fig:2}

\textbf{Related terms}

\begin{itemize}
  \item [(1)]
  Integrity verification of cloud storage: A mechanism by which the cloud storage server can prove to the verifier (user or TPA) that the data it stores is intact. It proves that the cloud storage server is actually storing the user's data.
  \item[(2)]
  Random sampling detection: the verifier does not need to download all the data actually stored from the cloud at the time of verification, but randomly extracts pre-processed metadata blocks to generate data holding probability. Most integrity auditing schemes are based on random sampling verification.
  \item[(3)]
  Random Oracle Model: The random oracle model and the standard model are two models used in the provable theory of cryptography. The standard model only uses the real hash function to complete the proof \cite{erway2015dynamic}. The random predictor model assumes that the hash function is absolutely safe, and other assumptions can be added in the proof to reduce the difficulty of proof. At present, most of the proofs of integrity verification schemes use random oracle model \cite{jiang2018id,fan2019one, tian2018attribute}.
  \item[(4)]
  Zero-knowledge proof: The prover proves the correctness of the statement to the verifier without revealing any useful information. In order to protect the privacy of data in public validation, the validation process is based on the zero-knowledge attribute \cite{zhu2011dynamic}, that is, the server does not have to provide stored data to the verifier to complete the validation.
  \item[(5)]
  MAC( Message Authentication Codes): Using Hash function to encrypt message digest to generate message authentication code, the security of which depends on Hash function, is a method to realize data integrity verification. 
  \item[(6)]
  Aggregate signature: Aggregate signature aggregates $p$ signatures of $m$ different messages from $n$ different users into a short signature $q$. By verifying $q$, you can prove the correctness of m messages from $n$ users. In batch authentication, this type of signature is used to complete the data integrity verification of multi-user and multi-replica.
  \item[(7)]
  Jump table: It is a data structure commonly used in dynamic verification, which is a special ordered linked list structure, in the dynamic operation to add multiple indexes, can quickly achieve insertion and deletion operations. The time complexity of these operations is $O(\log n)$ while The complexity of ordinary single linked list is $O(n)$. However, the jump table requires a large storage space. It is a space-for-time list structure.
  \item[(8)]
  MHT (Merkle Hash Tree): It is a binary tree structure for Hash values storage and is also one of the widely used authentication structures at present \cite{wang2010enabling}. The leaf node of the MHT stores the Hash value of the data block. In the integrity verification process, the value of the root node can be verified to determine whether the data is completely saved. In addition, a branch can be partially verified.
  \item[(9)]
  The Computational Diffie-Hellman problem: A difficult problem in mathematics. Given $g^a$ and $g^b$, calculating $g^{ab}$ is difficult.
\end{itemize}

\textbf{Evaluation criteria} With the deepening of the research on integrity auditing, the efficiency and flexibility of the PDP and POR schemes are constantly improving. According to the survey on recent articles, the evaluation criteria of an integrity auditing scheme is as follows:

\begin{figure}
  \centering
  \includegraphics[width=10cm]{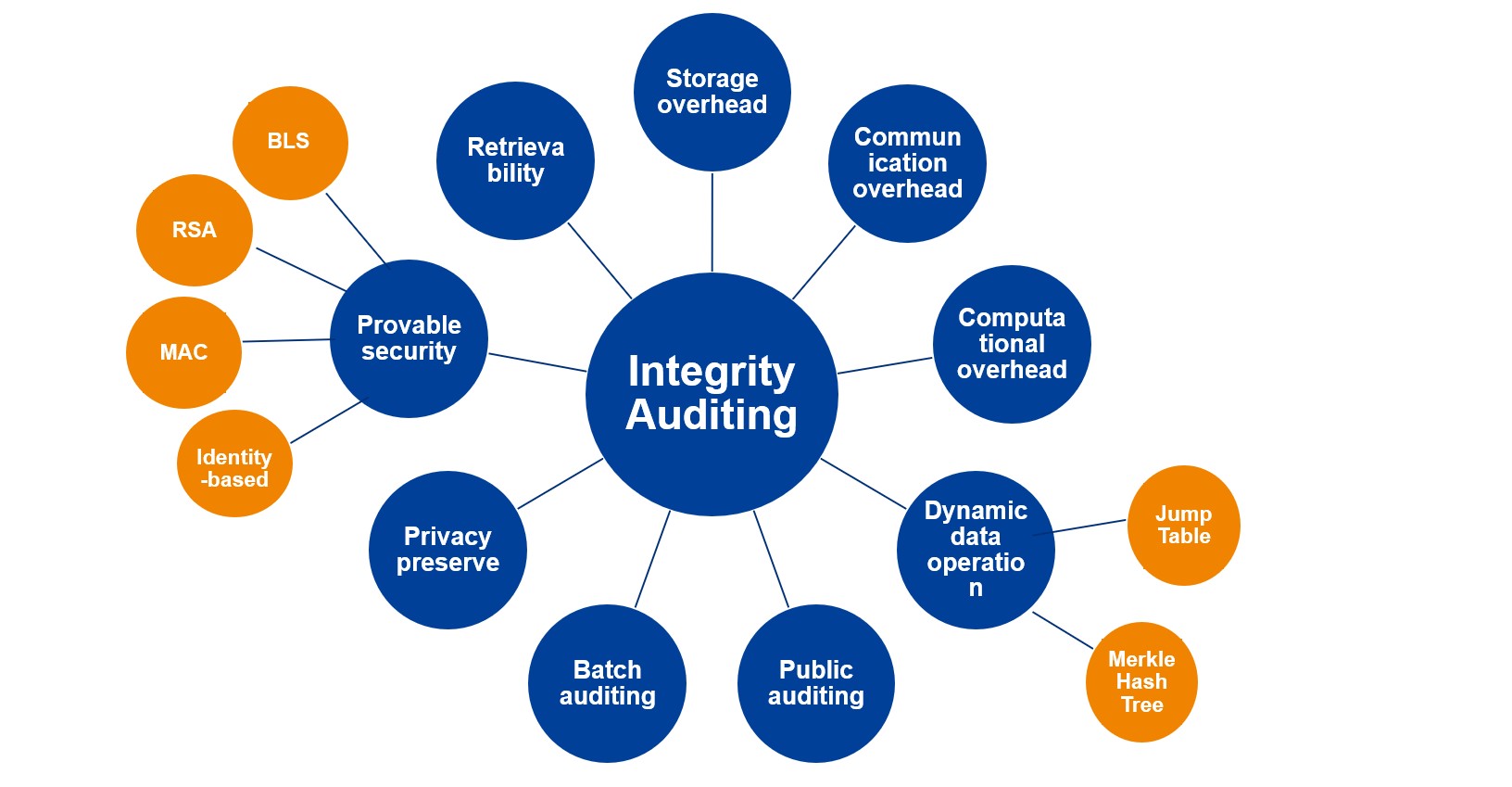}
  \caption{Evaluation Criteria for Integrity Auditing}
  \label{fig:3}
\end{figure}

\begin{itemize}
  \item \textit{Storage overhead} evaluates the storage occupation of the server and the user. Because the main reason for users to deliver the personal data to the cloud is the limitation of local storage, the audit solution should not bring too high storage costs to users.
  \item \textit{Communication overhead} represents the cost that occurs during the process of integrity auditing when a user needs to interact with the server.
  \item \textit{Computational overhead} Occurs when the cloud server generates proofs as well as user verifies proofs, during which the cloud generates labels based on the data stored by the user and sends them to the user, who verifies the proofs. The scalability of the scheme is largely based on the overall computational complexity.
  \item \textit{Dynamic data operation} refer to the operations that users can perform to store data in the cloud, such as adding or deleting data. An integrity audit solution that supports dynamic data operations enables more flexible cloud data management.
  \item \textit{Public auditing} that after users store data in the cloud, they do not directly participate in the audit verification process, but hand it over to a trusted third party. A solution that supports public auditing can greatly reduce the storage and computing overhead for users.
  \item \textit{Batching auditing} is to audit the data integrity of multiple users simultaneously. Supporting batch auditing can also effectively improve the efficiency of the verification scheme.
  \item \textit{Privacy protection} describes the capability at protecting the privacy of users during the audit process. In order not to disclose any data information about the user to anyone other than the user, including third parties to help audit, privacy is also a necessary condition for users to be willing to deliver their data to the cloud \cite{wenxiu2017privacy}.
  \item \textit{Provable security} is a measure of the definition of security that an audit scheme can achieve, judged by the model in cryptography, that is, whether the scheme is secure under the standard model or the random Oracle model.
  \item \textit{Retrievability} means whether the audit scheme can help the user recover the initial file even when the data is missing. 
\end{itemize}

\medskip\textit{C. Auditing Schemes Review}
\label{subsec:audit}

In this subsection, we will delivered a comprehensively overview of previous researches about data integrity auditing and compare reviewed schemes. The auditing schemes can be classified into three types: PDP schemes, POR schemes and multi-cloud schemes.
 
\textbf{Auditing Schemes Based on PDP} Prior to the formal definition of PDP in 2007, Deswarte et al. researched the integrity checking of remote cached data in \cite{chang2008remote} firstly. They used hash functions which is similar to the RSA algorithm to compute a tag of the file and generate challenges by calculating the checksum of the file. Their scheme uses public key algorithms and Diffie-Hellman key exchange based authentication protocol. However, this scheme is computationally expensive and very difficult to scale.

Then, Ateniese et al. proposed the first formal definition of the PDP scheme \cite{ateniese2007provable}. two PDP schemes designed in this paper are based on homomorphic verifiable tags. The user first divides the data into small blocks and computes a tag for each data block, then delivers these tags with the file to the cloud. In the process of verification, the challenger selects some blocks at random to challenge the verifier and asks the verifier to return relevant proof generated from the request blocks and corresponding tags. Due to the homomorphism property of the tags, labels of different file blocks can be represented as a simple value, which greatly reduces communication overhead and enables users to confirm data integrity without accessing the complete file. Their scheme supported third-party verifiability. The user is able to conducted integrity verification for any times. However, the modular operations used in RSA algorithms forced their scheme to support no dynamic operations.

According to the encryption mode of authentication data, PDP schemes can be divided into RSA, BLS, MAC and identity attribute based encryption. In addition to the RSA-based PDP schemes \cite{ateniese2007provable}, Shah et al. \cite{shah2008privacy} proposed a MAC-PDP scheme, which uses MAC as authentication metadata and randomly extracts some metadata blocks to complete remote data integrity verification. Although the computing and communication costs are reduced, this mechanism requires users to store a lot of authentication information. Then a verification scheme based on BLS signature was proposed in literature \cite{li2013multiple}. The key length of BLS short signature is $160$ bit, which is much smaller than the calculation amount of $1024$ bit key for RSA signature. In 2016, Yu et al. \cite{wang2016identity} proposes a authentication scheme based on identity attributes, that is, a authentication scheme that generates private keys combined with the user's identity attributes in the process of key generation. In recent years, most PDP schemes \cite{yu2016identity,tian2015dynamic} are based on these encryption methods, and have been optimized in terms of computation, communication, storage overhead.

In order to better fit the current application scenarios such as car networking, smart medical and smart city, the PDP scheme should support the verification of dynamic updates such as data insertion, modification and deletion. According to the survey, PDP schemes supporting dynamic operations can be divided into based on jump table based, Merkle Hash Tree, sequential index table and other linked list. Dynamic PDP research has also been done by Erway et al. \cite{erway2015dynamic}. One of their scheme used the jump table and the other was designed using the Tree structure. However, since the jump table is a space-for-time data structure, the storage overhead of this scheme is too high once large files are stored. Therefore, \cite{wang2010enabling} proposes another dynamic verification scheme based on Merkle hash tree, which stores the encrypted files in blocks on the leaf nodes of the Merkle hash tree, and realizes the full dynamic operation of the verification scheme at the data block level through the insertion, deletion and modification of the leaf nodes of the tree. In the scheme of Yao et al. \cite{yao2016efficient}, they used The Large branching 
 tree to design a dynamic PDP scheme. proposed a new dynamic PDP scheme by proposing a secure signature scheme and using the Large Branching Tree. By using the LBT instead of Merkle Hash Tree, they achieved better efficiency with less communication overhead. In 2019, Li et al. \cite{li2019method} changed the traditional dynamic PDP structure, designed on the bidirectional link information tables and position arrays. 

Shah et al. \cite{shah2007auditing} put forward the concept of third-party auditing while defining PDP, by differentiating external auditing and internal auditing. Shah et al. \cite{shah2008privacy} allows external agencies to regularly validate file and proactively communicate with users, thereby reducing the authentication burden on users without compromising user data and supporting the privacy protection of user data. Similarly, Wang et al. \cite{wang2010privacy} 's scheme also allows external agencies to verify the integrity of data on behalf of users, while enabling users to dynamically modify the data, with great flexibility. The solution improves on the previous cloud storage model using classic Merkle hash trees and bilinear aggregated signature techniques to perform batch auditing for multiple user Settings. After their work, Wang et al., \cite{wang2013panda} propose a new audit architecture (PANDA) that considers valid user revocation and the integrity of users personal data. The idea of re-signing with a proxy allows the cloud to re-sign blocks on behalf of existing users during user revocation so that existing users do not have to download and resign the block.

In general, the development of the PDP-based cloud storage data integrity verification mechanism mainly focuses on the following points: (i) Associated application scenarios: More effective verification strategies are proposed for different scenarios and different requirements; (ii) Strengthen security capabilities: As the openness of cloud storage increases, the security of stored data will also be more threatened, and the protection of data privacy becomes particularly important; (iii) Optimization of verification efficiency: Actual production puts forward higher requirements for the real-time efficiency of cloud services.

\textbf{Auditing Schemes Based on POR} Unlike PDP, POR were designed to focuses its research on the recovery of tampered files. Juesl et al. \cite{juels2007pors} first formally designed POR. their intuitive measure was to encrypt the file encoded before, and the users need to return the "sentinel" at some randomly chosen blocks of the file at the time of verification. Depending on the property of the erasure code used, the original file can be recovered from the deleted encoded file. However, according to their scheme, every challenge is generated by a different sentinel, thus users cloud only made finite times of challenges. Later, \cite{shacham2013compact} was proposed in 2013 for infinite times of challenges which is a limitation in \cite{juels2007pors}. A private auditing schemes using pseudorandom functions and a public auditing schemes based on the BLS signature was proposed in \cite{shacham2013compact}. In the meantime, both schemes used homomorphic authenticators to simplify the proofs by combining clocks and authenticators into a short tag due to the homomorphism property. Since then, research on POR schemes has become quite diverse using different Cryptography and computer technology. Different POR schemes can also be designed based on different application scenarios.

According to the way how was the data stored in the server audited, the POR schemes can also be divided into types: (i) private auditing schemes such that the auditing can 
 only be conducted by the users; (ii) public auditing schemes such that the verification process can be helped by a third party agency. Third-party audit can help users save computing resources, but it needs to consider the problem of privacy information disclosure. Therefore, according to the privacy requirements of different application scenarios, different POR schemes also selectively support public audit or just private audit. \cite{shi2013practical, armknecht2014outsourced,shin2017secure,omote2015md} are the POR schemes supporting third-party auditing whereas \cite{juels2007pors, jianchao2015proof,cash2017dynamic,vasilopoulos2016message} are private auditing researches. Only a few POR schemes is compatible with both private and public auditing \cite{shacham2013compact,huang2014enabling, zhang2014efficient}.

Similarly to the PDP schemes, the beginning POR schemes such as \cite{juels2007pors} and \cite{shacham2013compact,yuan2013secure, armknecht2014outsourced,thao2014sw,chauhan2014robust, omote2014new} are all schemes that is not compatible with dynamic operations. However, recent POR researches which supports for dynamic data operations include \cite{miller2014permacoin, huang2014enabling, shi2013practical, li2013efficient}. According to the survey on recent POR schemes, naturally, POR schemes are being built to support dynamic data operations, because dynamic POR is not only suitable for dynamic operations, but also capable to deal with static data that does not need to be updated. On the other hand, compared with the single server setting, the distributed server setting is more prominent because of the data corruption elasticity and backup, so the research of multi-server POR scheme is also an inevitable trend. In terms of data recovery, although erasure coding is still the dominant trend, in the future, network coding may be another wise direction because it is more resource and computational efficient than erasure coding in the data recovery process.

\textbf{Auditing Scheme for Multiple Replicas} As illustrated before, POR and PDP schemes for single cloud environment cannot deal with the badly damaged file. Thus, Multi-cloud integrity auditing can be an effective solution and have become very popular recent years. In this subsection, we will deliver an overview of the auditing schemes in distributed sever architecture. The beginning auditing scheme designed for multi-cloud models was proposed by Deswarte et al. \cite{deswarte2004remote}.However, their traditional protocol is inefficient and can be attacked by malicious users. And in their protocol, the verifier has to take some storage resource to cache tables of checksums.

Ateniese et al. \cite{ateniese2011remote} also considered a multi-copy storage scheme, but unfortunately this scheme proved to be insecure and the user's privacy would be compromised by multiple servers colluding. Next, Curtmola et al. \cite{curtmola2008mr} proposed a multi-copy PDP scheme that uses encryption tools and protects the privacy of user data. They demonstrate that storing multiple copies has less communication overhead during validation than than the single-cloud situation but may cause expensive storage and communication cost for the users. Based on \cite{curtmola2008mr}, a multi-part audit protocol supporting public verifiability is proposed by Hao et al. \cite{hao2010multiple} using BLS-based homomorphic authentication tags. This protocol has higher security and efficiency.

Because most schemes are designed based on public key cryptography, it is necessary to involve the distribution and management of the password, which will bring a large overhead. To solve this problem, Wang \cite{wang2014identity} proposes an identity-based data ownership scheme for distributed settings(ID-DPDP) that eliminates the need to manage passwords. For the dynamic operation of data, Long et al. \cite{long2019dynamic} proposed a multi-copy PDP scheme of cloud storage data based on AVL tree in their article. The model is designed on the Merkle hash tree scheme to provide security and reduce communication and storage overhead.

In multi-cloud architecture, When an error happens in a file, it is necessary to clarify where the error occurs. In recent years, Rakesh et al. \cite{rakesh2017distributed} designed an dynamic scheme with high efficiency that used homomorphic tokens to identify which server has the behavioral fault. 

\subsubsection{Complex Program Behavior Analysis Method for
Software Engineering Efficiency Improvement}

\textbf{\numberedParagraph{Introduction}}

As the complexity of business continues to grow, enterprise systems have evolved in intricacy, and multi-application systems like microservices have been increasingly used. Subsequently, problems that were not paid attention to gradually become prominent, such as implicit calls between applications, the influence of frameworks/configurations on calls, etc. The traditional techniques of static analysis and symbolic execution help understand and analyze the behavior of applications or systems; however, it is difficult for them to tackle emerging issues like scalability, especially when comprehending and analyzing the program behavior of complex software.

To shed light on software behavior comprehension and analysis, this part will first survey the methods of program behavior extraction that abstract complex behavior into a structure of entity and dependency. Then it will illustrate a representative application namely software refactoring recommendation based on code dependencies.  The two aspects will give a roadmap of how techniques have evolved to address complex software behavior analysis, thus facilitating effective and efficient software engineering tasks for green computation purposes.

In general, this part is organized into three sections:

\begin{itemize}
    \item 
    The first section introduces the fundamental concepts of code behavior, namely code entities and dependencies.

    \item The second section introduces the entity and dependency extraction that analyzes code behavior. This encompasses an exploration of traditional static analysis methods, heuristic-based analysis techniques, and the latest Artificial Intelligence (AI)-driven methods. Challenges and potential directions of AI-enabled code behavior analysis will also be discussed.

    \item The third section explores refactoring recommendations, including traditional methods, search-based methods, and the recent machine learning-based techniques. Prospective opportunities and challenges brought by AI models will be discussed.
\end{itemize}

\textbf{\numberedParagraph{Code Behavior: Entities and Dependencies}}

\emph{Entity} and \emph{dependency} are the essential concepts for code behavior. 
The analysis of source code entity and dependency is critical for modern architecture analysis, including architectural metrics \cite{maccormack2006exploring}\cite{mo2016decoupling},  anti-pattern observation~\cite{macia2012automatically}\cite{macia2012supporting}\cite{mo2015hotspot}, change impact analysis~\cite{li2013survey}, fault forecasting~\cite{zimmermann2008predicting}\cite{nagappan2006mining}, etc.  

\begin{figure}[htbp]
\centerline{\includegraphics[width=0.46\textwidth]{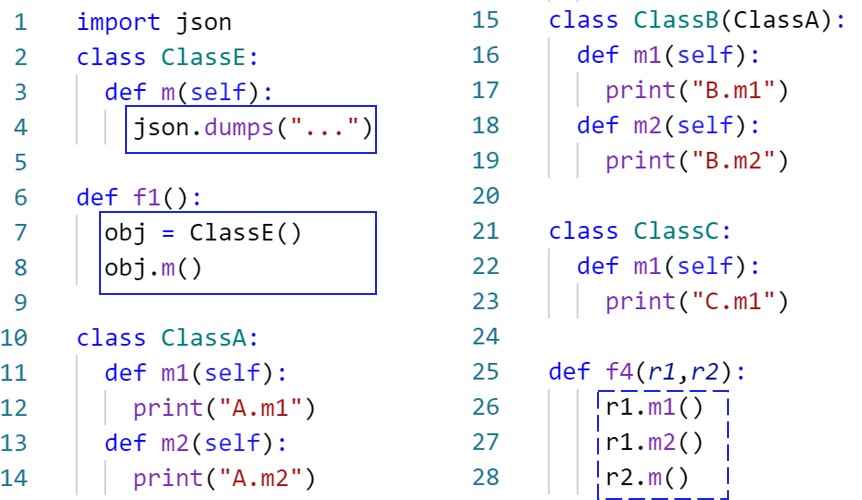}}
\caption{A python code snippet }
\label{fig:case}
\end{figure}

Dependency can be manifested in a variety of ways. If a dependency is extracted from the source code, then it becomes a \emph{syntactic dependency}. If it comes from the textual information of source code, it can be called \emph{semantic dependency}. If it is recorded in the revision history, it becomes a \emph{historical dependency}.
\cite{bavota2013empirical}\cite{gall1998detection}\cite{poshyvanyk2009using}\cite{arisholm2004dynamic}. 

Jin et al. \cite{jin2022evaluating} focused on syntactic dependencies. Normally, syntactic dependencies come from the output of static analysis procedures performed on the source code or its production. Dynamic programming characteristics would make syntactic dependencies impossible to be obtained directly from source code. These are referred to as \textit{possible dependencies}. They categorize the dependencies in dynamic programming languages into two different types: explicit dependency and possible dependency.

\textbf{\numberedParagraph{Entity and Dependency Extraction Methods}}
\label{sec:method}

With the development of static code analysis technology, a variety of source code dependency extraction methods have emerged in academia and industry. Traditionally, entity and dependency can be extracted by traversing AST. However, extracting source code entity and dependency can be difficult due to the existence of possible dependency. To address possible dependency, many different techniques have been implemented, ranging from heuristic type inference extraction to the use of deep learning. Now with the development of a large language model (LLM), traditional static code analysis methods have the opportunity to be enhanced with the LLM process.

\medskip\textit{A. Traditional Methods}

A programming language compiler usually consists of two parts: the front-end of the compiler is responsible for parsing the target program into a specific source language, and the back-end of the compiler generates code for the target machine. Between these two processes, an Abstract Syntax Tree (AST) is generated as a representation of the program at the level of abstract syntax. The significant advantage of using AST is the higher level of abstraction compared to source code. Therefore, the algorithm only needs to be developed once and can be used in programs written in a variety of different programming languages. \cite{4299919}. So, we can extract entity and relation dependencies from source code through visiting nodes from the AST, without considering specific programming language. 

In general, an AST-based source code dependency extraction method involves with several tasks:

\textbf{Parsing} This task aims to generate an AST model from the source code. Normally, this task is performed by a specific language parser. For example, we can use Eclipse JDT\footnote{\url{https://projects.eclipse.org/projects/eclipse.jdt}} for Java and LLVM\footnote{\url{https://llvm.org/}} for C/C++, et al.

\textbf{AST Iteration} Since the information we need to extract the dependency model is stored in nodes of AST, we need to traverse the AST to gather the data. During the iteration of AST, we can extract any information we need from the AST node. With the information, we can represent the dependency model using a specific entity relation representation model. 

The general process of extracting the dependency model from source code using AST is described in Fig. 2. 

\begin{figure*}[htbp]
\centering
\includegraphics[width=0.75\textwidth]{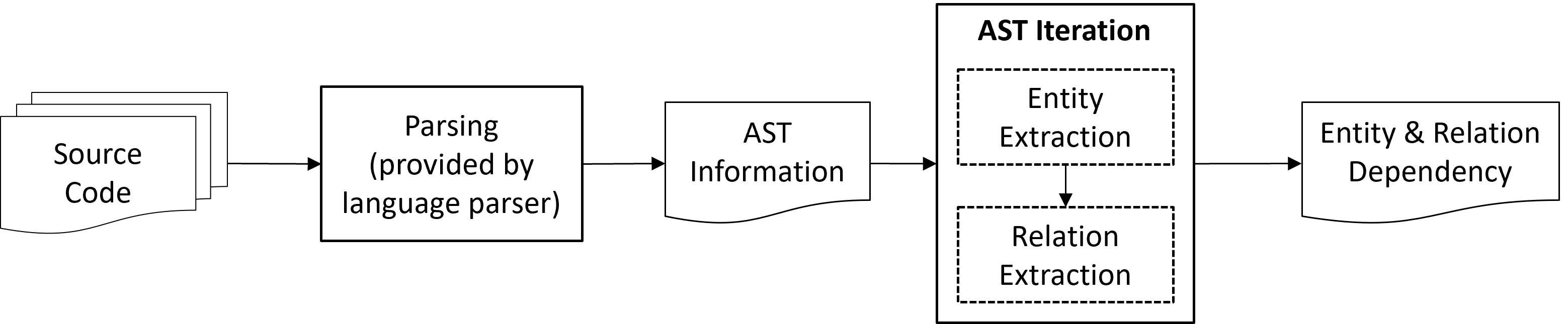}
\caption{Source code dependency extraction process based on AST}
\label{dep-ast}
\end{figure*}

Ralf Lämmel et al. \cite{10.1145/1982185.1982471} proposed a method of extracting API usage in large-scale Java projects through AST. They built a program fact database after resolving AST from source code, then performed queries on the fact database. Zhao, K. Xia et al. \cite{7424821} proposed an AST-based code plagiarism detection algorithm. They generated two abstract syntax trees for two code blocks and then performed an AST comparison algorithm for these two abstract syntax trees. To compare these two ASTs, the algorithm transformed the AST structure to a linked list by extracting information from AST node while traversing through AST. W. Jin et al. \cite{8802634} proposed a unified entity relation representation (UERR) and a method for extracting entity and relation dependencies from source code through traversing AST. They came up with a unified model of entity and relation representation to deal with multiple programming languages, then traversed AST which was generated by a language parser to extract information, and then used UERR to represent the dependency model of specific programming languages in multiple stages. Some widely used source code dependency extraction tools like depends\footnote{\url{https://github.com/multilang-depends/depends}} and understand\footnote{\url{https://scitools.com}} are AST-based as well. 

\medskip\textit{B. Heuristic Type Inference Extraction Methods}

Since possible dependency is not explicitly manifested and the information we need to extract possible dependency is not maintained in the node of the AST, it will be difficult to extract possible dependency just by traversing  the AST. To get the exact element type in dynamic programming languages such as Python and JavaScript, There are several methods to extract type inference based on heuristic methods. 

A common algorithm for doing type inference is the Hindley-Milner algorithm. It is an algorithm that infers the type of value based on its usage. It literally formalizes the intuition that types can be inferred from the functions it supports, to implement abstract, general-purpose algorithms that can automate type inference process. It was originally proposed by Haskell B. Curry and Robert Freys in 1958 \cite{curry1958} for the simple type lambda calculus. Roger Hindley \cite{Hindley1969ThePT} improved on this work and showed that the algorithm can produce the most general types in 1969. The equivalent algorithm was provided by Robin Milner \cite{MILNER1978348} in 1978. Finally, Luis Damas \cite{10.1145/582153.582176} proposed the proof that demonstrated that Milner's algorithm was sound.

Khedker et al. \cite{KHEDKER200315} proposed a data-flow based type inference extraction approach. First, they formulated information flow for type inference on flow graphs. Then they captured the impact of control flow on type information propagation to discover more precise type information.

\begin{figure*}[htbp]
\centering
\includegraphics[width=0.75\textwidth]{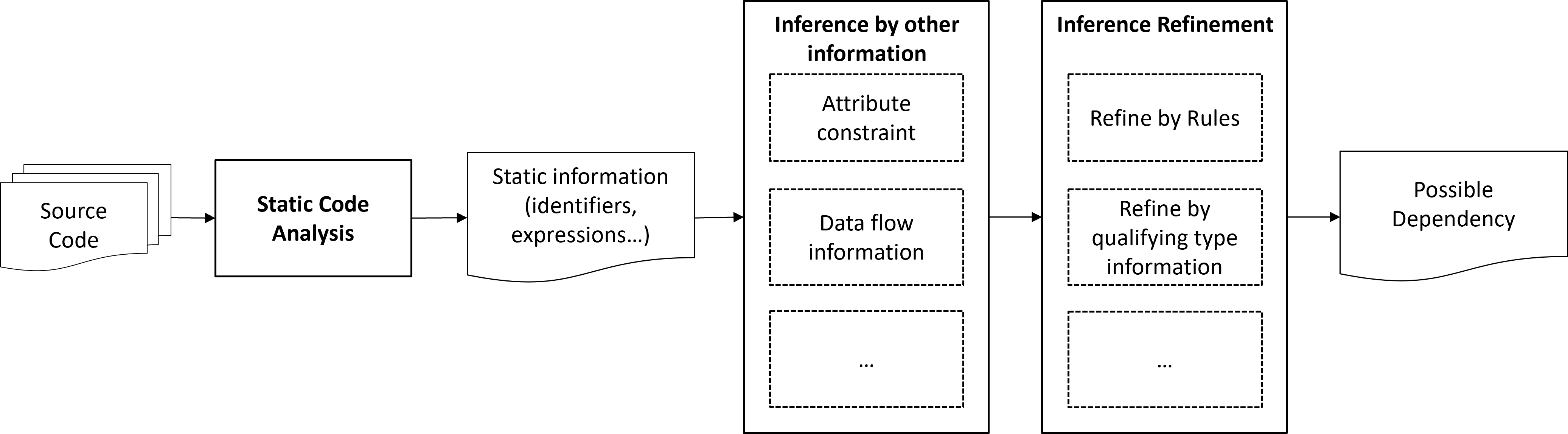}
\caption{Source code dependency extraction process based on type inference}
\label{typeinference}
\end{figure*}

W. Jin et al.\cite{9765666} proposed a system that can perform type inference and type hinting for possible dependencies in Python code to resolve possible dependencies that are difficult to obtain with other tools. Their approach assumed that possible dependencies can be obtained from type hints. First, it extracted all the necessary AST information from the source code using static analysis tools. They then evaluated the expressions whose related stub files were missing, and  used different rules to refine the set of possible dependencies. For those expressions whose source code's stub file were not missing, the source code files and stub files were merged to generate source code with type annotations. Finally, the types of entities contained in the expressions were extracted by leveraging \emph{Mypy}. In general, the process of heuristic type inference-based method is described by Fig. \ref{typeinference}.

\medskip\textit{C. Learning-based Methods}

For possible dependency, it is difficult to resolve dynamic programming features for popular dynamic programming languages such as Python and JavaScript. Traditional type inference tools normally address those challenges by only inferring types when they are certainly or very likely, which greatly restricts the amount of inferred types. \cite{10.1145/3368089.3409715} Therefore, several probabilistic approaches have been proposed for prediction types.

In general, a Learning-based source code dependency extraction method follows three steps:
\begin{itemize}
    \item 

\textit{Static Code Analysis}.
Normally the method needs to perform static code analysis to obtain all the necessary information from the source code, including code tokens, identifiers, comments, et al. The neural network needs to take the specific form of static code information as the input of training. 

\item 
\textit{Neural Prediction}.
With the information from static code analysis, the different layer of the neural network takes parts of the information and starts training based on different methods to form a type vector, which contains possible type for different elements in the source code. 

\item \textit{Search for consistent types}.
With the result of type prediction, we need to perform further type checking to get the actual type for specific elements in the source code. 
\end{itemize}

The general process of extracting source code dependency model based on Learning is described in Fig. 3. 

\begin{figure*}[htbp]
\centering
\includegraphics[width=0.75\textwidth]{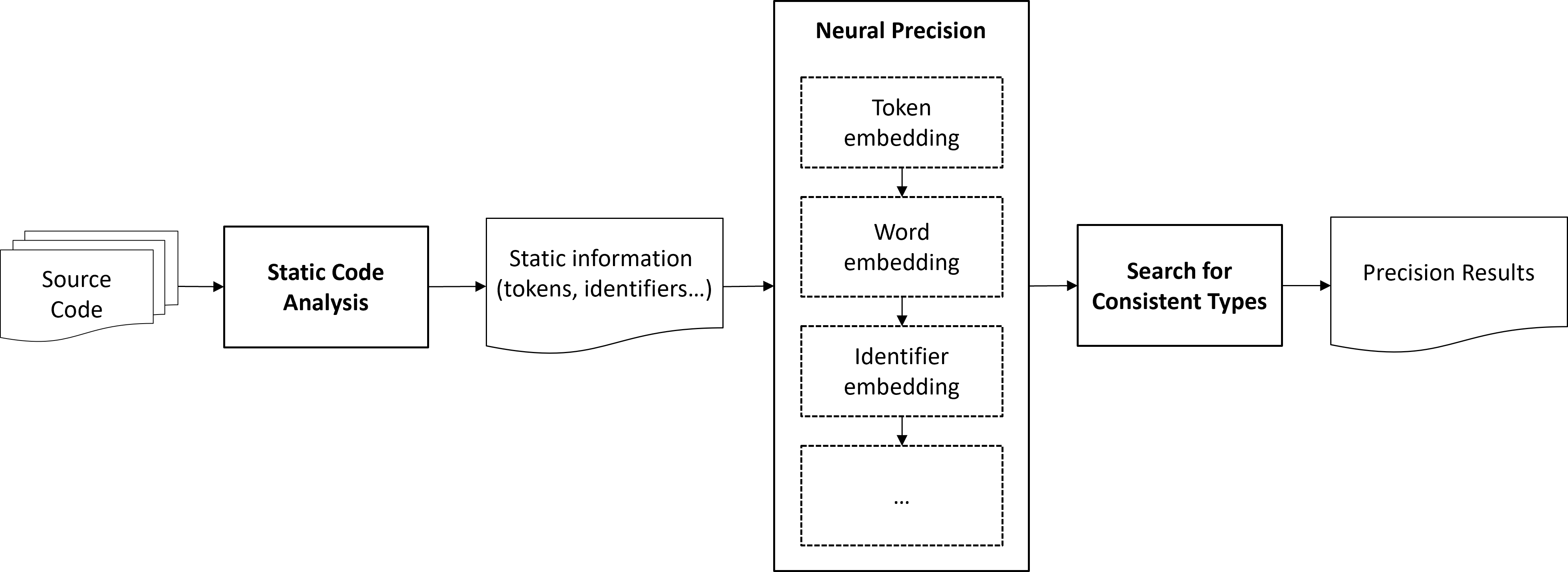}
\caption{Source code dependency extraction process based on Learning}
\label{dep-learning}
\end{figure*}

Pradel et al. \cite{10.1145/3368089.3409715} designed 4 individual sequence models to infer function types in Python. Initially, they used an static analysis approach based on AST to extract types and necessary contextual information for predicting types. Subsequently, they viewed the type prediction problem as a classification problem. A group of types was used for the model to predict the probability distributions. A single type of prediction was produced by the neural type prediction model by combining all kinds of information. After training, a ranked list of predictions was provided by the neural type prediction model for each missing annotation. Then they validated candidate assignments by leveraging the existing type checker as a filter.

Allamanis et al. \cite{10.1145/3385412.3385997} proposed a graph model as code representation and used KNN model to predict types. They developed a deep learning model by learning type spaces, along with deep similarity learning to avoid the model missing the opportunity to learn rare or previously unseen types. First, they represented Python code in graphs along with multiple design decisions. The graph encoded necessary AST information of each program. Afterwards, they used edges to encode relations between nodes, and GNN used these relations in the output representation. The AST encoded syntactic information that was traditionally used in type inference, so the GNN learned about these relationships. Finally, they connected all the identifiers to a unique node to represent the sub-token. Each of the identifier contained a sub-token. 

Wei et al. \cite{Wei2020LambdaNetPT} proposed a neural network to implement type inference for JavaScript programs. A type dependency graph was developed where nodes represented by type variables and relations between them were encoded by hyper-edges. They performed static code analysis on source code to extract the type dependency graph. Afterwards, they developed a neural architecture that consisted of two main parts: The first one was a graphical neural network. It produced a vector-valued embedding type variables by passing information along the type dependency graph. The second one was a pointer network that distributed the possible type assignments by comparing variable's type to the embedding vectors. 

\medskip\textit{D. Future Direction}

With the emergence and development of a large language model (LLM), numerous traditional technical domains now have the possibility of integrating innovation with the LLM. Static analysis methods and LLM also offer many potential directions for integration. 

The traditional static code analysis methods have consistency between multiple analysis results. They have strong interpretability, the result can be explained based on specific rules. Also, the traditional static code analysis algorithms are generalized. However, the traditional static code analysis methods are unable to effectively utilize code semantics, resulting in false positives and false negatives (FP and FN). The more comprehensive the analysis is, the higher the computational cost. Besides, the traditional static code analysis methods require specific adaptations for different programming languages.

The application of a large language model in static code analysis can effectively learn code semantics. The budgetary cost does not significantly vary based on the analysis's completeness. Also, LLM can collect multilingual datasets to support analysis in multiple programming languages. However, the application of LLM in code analysis has inconsistent results due to multiple invocations or different prompts. The interpretability of LLM is weak. Besides, it requires a high-quality training dataset while avoiding overfitting.

Taking into account the advantages and disadvantages of the two aforementioned technical methods, the static analysis method can be integrated with the LLM in the following four aspects:

\textbf{Improve Dataset Quality:}
Using static analysis to measure the code quality within the dataset to improve its quality and eliminate redundancy.

\textbf{Solve Low Complexity Problems:}
Utilizing static analysis to solve problems helps yield more accurate results for low-complexity problems.

\textbf{Rectify FP and FN:}
Rectifying the false positives (FP) and false negatives (FN) obtained from static analysis. These FP and FN results often pose a higher complexity for static analysis methods.

\textbf{Eliminate Syntax and Semantic Error:}
Performing static checks on the code along with type inference which are outputted by the large model to eliminate syntax and semantic errors or inconsistencies.

\textbf{\numberedParagraph{Dependency-based client applications: Software Refactoring Recommendations}}

Having established a solid foundation in accurately capturing code behavior, this section delves into practical software applications based on code behavior. Specifically, it focuses on automated refactoring recommendations, highlighting methods to facilitate intelligent and effective refactoring that enhance code quality.

Software refactoring is a prevalent technique utilized to enhance the internal structure of software systems while maintaining their external behaviors \cite{opdyke1992refactoring}. As software systems continue to evolve and grow, the internal architecture often needs to be restructured to adapt to new requirements, fix design defects, and maintain a high level of software quality. This makes refactoring one of the most frequent activities undertaken by developers during software maintenance and evolution, ensuring that the software system remains manageable despite increasing complexity.

\medskip\textit{A. Traditional Methods}

Historically, developers have relied on manual refactoring techniques. Fowler's seminal work provides a list of design issues in source code, termed "code smells." For each identified smell, he suggests a set of potential refactorings that developers can apply to ameliorate the issue \cite{fowler2018refactoring}. Du Bois et al. argue that refactoring opportunities align with the prospects to enhance cohesion and reduce coupling in the code \cite{du2004refactoring}. Their method, however, is limited to specific types of refactoring and a narrow range of quality metrics. Further, Murphy-Hill has introduced techniques and empirical studies to support refactoring practices \cite{murphy2008refactoring, murphy2011programmer}. In their subsequent work, they developed tools to aid developers in applying refactorings, such as selection assistants based on software structure information and program analysis techniques \cite{murphy2011programmer, murphy2008breaking}.

Innovative tools like GhostFactor, which allow developers to manually refactor code while automatically checking its correctness, have also been introduced \cite{ge2014manual}. Other tools like BeneFactor \cite{ge2011benefactor} and WitchDoctor \cite{foster2012witchdoctor} detect manual refactoring and then automate their completion.
Behavior preservation during refactoring is crucial. Some scholars suggest using invariants to identify parts of the program that need refactoring \cite{kataoka2001automated}. Additionally, Opdyke introduced the concept of defining and using pre and post-conditions along with invariants to ensure behavior is preserved during refactoring \cite{opdyke1992refactoring}.

Despite these advancements, manual refactoring continues to be a laborious task for developers, frequently necessitating a thorough exploration of the software system to identify the optimal refactoring solutions that enhance software quality and address design flaws.

\medskip\textit{B. Search-based Methods}

Search-based techniques are emerging as a popular method to automate software refactoring. This approach sees refactoring as an optimization problem, aiming to enhance system design quality based on a set of software metrics \cite{harman2012search}. Seng et al. introduced a single-objective optimization method utilizing genetic algorithms to improve various software metrics like coupling, cohesion, complexity, and stability \cite{seng2006search}. Kessentini et al. explored genetic algorithms to find the best sequence of refactorings, focusing on reducing detected design defects \cite{kessentini2011design}. In the work of Lin \cite{lin2016interactive}, hill-climbing algorithm was used to search for refactoring solutions that reduce the number of architectural inconsistencies.

However, single-objective optimization algorithms can't address the multiple dimensions of refactoring needs effectively.  In recent years, more and more works utilize multi-objective optimization-based methods for automatic refactoring recommendations. Alizadeh et al. \cite{alizadeh2018interactive} used the NSGA-II algorithm to find a set of refactoring solutions. Developers can approve, modify, or reject each refactoring, then their approach could get the refactoring recommendation that best meets the developer's preference through interaction. Rebai et al. \cite{rebai2020recommending} proposed the algorithm RefCom, which understands the developer's intention (the files expected to be refactored and the quality attributes expected to be improved) by analyzing the refactoring information committed by the developer in history. Then the NSGA-II algorithm was used to search the refactoring sequence that optimizes the software quality, and finally, the desired refactoring operation was filtered according to the identified developer intention. Abid et al. \cite{abid2021intelligent} introduced smart crossover and mutation operators that take refactoring dependencies into account to prevent the generation of invalid refactorings. They discovered that utilizing the NSGA-II algorithm can yield effective refactoring solutions, which in turn enhance the software quality. Additionally, they proposed knowledge-driven change operators along with an enhanced seeding strategy. These innovations were incorporated within a multi-objective genetic algorithm, as discussed in their subsequent work \cite{abid2021x}.

\medskip\textit{C. Machine Learning-based Methods}

Machine learning's prowess in discerning patterns and making accurate predictions has made it a valuable tool in various areas of software engineering. Over the past few years, there has been significant interest in how machine learning can be applied to the domain of software refactoring. 

In light of this trend, numerous studies have delved deeper into specific facets of refactoring through machine learning. 
Alenezi et al. \cite{alenezi2020harnessing} utilized the Gated Recurrent Unit (GRU) to predict refactoring needs at the class level. Meanwhile, Kumar et al. \cite{kumar2019method} focused on method-level predictions, employing a diverse set of classifiers and metrics. On the other hand, Sidhu et al. \cite{sidhu2022machine} proposed a neural network-based approach for refactoring models of object-oriented software impacted by functional decomposition. The approach was trained on design quality attributes to identify functional decomposition and highlight refactoring candidates. 

Adding to the dimension of refactoring categorization, RMove refactoring approach \cite{cui2022rmove} is enhanced by integrating both structural and semantic code representations. Inspired by the graph embedding techniques and the AST path techniques, the RMove approach begins by extracting both structural and semantic information from the code. Subsequent steps involve creating, normalizing, and fusing these representations. Ultimately, it trains a machine learning classifier that can recommend moving a target method to a class with closer structural and semantic alignments. Nyamawe et al. \cite{nyamawe2022mining} presented a two-tiered approach in refactoring prediction. The study introduces a two-tiered approach for refactoring prediction. Initially, a binary classifier evaluates the need for refactoring based on feature requests and code smells. Only feature requests deemed necessary for refactoring proceed to the next stage. Subsequently, a multi-label classifier identifies the specific refactoring types required for each selected request. This method effectively filters and categorizes refactoring needs. Concurrently, the research in \cite{aniche2020effectiveness} framed the prediction of refactoring opportunities as a binary classification problem. For each specific refactoring operation, a distinct model is trained to determine whether a particular code segment should be refactored accordingly. Using six machine learning algorithms, the study processed over two million labeled refactoring operations from open-source projects spanning the Apache, F-Droid, and GitHub ecosystems. These models, covering 20 varied refactoring operations at different code levels, consistently showcased an accuracy rate above 90 percent. 

The Machine Learning-based refactoring recommendation process can be distilled into the following steps, which are shown in Figure \ref{MLcase}:

\begin{figure*}[htbp]
\centering
\includegraphics[width=0.75\textwidth]{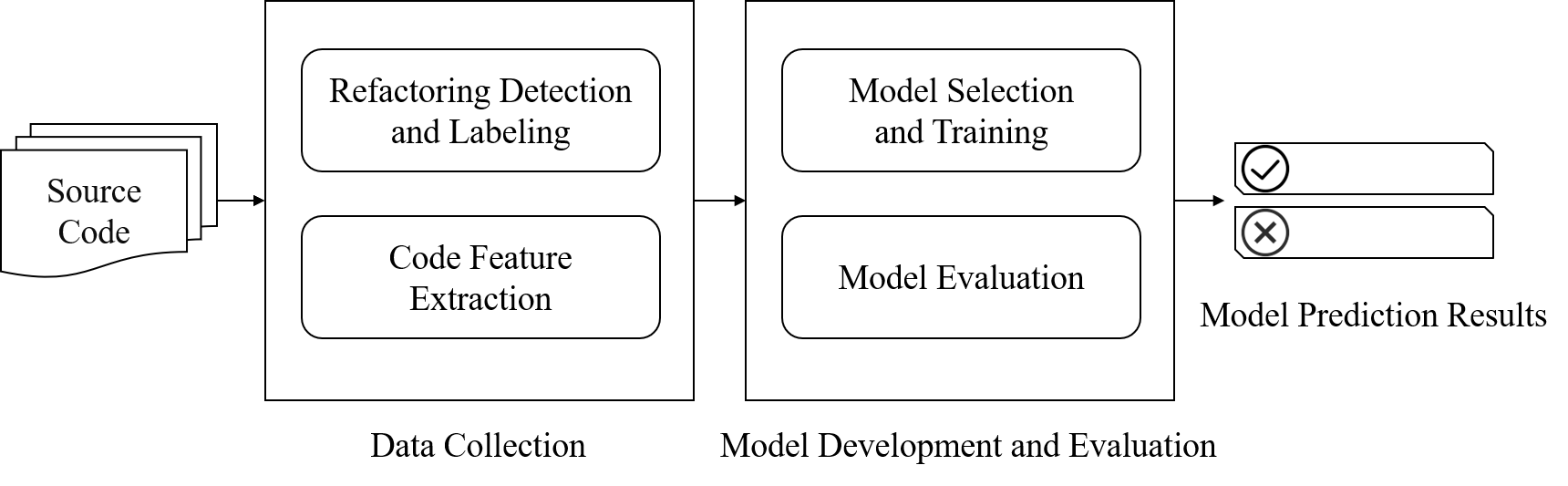}
\caption{Refactoring recommendation process based on machine learning}
\label{MLcase}
\end{figure*}

\begin{itemize}
    \item 

\textit{Refactoring Detection and Labeling.}
First, The commit history of the code is extracted, and the occurrence of refactoring operations is identified using a refactoring detection tool such as RefactoringMiner \cite{tsantalis2020refactoringminer}. Based on the results obtained from the tool, code entities are categorized accordingly:
If a refactoring operation is detected, the concerned code entity is labeled as ``Refactored''. 
Conversely, in the absence of detected refactorings, the code is labeled as ``Not Refactored''.
\item 
\textit{Code Feature Extraction.}
Code metrics, process metrics, and ownership metrics can be calculated for each code entity, specifically from the version prior to the refactoring act. The rationale behind this is to enable models to discern refactorings based on the code's state before the operation.

\item
\textit{Model Selection and Training.}
Before inputting the data into machine learning algorithms, it's crucial to preprocess the data, including merging, balancing, and scaling. For the machine learning aspect, various algorithms can be considered, such as Logistic Regression, Naive Bayes, SVM, Decision Trees, Random Forest, Neural Networks, among others. Each algorithm comes with its set of hyperparameters, which can greatly influence the performance of the model. These hyperparameters should be meticulously tuned, using methods like grid search or random search, to ensure optimal model performance.

\item 
\textit{Model Evaluation.}
Models can be validated through a stratified 10-fold cross-validation, wherein 9 folds are reserved for training, and the remaining fold serves as the test set, providing insights into the precision, recall, and accuracy of each model.

\item 
\textit{Model Prediction Results.}
Lastly, upon the culmination of the training and validation phases, each model is poised to make predictions. Given a code entity, the trained model predicts whether the entity requires a refactoring operation or not.
\end{itemize}

\medskip\textit{D. Future Direction}

Traditionally dominated by manual methodologies, the refactoring sphere is now experiencing a paradigm shift with the emergence of search-based and machine learning-based techniques. While both possess their unique strengths and have individually validated their efficacy, the intricacies and comparative advantages remain to be deeply explored.


\textbf{Challenges with ML-Based Refactoring} Machine learning (ML)-based techniques for code refactoring stand at the forefront of modern software development, promising efficiency, precision, and higher-quality recommendations. However, as with any rapidly evolving domain, there exist intrinsic challenges that need addressing to harness its full potential.

\textit{Language Bias and Multi-Language Platform Issues.} A significant chunk of research has been Java-centric, leaving a void in understanding refactoring in other languages \cite{naik2023deep}. Furthermore, with the rise of multi-language platforms, there's an urgent need for tools and methodologies that can ensure consistent refactoring across different languages, preserving the cohesiveness and functionality of multi-language software systems.

\textit{Inconsistent Representation Techniques.} ML models for code refactoring use various embedding techniques or code metrics to capture the essence of source code. The inconsistency in the adoption of these techniques and metrics across research works makes it difficult to determine a standardized approach \cite{naik2023deep}. Such discrepancies can lead to varying results, even when applied to similar codebases, complicating the selection of the optimal refactoring strategy.

\textit{Absence of Architectural Perspective.} Most ML-based refactoring approaches focus on the code's granular aspects, with limited attention to the software's overarching architecture \cite{samarthyam2016refactoring,ivers2022industry}. Refactoring from an architectural viewpoint is crucial as it can lead to systemic improvements in the software's design, performance, and maintainability. The lack of tools and methodologies that can refactor and assess the effects of such changes from an architectural lens is a significant gap in the current landscape.


\textit{Lack of Direct Comparisons on Effectiveness and Performance.} Both search-based and ML-based refactoring techniques have made a significant impact on software development. They offer increased speed, efficiency, and reduced defects. However, there remains a research gap, no studies have directly compared the performance and effectiveness differences between the search-based and ML-based refactoring recommendations.

In conclusion, while ML-based refactoring techniques present an innovative approach with numerous advantages, several challenges persist. Addressing these challenges, from language biases to the need for a broader architectural perspective, is paramount to unlocking the full potential of ML in the realm of code refactoring. Only by thoroughly understanding and addressing these issues can the software development community realize the complete benefits and transformative capabilities of ML-driven refactoring.

\textbf{Exploring the Opportunity of Language Learning Models (LLM) in Refactoring} With advancements in artificial intelligence, Language Learning Models are making notable strides in the world of software development and refactoring. Recent studies give us deep insights into the potential and challenges presented by LLM in this domain.

\textit{Data-guided and Pseudo-code Refactoring Patterns.}
The research presented in \cite{white2023chatgpt} introduces novel approaches utilizing Language Learning Models (LLM) for enhancing various facets of software development. Specifically, the paper introduces two pivotal patterns: the Data-guided Refactoring Pattern and the Pseudo-code Refactoring Pattern. Both patterns aim to simplify and automate aspects of code improvement and modification. The Data-guided Refactoring Pattern focuses on automatically altering code based on a new data format, while the Pseudo-code Refactoring Pattern allows users to guide LLM-driven code refactoring through the provision of pseudo-code outlines, reducing the need for manually specifying the minutiae.

\textit{Refactoring in Domain-Specific Programming.}
The study \cite{tarassow2023potential} delves into the instrumental role of Language Learning Models (LLM) in refining and refactoring Gretl code. Emphasizing the refactoring methodology, which aims to enhance code efficiency, maintainability, and readability without altering its core functionality, the study provides detailed examples that critically analyze the refactoring for specific functions. Although LLM demonstrates a commendable capacity to comprehend and refine intricate code structures, aligning them with the ``Clean Code'' principle, the paper highlights the imperative of vigilance. Outputs derived from LLM might contain inaccuracies, potentially leading to significant bugs, emphasizing the necessity for thorough examination.

In conclusion, LLM presents a promising frontier in the field of code refactoring. While they bring automation and precision, caution must be exercised to ensure the output's accuracy, underlining the need for a balance between automated and manual validation processes.

\textbf{\numberedParagraph{Conclusion}}

In this study, we provide a comprehensive survey of code behavior, starting with foundational entities and dependencies, and progressing to advanced techniques of code dependency extraction. Our explorations range from traditional static methods to heuristic-based and AI-driven techniques. Building on the foundation of code behavior analysis, we further survey the techniques in one representative software engineering practice namely refactoring recommendations. We outline the evolution from traditional to search-based methods, culminating in recent machine learning-driven techniques. On the whole, integrating AI into software practices highlights numerous prospective opportunities, yet also suffers from pain points in achieving seamless integration within this domain. 

\section{Conclusion}
In this paper, we have provided a detailed and up-to-date review of the current progress of research on green computing. We have discussed methods to measure green computing, techniques for improving efficiency in model training and inference phrase especially in large language model and solutions to design sustainable system. In addition to the above, we also provide numerous use cases of green computing technologies for environmental sustainability and engineering efficiency.

With the development of computationally intensive technologies such as large language models and blockchain, computing resources have become a consistent bottleneck in driving industry growth. Despite the achievements of green computing in energy efficiency and carbon reduction, it still requires further investment and effort to address this challenge. Here we list several future research directions for Green Computing.

\textbf{Model Board Include Greenness Measurements} As we mentioned in Section \ref{ai_r_and_d_trend}, Only a small portion of models report results for efficiency related measurements. If major leaderboards include measurement for "greenness", it will increase everyone's attention to the efficiency of AI solutions. With more models reporting such indicators, it will also encourage researchers to degine more environmental friendly AI techniques.

\textbf{A Widely Accepted Green Measure Framework and Toolkit} A widely accepted evaluation framework makes it easy to compare greenness of models. Due to the lack of consensus in the academic community on the green assessment framework, it is necessary to design and promote a widely used green assessment benchmark (like ImageNet). This not only allows us to evaluate the greenness of an AI solution but also enables us to track the overall development of green computing technology. 

\textbf{Exploring the Opportunity of "Smaller" Language Learning Models} As we entered the age of Large Language Models(LLM), it has widely accepted that models with more parameters are expected to understand the context better, make fewer mistakes and provide better answers. But as the models grew larger, the demand for computational resources also increased. Training these LLMs become an expensive and unsustainable task. For example, Due to the sheer size of the model GPT-4, tt is impossible to run locally on personal mobile devices or laptop even with its code open-sourced, thus only institutions or companies with abundant computing power become the only players to run them, which prevents normal users to aceess to such technology. Instead of merely adding more parameters, we should rethink the strategies: how to develop smaller yet powerful LLMs with more efficient technology that offer a better balance of performance, cost, and equality. 

\textbf{More Industrial Applications are welcomed} With its ability to handle large and complex data, AI play an active role in a variety of green applications to address sustainability challenges. We need more AI applications to reduce the impact to environments in industries like power generation, agriculture, livestock farming, transportation, etc. This also requires the government to promote corresponding policies and regulations to encourage the adoption of green computing technologies by businesses and provide financial and tax support. Additionally, we encourage research institutions and companies to work together and pay more attention to innovation in green computing, as well as share best practices and experiences.

\section*{Acknowledgments}
This work was supported by CCF-AFSG Research Fund.

\bibliographystyle{IEEEtran} 
\bibliography{main} 

\end{CJK}
\end{document}